\documentclass[twoside]{article}

\usepackage[accepted]{aistats2025}
\usepackage[round]{natbib}

\usepackage{hyperref}
\usepackage{url}
\usepackage{graphicx}
\usepackage{booktabs}
\usepackage{multirow}
\usepackage{algorithm}
\usepackage{enumitem}
\usepackage{algpseudocode} 
\usepackage{subfigure}
\usepackage{siunitx}
\usepackage{color}
\usepackage{subcaption}
\usepackage{hyperref}       
\usepackage{url}            
\usepackage{booktabs}       
\usepackage{amsfonts}       
\usepackage{nicefrac}       
\usepackage{microtype}      
\usepackage{xcolor}         
\usepackage[utf8]{inputenc} 
\usepackage[T1]{fontenc}    
\usepackage{hyperref}       
\usepackage{url}            
\usepackage{booktabs}       
\usepackage{amsfonts}       
\usepackage{nicefrac}       
\usepackage{microtype}      
\usepackage{xcolor}         

\usepackage{amsmath}
\usepackage{amssymb}
\usepackage{mathtools}
\usepackage{amsthm}
\usepackage{caption}

\begin{document}

%

%

\twocolumn[

\aistatstitle{A Unified Evaluation Framework for Epistemic Predictions}

\aistatsauthor{
\centering
Shireen Kudukkil Manchingal\textsuperscript{$1$}\textsuperscript{$\dagger$} \And 
Muhammad Mubashar\textsuperscript{$1$} \AND  
Kaizheng Wang\textsuperscript{$2$, $3$} \And 
Fabio Cuzzolin\textsuperscript{$1$}
}

\aistatsaddress{\\\textsuperscript{$1$}Artificial Intelligence, Data Analysis and Systems (AIDAS) Institute, \\School of Engineering, Computing and Mathematics, Oxford Brookes University, UK \\\textsuperscript{$2$}M-Group and DistriNet Division, Department of Computer Science, KU Leuven, Belgium \\\textsuperscript{$3$}Flanders Make@KU Leuven, Belgium\\}
]

\begin{abstract}
Predictions of uncertainty-aware models 
are diverse,
ranging from single point estimates (often averaged over prediction samples) to predictive distributions, to 
set-valued or credal-set representations. 
We propose a novel unified {evaluation} framework for uncertainty-aware classifiers, applicable to a wide range of model classes, which allows users to tailor the trade-off between accuracy and precision of predictions via a suitably designed performance metric. This 
makes possible the selection of
the most suitable model for a particular real-world application as a function of the desired trade-off. 
Our experiments, concerning 
Bayesian, ensemble, evidential, deterministic, credal and belief function classifiers on the CIFAR-10, MNIST and CIFAR-100 datasets,
show that the metric behaves as desired. 
\end{abstract}

\vspace{-6pt}
\section{INTRODUCTION}\label{sec:intro}

Awareness of uncertainty enables machine learning models to offer more
reliable and interpretable predictions, fostering trust and transparency. 
This is 
crucial in safety-critical domains such as medical diagnosis \citep{vega2022variational},
autonomous vehicles \citep{hoel2023ensemble}
or finance \citep{walters2023investor}, 
where inaccurate predictions may lead to adverse negative consequences.

In 
classification, 
in particular,
it is imperative to account for uncertainty inherent in the predictive process, called \emph{predictive uncertainty}. 
It can be represented using the predictive distribution $\hat{p}(y \mid \mathbf{x}, \mathbb{D})$, where $y \in \mathbf{Y}$ is a label, $\mathbf{x} \in \mathbf{X}$ an input instance, and $\mathbb{D} = \{ (\mathbf{x}_i, y_i)\}_{i=1}^\mathcal{N} \in \mathbf{X} \times \mathbf{Y}$ 
the available training set, $\mathcal{N}$ being the number of 
training instances.
In 
Bayesian Neural Networks (BNNs) \citep{Buntine1991BayesianB, neal2012bayesian, DBLP:journals/corr/abs-2007-06823, https://doi.org/10.48550/arxiv.1312.6114}, this uncertainty is explicitly represented through posterior predictive distributions over the parameter space. 
In Deep Ensembles (DEs) \citep{lakshminarayanan2017simple}, a predictive distribution is formed by aggregating the individual predictions generated by multiple independently trained models. 
In Evidential Deep Learning (EDL), instead, predictions are parameters of a Dirichlet posterior in the label space. In Deep Deterministic Uncertainty (DDU) models, predictions are point estimates obtained from a final softmax layer, similar to those of standard neural networks (SNN).
Finally, in imprecise-probabilistic models \citep{caprio2023imprecise, wang2024creinns}, predictions correspond to either credal sets \citep{levi1980enterprise}
or random sets
\citep{tong2021evidential, manchingal2025randomset} on the target space, 
rather than single, precise probability vectors. 

No common evaluation setting exists to date for all these diverse kinds of predictions, making it difficult for practitioners to consistently evaluate or rank uncertainty models based on their predictive performance.
In this paper, we propose a novel \emph{unified evaluation framework} 
which provides a holistic assessment of predictions produced by uncertainty-aware classifiers for model selection.
Our framework carefully balances two crucial facets: the need for accurate predictions (the distance-based ‘accuracy’ of a prediction, as detailed in Sec. \ref{sec:evaluation}) and the acknowledgment of the inherent precision or imprecision of such predictions (i.e., how ‘vague’ predictions are).

Our aim is to \emph{enable the selection of the most fitting uncertainty-aware machine learning model}, 
aligning with the requirements of specific applications.
For instance, in financial forecasting, where the stakes are high and decisions rely on precise predictions, practitioners may prioritise models that offer greater accuracy, even at the cost of slightly higher uncertainty. Conversely, in environmental monitoring applications, where adaptability and resilience are key, models providing more precise uncertainty estimates might be favored, even if predictions are slightly less accurate.
Our objective, note, is \emph{not} to propose yet another measure of uncertainty.

\begin{figure}
\vspace{-6mm}
    \centering
    \hspace*{-0.6cm}
\includegraphics[width=0.55\textwidth]{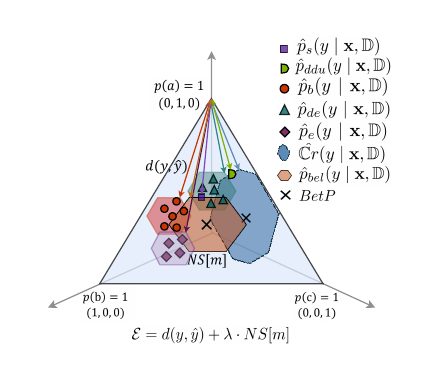}
    \vspace{-40pt}
    \caption{
    Different types of uncertainty-aware model predictions, shown in a unit {simplex} of probability distributions defined on the list of classes 
    $\mathbf{Y}= \{a, b, c\}$.
    Our proposed evaluation framework uses a metric which combines, for each input $\mathbf{x}$, a distance (arrows) 
    between the corresponding ground truth (e.g., $(0,1,0)$) and the \textit{epistemic predictions} generated by the various models (in the form of credal sets), and a measure of the extent of the credal prediction (\emph{non-specificity}).
    }
    \label{fig:simplex_KL}
    \vspace{-6.4mm}
\end{figure}

Our unified framework (Fig. \ref{fig:simplex_KL}) harnesses the fact that predictions generated by Bayesian, Ensemble,
Evidential, Deterministic
and Imprecise-probabilistic models
can all be represented as 
\emph{credal sets}, i.e., convex sets of probability vectors, within the 
simplex of all probability distributions defined on the label space,
ensuring compatibility across various model architectures and types of uncertainty model.

The \textbf{contributions} 
of our paper are: (1) A novel versatile {evaluation} framework for 
{ranking} uncertainty-aware predictions produced by a wide range of models
to facilitate model selection under uncertainty; 
(2) Instrumental to the above, a method for transforming 
predictions to credal sets in the prediction simplex;
(3) A novel metric designed to assess the trade-off between accuracy and precision of predictions; 
(4) A proof that such metric reduces to the standard KL divergence for point-wise predictions; 
(5) An experimental validation of the 
soundness
of the evaluation framework, demonstrated 
through suitable ablation studies, especially on the main parameter determining the trade-off between accuracy and precision.

\textbf{Paper outline}. After of a review of related work (Sec. \ref{sec:sota}), we examine (Sec. \ref{sec:background}) the predictions generated by various uncertainty-aware models and recall the notion of credal set. 
{In Sec. \ref{sec:credal_set}, we explain a method for transforming predictions to credal sets.}
Sec. \ref{sec:evaluation} elucidates the evaluation framework 
and introduces the novel performance metric.
Sec. \ref{sec:experiments} presents extensive experimental evidence validating our approach. 
Finally, Sec. \ref{sec:conclusion} concludes and outlines future work. 
The Appendix provides further analyses on the theory behind our evaluation approach (\S{\ref{app:theory}}), uncertainty estimation in the baseline models (\S{\ref{app:A}}) credal set computation (\S{\ref{app:approx-credal}),
experiment implementation details (\S{\ref{sec:details})
and further experimental validation and ablation studies (\S{\ref{app:additional_exp_main}}).

\vspace{-4pt}
\section{RELATED WORK} \label{sec:sota}
\vspace{-1mm}

\textbf{Aleatory vs epistemic uncertainty}.
Most scholars distinguish
\emph{aleatoric} uncertainty, caused by randomness in the data, and \emph{epistemic} uncertainty, stemming from a lack of knowledge about data distribution or model parameters \citep{kendall2017uncertainties,hullermeier2021aleatoric,manchingal2022epistemic,Zaffalon}.
Different metrics for measuring those (see \S{\ref{app:A}}) are used for different kinds of models. E.g., Bayesian models assess epistemic uncertainty using Mutual Information, while Deep Ensembles use predictive variance. 
As noted, we do not propose new uncertainty measures, but employ existing ones to propose a new fair metric and model selection approach.

\textbf{Metrics for evaluation under uncertainty}.
Evaluation metrics for probabilistic graphical models (e.g., Bayesian networks) were detailed in \citet{MARCOT201250}, including model sensitivity analysis \citep{thogmartin2010sensitivity} 
and a model emulation strategy to enable faster evaluation using surrogate models. 
\citep{SNOWLING200117}, instead, considered uncertainty for model selection, using uncertainty quantification for informed decision-making.
An `alternate hypotheses' approach was proposed by \citep{zio1996two} for evaluating model uncertainty which identifies a family of possible alternate models and probabilistically combines their predictions based on Bayesian model averaging 
\citep{droguett2008bayesian}. In 
\citep{park2011quantifying},
the model probability in the alternate hypotheses approach was further quantified by the measured deviations between data and model predictions. 
Our approach for model selection of uncertainty-aware models is unique, especially as we no longer rely solely on test accuracy to select the best model, but can instead establish a trade-off between accuracy and imprecision.
\section{CLASSES OF EPISTEMIC PREDICTIONS}
\label{sec:background}
\vspace{-3mm}

The predictions of a classifier can be plotted in the simplex (convex hull) $\mathcal{P}$ of the one-hot probability vectors assigning probability 1 to a particular class.
For instance, in a $3$-class classification scenario ($\mathbf{Y}= \{a, b, c\}$), 
the simplex would be a 2D simplex (triangle) connecting three points, each representing one of the classes, as shown in Fig. \ref{fig:simplex_KL}, which 
depicts all types of model predictions considered here.

\textbf{Standard Neural Networks} (SNNs)
predict a vector of $N$ scores, one for each class, duly \emph{calibrated} to a probability vector representing a (discrete, categorical)
probability distribution over the list of classes $\mathbf{Y}$, 
$
    \hat{p}_{s}(y \mid \mathbf{x}, \mathbb{D}),
$
which represents the probability of observing class $y$ given the input $\mathbf{x}$.

\textbf{Bayesian Neural Networks} 
(BNNs) \citep{lampinen2001bayesian, titterington2004bayesian, goan2020bayesian, hobbhahn2022fast} 
compute a predictive distribution $\hat{p}_{b}(y \mid \mathbf{x}, \mathbb{D})$ by integrating over a learnt posterior distribution of model parameters $\theta$ given training data $\mathbb{D}$. 
This
is often infeasible due to the complexity of the posterior, leading to the use of \emph{Bayesian Model Averaging} (BMA), which approximates the predictive distribution by averaging over predictions from multiple samples.
BMA inadvertently smooths out predictive distributions, diluting the inherent uncertainty present in individual models \citep{hinne2020conceptual, graefe2015limitations} as shown in Fig. \ref{fig:BMA_vs_nonBMA}, \S{\ref{app:A_BNN}}}. 
When applied to classification, BMA yields
point-wise predictions. For fair comparison and to overcome BMA's limitations, 
here we also use sets of prediction samples obtained from the different posterior weights \emph{before} averaging.

In \textbf{Deep Ensembles} 
(DEs) \citep{lakshminarayanan2017simple}, 
a prediction 
$\hat{p}_{de}(y \mid \mathbf{x}, \mathbb{D})$ for an input $\mathbf{x}$ is obtained by averaging the predictions of $K$ individual models:
$
   \hat{p}_{de}(y \mid \mathbf{x}, \mathbb{D}) = \frac{1}{K} \sum_{k=1}^{K} \hat{p}_k(y \mid \mathbf{x}, \mathbb{D}) ,
$
where $\hat{p}_{k}$ represents the prediction of the $k$-th model, trained independently with different initialisations or architectures. Uncertainty is then quantified as in \S{\ref{app:A_DE}}.

\textbf{Evidential Deep Learning} (EDL) models \citep{sensoy} make predictions $\hat{p}_{e}(y \mid \mathbf{x}, \mathbb{D})$ as parameters of a second-order Dirichlet distribution on the class space, instead of softmax probabilities. EDL uses these parameters to obtain a pointwise 
prediction.
As for BNNs, averaging may not be optimal; hence, in our evaluation, we consider individual prediction samples.

\textbf{Deep Deterministic Uncertainty} (DDU) \citep{mukhoti2021deep} models 
differ from other uncertainty-aware baselines as they do not represent uncertainty in the prediction space, but do so in the input space by identifying whether an input sample is in-distribution (iD) or out-of-distribution (OoD). As a result, DDU provides predictions $\hat{p}_{ddu}(y \mid \mathbf{x}, \mathbb{D})$ in the form of softmax probabilities akin to standard neural networks (SNNs).

\textbf{Credal Models}. 
A \emph{credal set} \citep{levi80book,zaffalon-treebased,cuzzolin2010credal,antonucci10-credal,antonucci2010credal,cuzzolin2008credal} is a convex set of probability distributions on the target (class) space (see Fig. \ref{fig:simplex_KL}). 
Credal sets can be elicited, for instance, 
from predicted
probability intervals \citep{wang2024creinns, caprio2023credal} 
$[\hat{\underline{p}}(y), \hat{\overline{p}}(y)]$, encoding lower and upper bounds, respectively, to the probabilities of each of the classes \citep{probability_interval_1994}:
\begin{equation}\label{eq:credal_interval}
    \hat{\mathbb{C}r} (y \mid \mathbf{x}, \mathbb{D}) = \{ p \in \mathcal{P} \ | \ \hat{\underline{p}}(y) \leq p(y) \leq \hat{\overline{p}}(y), \forall y \in \mathbf{Y} \}.
\end{equation}
A credal set is efficiently represented by its extremal points; their number can vary, depending on the size of the class set and the complexity of the network prediction the credal set represents.

\textbf{Belief Function Models}. 
\label{sec:belief-functions}
\emph{Belief functions} \citep{Shafer76} 
are non-additive measures independently assigning a degree of belief to each subset $A$ of their sample space, 
indicating the support for that subset (see \S{\ref{app:A_BF}}). 
\\
A predicted belief function $\hat{Bel}$ on 
$\mathbf{Y}$ is mathematically equivalent to the credal set 
\begin{equation} \label{eq:consistent}
\mathbb{C}r_{\hat{Bel}} (y \mid \mathbf{x}, \mathbb{D}) 
=
\big \{ 
p \in \mathcal{P}  \; \big | \; p(A) \geq \hat{Bel}(A) 
\big \}.
\end{equation} 
Its center of mass, termed 
\emph{pignistic probability} \citep{SMETS2005133} $BetP[\hat{Bel}]$,
assumes the role of the predictive distribution for belief function models
 \citep{tong2021evidential, manchingal2025randomset}:
$
   \hat{p}_{bel}(y \mid \mathbf{x}, \mathbb{D}) = 
    BetP[\hat{Bel}]. 
$

In summary, predictive distributions play a pivotal role in all uncertainty-aware classification models, as they encapsulate both a model's uncertainty and the inherent variability in the data. 
Our proposed unified evaluation framework (Sec. \ref{sec:evaluation}) relies on mapping all such types of predictive distributions to a credal set.

\section{MAPPING PREDICTIONS TO CREDAL SETS}
\label{sec:credal_set}
\vspace{-2pt}

Credal and belief function models directly output predictions in the form of credal sets. Traditional networks and Bayesian BMA also generate a credal set: the trivial one containing a single probability vector.
\\
Below we explain how predictions from Bayesian, Ensemble and Evidential models can also be represented by credal sets, leveraging \emph{coherent lower probabilities}. 

\textbf{Coherent lower probabilities} 
\citep{pericchi1991robust, miranda2023inner, miranda2021selection, miranda2008survey} 
model partial information about a
probability distribution (\S{\ref{app:coherent-theory}}).
Namely, 
a \emph{lower probability} is a function $\underline{P}$ 
from the power set $\mathbb{P}(\mathbf{Y})$ of all subsets of $\mathbf{Y}$ into
$[0,1]$, that 
(i) is monotone, i.e., $\underline{P}(A) \leq \underline{P}(B)$ for all $A \subseteq B$; (ii)
satisfies 
$\underline{P}(\emptyset) = 0$ and $\underline{P}(\mathbf{Y}) = 1$. 

Given such a lower probability $\underline{P}$, the associated credal set $\mathbb{C}r(\underline{P})$ comprises all probability measures in $\mathcal{P}(\mathbf{Y})$ that 
have it as lower bound: 
\begin{equation} \label{eq:lower_credal}
\mathbb{C}r(\underline{P}) = \{ P \in \mathcal{P}(\mathbf{Y}) \mid P(A) \geq \underline{P}(A) \quad  \forall \ A \subseteq \mathbf{Y} \}.
\end{equation}
$\underline{P}$ is considered \emph{coherent} if it can be computed as:
$\underline{P}(A) = \min_{P \in \mathbb{C}r(\underline{P})} P(A) \quad \forall A \subseteq \mathbf{Y}$.
From (\ref{eq:consistent}), belief functions are also coherent lower probabilities. The Möbius inverse of a coherent lower probability that is not a belief function may yield negative values. To ensure a coherent lower probability, we set all negative masses to zero.


\textbf{Computing lower probabilities from a sample of probability vectors.}
Given the multiple probability vectors predicted by
Bayesian, Ensemble or Evidential models, 
a lower probability can 
be obtained by computing the probability of any event $A$ (as the sum of the probabilities of the elements of A) for each probability vector and then taking the minimum of such $P(A)$ over the samples. 
When $|\mathbf{Y}| = N$ is large, it can be highly inefficient to compute probabilities for all $2^N$ events in the powerset $\mathbb{P}(\mathbf{Y})$. 
We thus
adopt a clustering technique \citep{manchingal2025randomset} that provides 
a fixed budget of most relevant
subsets $A \in \mathbb{P}(\mathbf{Y})$, and compute lower probabilities only for those.

\textbf{Computing masses by Moebius inverse}. To efficiently handle the credal set (\ref{eq:lower_credal}), we need to compute its vertices.
This can be done via the \emph{Möbius inverse}
\begin{equation}\label{eq:mobius_lower_prob}
    m_{\underline{P}}(A) = \sum_{B\subseteq A}{(-1)^{|A\setminus B|} {\underline{P}}(B)} \quad \forall A \subseteq \mathbf{Y},
\end{equation}
whose output is a \emph{mass function} 
\citep{Shafer76}, i.e., a set function \citep{denneberg99interaction} $m : \mathbb{P}(\mathbf{Y}) \rightarrow [0,1]$ such that $m (\emptyset) = 0$, $\sum_{A\subset\Theta} m(A) = 1$. 



\textbf{Computing the vertices of the credal set.} 
Once mass functions are computed, 
we can compute 
the vertices of the credal set $\mathbb{C}r(\underline{P})$ (\ref{eq:lower_credal}) identified by the computed lower probabilities.
Such a special credal set
has, as vertices, all the distributions $p^\pi$ induced by a permutation 
$\pi = \{ x_{\pi(1)}, \ldots, x_{\pi(|\mathbf{Y}|)} \}$ of the elements of the sample space $\mathbf{Y}$, 
of the form \citep{Chateauneuf89,cuzzolin08-credal} 
\begin{equation} \label{eq:prho}
p^\pi (x_{\pi(i)}) = \sum_{\substack{A \ni x_{\pi(i)}; \; A \not\ni x_{\pi(j)} \; \forall j<i}} m_{\underline{P}}(A).
\end{equation}
However, we do not need to compute all such vertices; instead, an approximation using only 
a subset of vertices
is often sufficient and even preferable (see \S{\ref{app:approx-credal}}).

The \textbf{overall procedure} is:
(1) Given a sample of predicted probability vectors, the lower probability of each subset of classes $A$ is computed; (2) A mass function is calculated by Möbius inverse (\ref{eq:mobius_lower_prob}); (3) The vertices of the credal set associated with the lower probability (and the original predictions) are computed.

\section{EVALUATION OF EPISTEMIC PREDICTIONS}
\label{sec:evaluation}
\vspace{-1mm}

We propose a novel \emph{unified evaluation framework} for a comprehensive assessment of these uncertainty models for \textit{model selection}, extending beyond those specifically addressed in this paper, and understand the fundamental trade-off between predictive accuracy and the imprecision inherent in uncertainty models.

The evaluation metric (see Fig. \ref{fig:simplex_KL}) combines a component measuring the distance in the probability simplex (in particular, the classical {Kullback-Leibler (KL) divergence}) between the ground truth and the prediction, represented by a credal set as explained in Secs. \ref{sec:background} and \ref{sec:credal_set}, with a \emph{non-specificity} measure \citep{dubois1987properties, dubois1993possibility, klir1987we} assessing the `imprecision' of a model, i.e., how far the prediction is from being a precise probability vector.

Classical performance metrics assume a single (max likelihood) class is predicted; thus, they do not act directly on the predicted probability vectors. By doing so, they discard a lot of information. 
We propose, instead, to evaluate predictions 
\emph{as they are} (either credal sets or as single probabilities, as special case).

\textbf{Metric}.
Mathematically, the evaluation metric $\mathcal{E}$ can be expressed, 
for a single data point, as:
\vspace{-1mm}
\begin{equation} \label{eq:metric}
    \mathcal{E} = d(y,\hat{y}) + \lambda \cdot 
    NS[m],
    \vspace{-1mm}
\end{equation}
where $d$ is the distance, however defined, between the ground truth $y$ and epistemic prediction $\hat{y}$, $\lambda$ is 
a \emph{trade-off parameter}, and 
$NS[m]$
is a non-specificity measure using as input the mass function representation (\ref{eq:mobius_lower_prob}) of the prediction.
Over a test set, the average of (\ref{eq:metric}) is measured. Low values are desirable.

The trade-off parameter, $\lambda$, controls the balance between the two terms $d$ and $NS$ in the evaluation metric $\mathcal{E}$.
Adjusting the value of $\lambda$ allows the user to emphasize either the accuracy of predictions or the precision/imprecision of the model determined by the uncertainty estimates, based on their specific requirements or preferences.
In Sec. \ref{sec:exp_trade_off}, we show experimental results on several values of $\lambda$.


\textbf{Distance computation}.
To compute the distance $d$ between (epistemic) prediction and ground truth, here we adopt, in particular, the classical
Kullback-Leibler (KL) divergence
    $
    D_{KL}(y || \hat{y})=\int_{x} y(x) \log\frac{y(x)}{\hat{y}(x)}\,dx
    $
between the ground truth $y$ and the boundary of the epistemic prediction $\hat{y}$. 
Nevertheless, our framework is agnostic to that: ablation studies on different distance measures (specifically, the Jensen-Shannon (JS) \citep{menendez1997jensen} divergence) and their effect on $\mathcal{E}$ and model rankings are shown in \S{\ref{app:ablation-kl}}.

Whatever the chosen distance,
because of the convex nature of credal sets, this reduces to computing the 
distance to the closest vertex of the credal prediction
(see \S{\ref{app:kl-theory}}).
For standard networks, this amounts to computing the KL (or JS) divergence between the ground truth distribution $y$ and the single prediction. 
For completeness, in our experiments KL is also applied to the pointwise predictions generated by Model Averaging, in the Bayesian, Ensemble and Evidential cases. 

\textbf{Non-specificity}.
Assuming a mass representation $m$ is available for the prediction (Sec. \ref{sec:credal_set}),
a \emph{non-specificity} measure is used to determine how imprecise the former is: 
the greater the non-specificity, the larger the imprecision. Non-specificity, due to \citet{dubois1987properties}, \citep{dubois1993possibility, klir1987we}, is given by the degree of concentration of the mass 
assigned to focal elements $A$ (sets of labels) in the label space $\mathbf{Y}$:
\begin{equation}\label{eq:non_spec}
    NS[m] = \sum_{A \subseteq \mathbf{Y}} {m(A) \log |A|}.
\end{equation}
Non-specificity captures epistemic uncertainty (see \S{\ref{app:nonspec-theory}}), as it is higher for 
predictions associated with larger sets of classes (or
higher mass values for those), signifying lack of confidence in the predicted probability. 
The log function in (\ref{eq:non_spec}) grows more slowly than a linear function, providing diminishing returns as the set size increases. 
Non-specificity is also tied to the size of the predicted credal set, another popular measure of epistemic uncertainty \citep{hullermeier2022quantification}; as the credal set increases in size, so does non-specificity.

Several measures of non-specificity have been proposed over the years \citep{kramosil1999nonspecificity, pal1993uncertainty, huang2014evidence, smarandache2011contradiction, abellan2000non, abellan2005difference} (\S{\ref{app:nonspec-theory}}).
Once again, our evaluation metric can be used with any suitable such measure. An ablation study is shown in \S{\ref{app:ablation-ns}}.

Note that, for 
pointwise predictions (including those generated by BMA and ensemble averaging) non-specificity goes to zero (as all non-singleton focal elements $|A|>1$ have mass 0, indicating perfect precision) and the credal set collapses to a single point. Thus, for precise predictions, (\ref{eq:metric}) reduces to the classical KL.

\textbf{Rationale and interpretation}.
The rationale of
(\ref{eq:metric}) is that
a `good' model's predictions should 
(on average)
be accurate and specific (i.e., exhibit low uncertainty), so that the model can confidently assign probabilities to different outcomes. 
In opposition, a `bad' model is 
one whose
predictions are less accurate and/or have higher uncertainty, resulting in a wider range of possible outcomes with less confidence in their probabilities.

When evaluating models, the distinction between \emph{correct} and \emph{incorrect} predictions is also important.
An
{ideal model} should 
exhibit low distance to the ground truth and low non-specificity for correct predictions. For incorrect ones, it should display high non-specificity to reflect uncertainty, and a controlled increase in its deviation from the ground truth, to prevent the prediction being close to an incorrect class altogether. Such a balance ensures that the model is both reliable and aware of its limitations, providing accurate predictions when it is confident and demonstrating uncertainty when it is not.
Hence, the best model selected by our algorithm is the one that minimizes $\mathcal{E}$. We show the evaluation metric for correct and incorrect predictions (see Tab. \ref{tab:cc_icc}) on CIFAR-10 (Fig. \ref{fig:kl_ns_eval}), MNIST (Fig. \ref{fig:app_kl_ns_eval_mnist}) and CIFAR-100 (Fig. \ref{fig:app_kl_ns_eval_cifar100}).

\textbf{Model selection and scenarios.} 
In Algorithm \ref{alg:model-selection}, we outline the process for model selection based on the evaluation metric $\mathcal{E}$. 
Using the following scenarios, we demonstrate (1) how to select the optimal $\lambda$ for a specific application, (2) its impact on the evaluation metric, and (3) how model selection differs when abstaining from a decision is allowed vs when a decision is mandatory.

\vspace{-3mm}
\begin{algorithm}[H]
\caption{Model Selection based on $\mathcal{E}$.}
   \label{alg:model-selection}
\begin{algorithmic}
   \State {\bfseries Input:} Model predictions $\hat{y}_K$ for a list of $K$ models, ground truth labels $y$, trade-off parameter $\lambda$.
    \State Select $\lambda$ based on the required precision for the task (high precision requires a high $\lambda$ and vice versa).
   \State {\bfseries Output:} Selected model $K$ with the lowest Evaluation Metric $\mathcal{E}$
   \For{each model $K$}
       \State $\circ$ Obtain predictions $\hat{y}_K$ for all available models on the test set.
       \State Compute lower probabilities $\underline{P}(\hat{y}_K)$ and mass functions $m_{\underline{P}}(\hat{y}_K)$ using (\ref{eq:mobius_lower_prob}) over the test set.
       \State $\circ$ Compute vertices of the credal set using (\ref{eq:prho}).
       \State $\circ$ Compute the minimum KL divergence between ground truth and predictions, $d(y,\hat{y}_K) = KL(y||\hat{y}_K)$ over the entire test set.
       \State $\circ$ Compute the non-specificity, $NS[m]$, of predictions on the test set using (\ref{eq:non_spec}).
       \State $\circ$ Compute $\mathcal{E} = KL(y||\hat{y}_K) + \lambda \cdot NS[m]$.
   \EndFor
   \State Select the model $K$ which has the lowest $\mathcal{E}$. 
\end{algorithmic} 
\end{algorithm}

\vspace{-5mm}
Consider crop disease classification from drone images, aiming to categorize crops as healthy, bacterial, fungal, or viral. 
Ideally, we require a model with low KL divergence and low non-specificity. But since we can choose to abstain from making a decision, we can allow non-specificity to be high. Therefore, we assign a low value to $\lambda$ as we want to favour the KL divergence more. Finally, we select the model which has lowest $\mathcal{E}$.
If the decision-making process in this scenario required more precision (i.e., decision is mandatory), even at the risk of incorrect predictions, we would increase $\lambda$ to penalize models with higher uncertainty.

Alternatively, consider the following safety-critical example: an autonomous vehicle driving through a crowded urban environment. The vehicle must make split-second decisions about detecting and responding to pedestrians, traffic signals, vehicles, and other objects. In this high-stakes scenario, \textit{abstention} or making no decision is not an option because failing to act could result in accidents, injury, or property damage. 
Here, precision and timeliness in decision-making is crucial, and a model with high uncertainty (non-specificity) cannot be allowed to abstain.
Here, the parameter $\lambda$ 
would need to be set to a higher value to ensure that the model’s non-specificity is minimized. The model must be decisive, even if there is some risk of making incorrect predictions.

This illustrates why our model selection approach is superior to conventional methods that is solely based on accuracy. In crop disease classification, the choice to abstain from decisions allows us to prioritize models that balance KL divergence and non-specificity, leading to \textit{more informed interventions}. In contrast, for autonomous vehicles, where abstention is not an option, our method focuses on minimizing non-specificity and ensuring \textit{decisiveness even under uncertainty}. This approach tailors model reliability and suitability to the specific decision-making context, beyond just accuracy.

\begin{table*}[!ht]
\caption{Comparison of Kullback-Leibler divergence (KL), Non-Specificity (NS) and Evaluation Metric ($\mathcal{E}$) for uncertainty-aware classifiers (trade-off $\lambda = 1$). Mean and standard deviation are shown for CIFAR-10, MNIST and CIFAR-100 datasets. 
}
\vspace{-1mm}
\label{tab:kl_ns_e}
\centering
\resizebox{\textwidth}{!}{
\begin{tabular}{lcccccc}
\toprule
\multirow{1}{*}{Dataset} &
  \multicolumn{1}{c}{\multirow{1}{*}{Model}}&
  \multicolumn{1}{c}{\multirow{1}{*}{Test accuracy (\%)($\uparrow$)}}&
  \multicolumn{1}{c}{\multirow{1}{*}{ECE ($\downarrow$)}}&
  \multicolumn{1}{c}{\multirow{1}{*}{KL divergence (KL)}}&
  \multicolumn{1}{c}{\multirow{1}{*}{Non-Specificity (NS)}}&
  \multicolumn{1}{c}{\multirow{1}{*}{Evaluation metric ($\mathcal{E}$)($\downarrow$)}} \\
  \midrule
\multicolumn{1}{c}{\multirow{10}{*}{CIFAR-10}}
  & LB-BNN & $89.24$ & $0.0565$ & $0.243 \pm 1.315$ & $0.166 \pm 0.398$ & $0.409 \pm 1.381$\\
  & DE & $\mathbf{93.77}$ & $\mathbf{0.0075}$ & $0.031 \pm 0.367$ & $0.385 \pm 0.715$ & $0.415 \pm 0.805$\\
  & EDL & $59.13$ & $0.0491$ & $0.002 \pm 0.011$ & $2.267 \pm 0.067$ & $ 2.270 \pm 0.066$\\
  & CreINN & $88.36$ & $0.0108$ & $0.058 \pm 0.374$ & $0.596 \pm 0.812$ & $0.654 \pm 0.892$\\
  & E-CNN & $83.5$ & $0.6497$ & $0.193 \pm 0.215$ & $1.609 \pm 0.003$ & $1.802 \pm 0.215$\\
  & RS-NN & $92.99$ & $0.0509$ & $0.398 \pm 1.895$ & $0.009 \pm 0.052$ & $\mathbf{0.407} \pm \mathbf{0.500}$\\
  \cmidrule{2-7}
  & SNN & $90.25$ & $0.0668$ & $0.481 \pm 1.797$ & $0.000 \pm 0.000$ & $0.481 \pm 1.797$\\
  & LB-BNN Avg & $89.24$ & $0.0565$  & $0.420 \pm 1.520$ & $0.000 \pm 0.000$ & $0.420 \pm 1.520$\\
   & DE Avg & $\mathbf{93.77}$ & $\mathbf{0.0075}$  & $0.195 \pm 0.763$ & $0.000 \pm 0.000$ & $\mathbf{0.195} \pm \mathbf{0.763}$\\
   & DDU & $91.34$ & $0.0439$ & $0.309 \pm 1.115$ & $0.000 \pm 0.000$ & $0.309 \pm 1.115$\\
  \midrule
\multicolumn{1}{c}{\multirow{10}{*}{MNIST}}
  & LB-BNN & $99.55$ & $0.0018$ & $0.002 \pm 0.126$ & $0.091 \pm 0.380$ & $0.093 \pm 0.401$\\
  & DE  & $99.32$ & $\mathbf{0.0012}$ & $0.002 \pm 0.072$ & $0.067 \pm 0.320$ & $0.070 \pm 0.331$\\
  & EDL & $94.42$ & $0.2418$ & $0.00007 \pm 0.002$ & $2.260 \pm	0.054$ & $2.260 \pm 0.054$\\
  & CreINN & $98.23$ & $0.0105$ & $0.071 \pm 0.609$ & $0.005 \pm 0.043$ & $0.077 \pm 0.612$\\
  & E-CNN & $99.27$ & $0.7878$ & $0.037 \pm 0.065$ & $1.608 \pm 0.004$ & $1.645 \pm 0.064$\\
  & RS-NN & $\mathbf{99.71}$ & $0.0059$ & $0.053 \pm 0.740$ & $0.001 \pm 0.016$ & $\mathbf{0.054} \pm \mathbf{0.741}$\\
  \cmidrule{2-7}
 & SNN & $98.90$ & $0.0057$ & $0.043 \pm 0.497$ & $0.000 \pm 0.000$ & $0.043 \pm 0.497$\\
    & LB-BNN Avg & $99.55$ & $0.0018$ & $0.016 \pm 0.251$ & $0.000 \pm 0.000$ & $\mathbf{0.016} \pm \mathbf{0.251}$\\
 & DE Avg & $99.32$ & $\mathbf{0.0012}$ & $0.020 \pm 0.198$ & $0.000 \pm 0.000$ & $0.020 \pm 0.198$\\
 & DDU & $99.28$ & $0.0028$ & $0.028 \pm 0.336$ & $0.000 \pm 0.000$ & $0.028 \pm 0.336$\\
  \midrule
  \multicolumn{1}{c}{\multirow{10}{*}{CIFAR-100}}
  & LB-BNN & $71.34$ & $0.1332$ & $0.146 \pm 0.504$ & $2.348 \pm 1.771$ & $2.494 \pm 1.781$\\
  & DE  & $\mathbf{74.08}$ & $\mathbf{0.0377}$ & $0.019 \pm 0.245$ & $3.182 \pm 1.909$ & $3.201 \pm 1.906$\\
  & EDL & $45.76$ & $0.3558$ & $0.010 \pm	0.192$ & $3.434 \pm 1.843$ & $3.445 \pm 1.840$\\
  & CreINN & $44.30$ & $0.1831$ & $0.723 \pm 0.646$ & $2.050 \pm	1.188$ & $2.774 \pm 0.945$\\
  & E-CNN & - & - & - & - & -\\
  & RS-NN & $71.17$ & $0.1336$ & $1.518 \pm 3.966$ & $0.569 \pm 1.164$ & $\mathbf{2.088} \pm \mathbf{4.025}$\\
  \cmidrule{2-7}
    & SNN & $65.51$ & $0.2357$ & $2.293 \pm 4.199$ & $0.000 \pm 0.000$ & $2.293 \pm 4.199$\\
  & LB-BNN Avg & $71.34$ & $0.1332$ & $1.617 \pm 2.886$ & $0.000 \pm 0.000$ & $1.617 \pm 2.886$\\
 & DE Avg &  $\mathbf{74.08}$ & $\mathbf{0.0377}$ & $1.062 \pm 1.924$ & $0.000 \pm 0.000$ & $\mathbf{1.062} \pm \mathbf{1.924}$\\
 & DDU & $73.44$ &$0.1142$ & $1.180 \pm 2.260$ & $0.000 \pm 0.000$ & $1.180 \pm 2.260$ \\
\bottomrule
\end{tabular}
} 
\vspace{-3mm}
\end{table*}

\section{EXPERIMENTS}
\label{sec:experiments}

We designed a set of baseline experiments to evaluate the behaviour of the proposed composite performance metric (Sec. \ref{sec:analysis}), using a representative for each class of predictions/models, and how these models are ranked (Sec. \ref{sec:exp_model_selection}).
We also ran several ablations studies: (1) on different measures of divergence (Kullback-Leibler vs Jensen-Shannon) (\S{\ref{app:ablation-kl}}), their effect on model selection (Tab. \ref{tab:model_rank_ablation}); (2) on different non-specificity measures (\S{\ref{app:ablation-ns}}, Tab. \ref{tab:model_rank_ablation_ns}), and why our choices of KL and NS are better; (3) on the trade-off parameter $\lambda$, to understand its effect on the ranking (Sec. \ref{sec:exp_trade_off}); (4) on the effect of the number of sample predictions produced by Bayesian and Ensemble models (\S{\ref{sec:exp_ablation_samples}}).
All implementation details are discussed in \S{\ref{sec:implementation-details}}.

\begin{figure*}[!h]
    \begin{minipage}[t]{0.47\textwidth}
    \includegraphics[width=\textwidth]{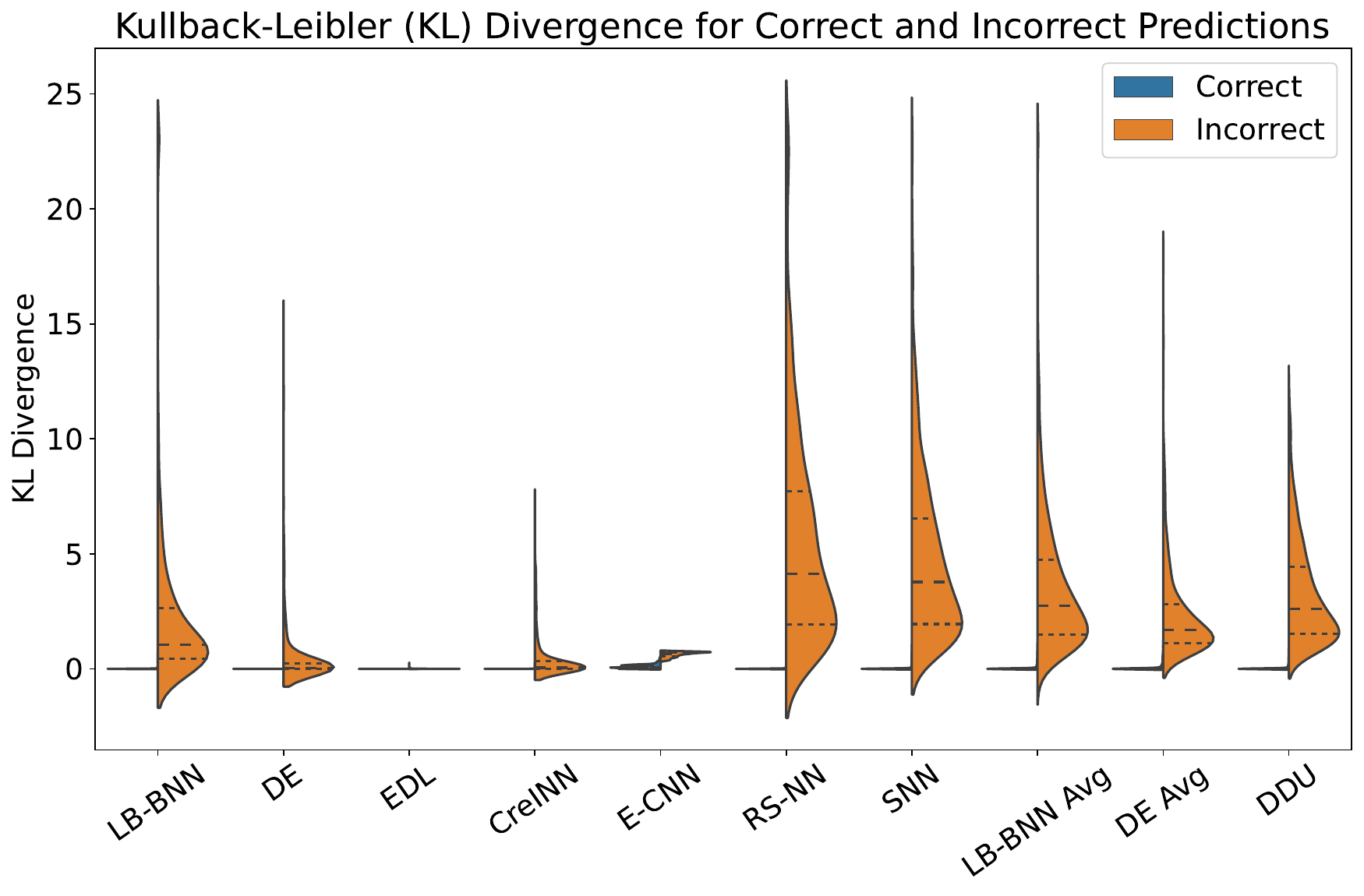}
    \end{minipage}\hspace{0.01\textwidth}
    \begin{minipage}[t]{0.47\textwidth}
    \includegraphics[width=\textwidth]{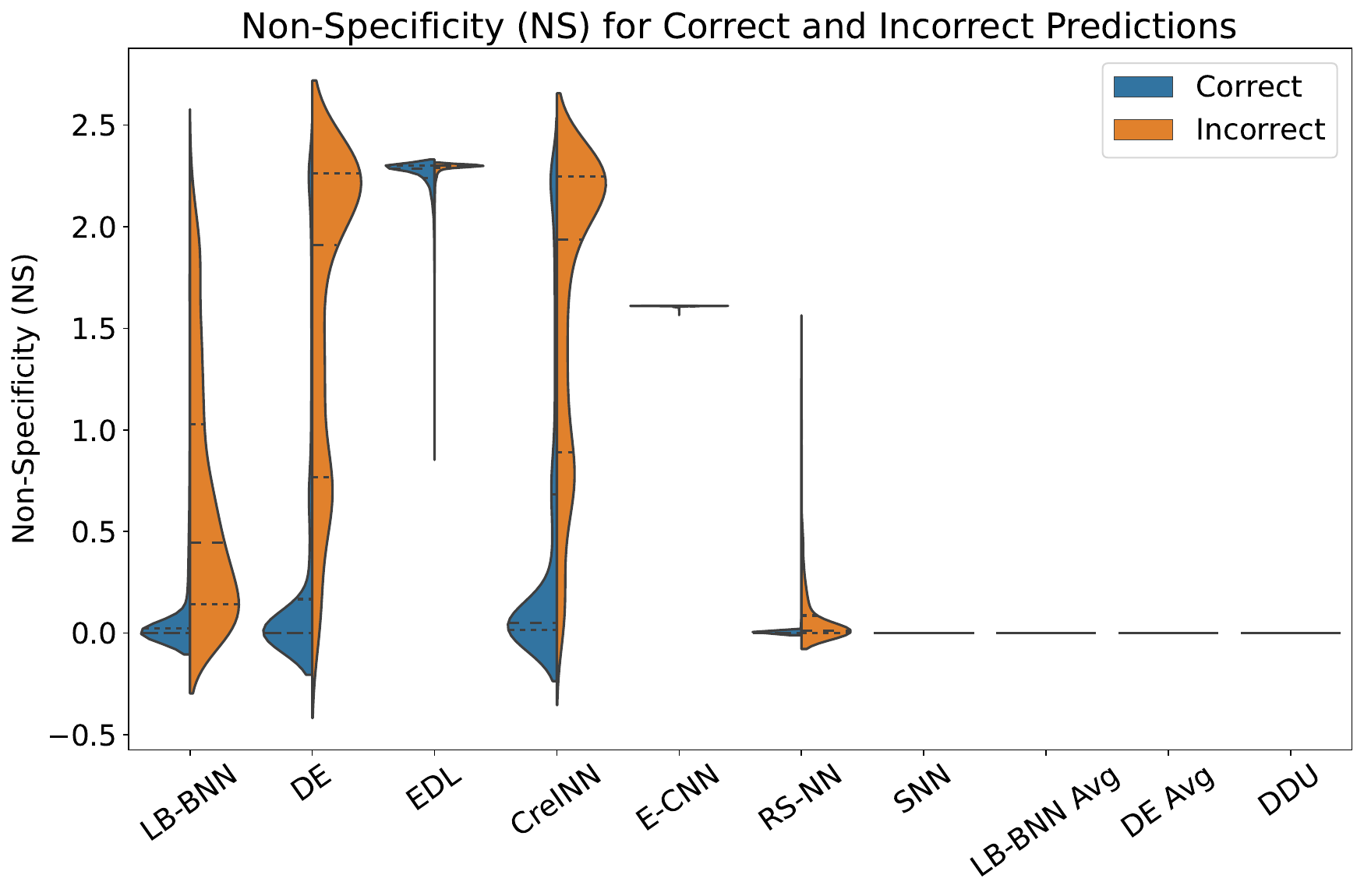}
    \end{minipage}\hspace{0.02\textwidth}
            \vspace{-5mm}
    \begin{minipage}[t]{0.47\textwidth}
    \includegraphics[width=\textwidth]{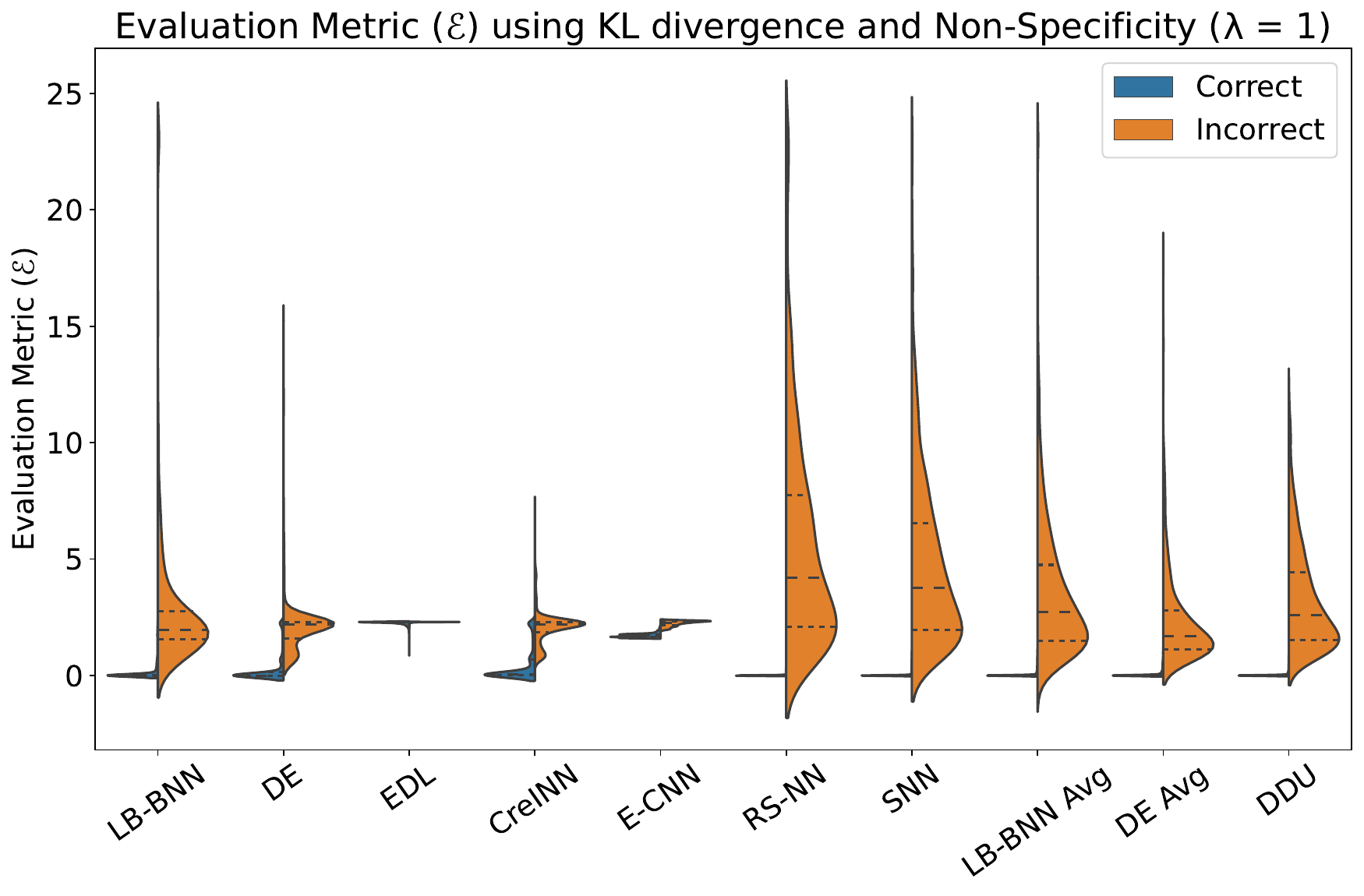}
    \end{minipage} \hspace{0.022\textwidth}
            \vspace{-5mm}
    \begin{minipage}[t]{0.5\textwidth}
    \vspace{-150pt}
    \includegraphics[width=\textwidth]{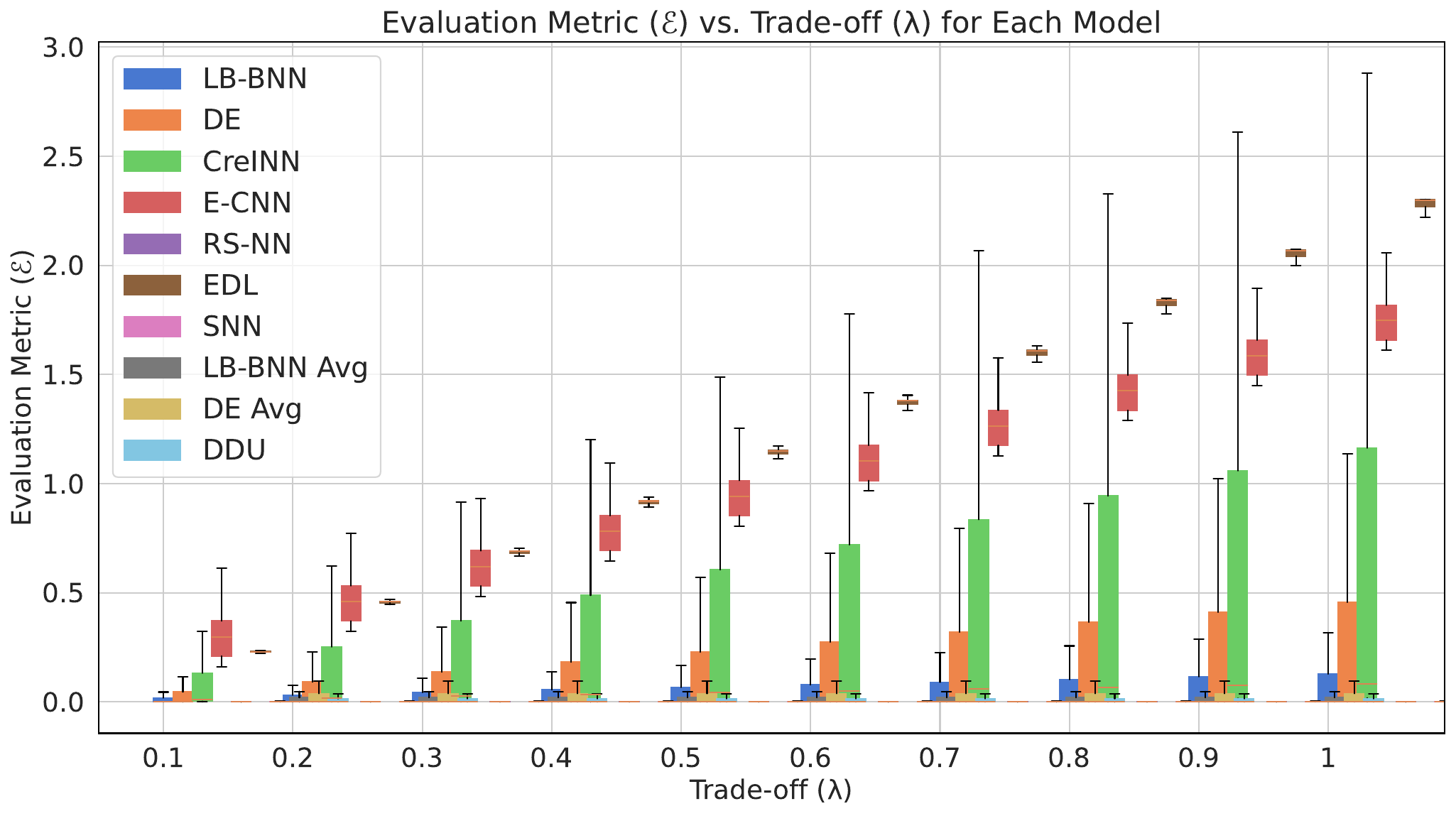}
    \vspace{-10pt}
    \end{minipage}
    \vspace{2mm}
    \caption{Measures of 
    KL divergence (top left), Non-specificity (top right), Evaluation Metric (bottom left) for both Correctly (CC) and Incorrectly Classified (ICC) samples from CIFAR-10, and Evaluation metric vs trade-off parameter (bottom right), for all models, on the CIFAR-10 dataset.
    }
    \label{fig:trade_off}
    \label{fig:kl_ns_eval}
    \vspace{-3mm}
\end{figure*}

\textbf{Baselines and datasets}. To assess the behaviour of our evaluation metric we adopted the following baselines: (1) Standard Neural Network (SNN) with no uncertainty estimation, (2) Laplace Bridge Bayesian Neural Network (LB-BNN) \citep{hobbhahn2022fast}, 
(3) Deep Ensemble (DE) \citep{lakshminarayanan2017simple},
(4) Evidential Deep Learning (EDL) \citep{sensoy}, (5) Deep Deterministic Uncertainty (DDU) \citep{mukhoti2021deep},
(6) Credal-Set Interval Neural Networks (CreINN) \citep{wang2024creinns}, (7) Evidential Convolutional Neural Network (E-CNN) \citep{tong2021evidential}, and (8) Random-Set Neural Network (RS-NN) \citep{manchingal2025randomset}.
We use MNIST \citep{LeCun2005TheMD}, CIFAR-10 \citep{CIFAR10} and CIFAR-100 \citep{krizhevsky2009learning} as datasets. 

All baselines and training details are discussed in more detail in \S{\ref{sec:details}}. Additionally, our approach is quick to implement, the entire Python notebook takes approximately 70 seconds for CIFAR-10.

\textbf{Obtaining epistemic predictions}. 
For LB-BNN, EDL and DE, we present results on both averaged predictions and credal sets generated from multiple prediction samples before averaging (100 prediction samples for LB-BNN and EDL, 15 ensembles for DE). An ablation study on number of samples vs evaluation metric $\mathcal{E}$ for these models is discussed in \S{\ref{sec:exp_ablation_samples}}.  
For CreINN, 10 samples were generated per lower and upper probability prediction. 
E-CNN directly predicts masses for all possible outcome sets (for CIFAR-10 with 10 classes, $2^{10} = 1024-1 = 1023$ outcomes, $\emptyset$ excluded), making it
computationally infeasible
for large datasets (e.g, CIFAR-100 where computing $2^{100}$ scores is inefficient). 
RS-NN generates belief functions for a budgeted set of outcomes. On CIFAR-10 and MNIST, it predicts 30 class sets 
(10 classes + 20 subsets), and 300 for CIFAR-100 (100 singletons + 200 subsets).

\vspace{-1.5mm}
\subsection{Analysis of the Evaluation Metric} \label{sec:analysis}
\vspace{-1mm}

Tab. \ref{tab:kl_ns_e} shows the 
mean and standard deviation of
Kullback-Leibler divergence (KL), 
Non-Specificity (NS) (Eq. \ref{eq:non_spec}), Evaluation Metric ($\mathcal{E}$) (Eq. \ref{eq:metric}), test accuracy (\%) and Expected Calibration Error (ECE) for 
the chosen
LB-BNN, DE, EDL, CreINN, E-CNN and RS-NN models. 
We also report these for models with point predictions only, i.e., SNN, Bayesian Model Averaged LB-BNN ({LB-BNN Avg}), averaged ensemble predictions ({DE Avg}) and DDU, for trade-off $\lambda = 1$.

Tab. \ref{tab:kl_ns_e} shows that for CIFAR-10, DE outperforms all other models with a test accuracy of $93.77\%$, a low ECE, while DE Avg and RS-NN has the lowest $\mathcal{E}$ score. Similarly, in MNIST, RS-NN achieves the highest accuracy and the LB-BNN Avg and RS-NN has the lowest $\mathcal{E}$. Conversely, for CIFAR-100, DE has a higher test accuracy of $74.08\%$, but RS-NN has the lowest $\mathcal{E}$ score. The performance is markedly low for EDL and CreINN, with E-CNN showing no results due to computational infeasibility. This model selection in Tab. \ref{tab:kl_ns_e} is reflected in the ranking for $\lambda = 1$ in Tab. \ref{tab:model_rank}.

\textbf{KL Divergence (KL)}. For point predictions generated by SNN, LB-BNN Avg, DE Avg and DDU, we computed the KL divergence between the ground truth and the point prediction. For all other models, we computed the KL between the ground truth and the boundary of the credal set (i.e., its closest vertex) (see \S{\ref{app:kl-theory}}).
The violin plot of Fig. \ref{fig:kl_ns_eval} (top left) visualises the distribution of KL divergence values for correct and incorrect predictions. 
For each model class, the KL divergence for correct predictions is represented by a single horizontal line (in blue, left), indicating that 
KL values for correct predictions are highly concentrated. 
This implies that, when the model makes correct predictions, the boundary of the predicted credal set is very close to the ground truth.
The KL divergence for incorrect predictions (in orange, right) is more broadly distributed 
for all models except EDL and E-CNN, which show consistently low KL values regardless of predictions being correct or incorrect. This could be attributed to the large size of its credal set (see \S{\ref{app:credal_set_size}}).

\textbf{Non-specificity (NS).} For models making pointwise predictions, Non-Specificity (NS) is trivially zero, as the credal set reduces to a single point 
(Sec. \ref{sec:credal_set}). For all other models, 
NS (\ref{eq:non_spec}) is shown in Fig. \ref{fig:kl_ns_eval} (top right),
and 
is generally broadly distributed across all models except, again, E-CNN and EDL, for both correct and incorrect predictions. Both E-CNN and EDL exhibits high non-specificity for all predictions, indicating that it is a more imprecise, less informative model. 

\textbf{Evaluation Metric ($\mathcal{E}$).} For point predictions, the Evaluation Metric ($\mathcal{E}$) reduces to the KL divergence between ground truth and the prediction. Fig. \ref{fig:kl_ns_eval} (bottom left) shows the Evaluation Metric ($\mathcal{E}$) for all models with the trade-off parameter $\lambda = 1$. 
One can see that, when adding NS to the mix, the concentrated distributions observed in the KL divergence for correct predictions become more distributed. 
Higher NS means that the credal set is larger, which causes more variability in $\mathcal{E}$. 
EDL shows the lowest KL for both correct and incorrect predictions 
across datasets 
but struggles with non-specificity (unusually high overall), while DE and LB-BNN balance low KL with better uncertainty handling, showing higher NS for incorrect predictions and RS-NN performs consistently well on $\mathcal{E}$ for correct predictions.
A more detailed discussion can be found in Tab. \ref{tab:cc_icc}, \S{\ref{app:addional_dataset_eval}}. 

\subsection{Evaluation of the trade-off parameter} \label{sec:exp_trade_off}

In Fig. \ref{fig:trade_off} (bottom right), a box plot is used to show the variation in Evaluation Metric ($\mathcal{E}$) values across different trade-off ($\lambda$) settings for each model. Each box in the plot represents the spread of $\mathcal{E}$ values for a specific model. The bottom and top edges of the box correspond to the 25\% and 75\% percentiles, respectively, while the horizontal lines (orange) inside the boxes indicate the median $\mathcal{E}$ value. 
For LB-BNN, DE, and CreINN, $\mathcal{E}$ steadily increases with $\lambda$, reflecting low KL and moderate non-specificity, also shown in Tab. \ref{tab:lambda-1} (\S{\ref{app:trade_off_parameter}}). In contrast, E-CNN and EDL show a shift in the range of $\mathcal{E}$ across $\lambda$ values while maintaining a consistent plot height, indicating very low KL and unusually high NS. This means that despite changes in the trade-off parameter $\lambda$, the variation of the evaluation metric $\mathcal{E}$ for E-CNN and EDL remains largely unchanged, due to these models being inherently designed to be \textit{highly imprecise} under all conditions.

\vspace{-2mm}
\subsection{Model selection} \label{sec:exp_model_selection}
\vspace{-1mm}

\begin{table}[!h]
\centering
    \caption{Model Rankings Based on KL and NS on the CIFAR-10 dataset for different values of trade-off $\lambda$. Model selection is based on the mean of Evaluation Metric ($\mathcal{E}$) with models with the lowest $\mathcal{E}$ ranking first.}
    \vspace{-2mm}
    \label{tab:model_rank}
    \resizebox{\linewidth}{!}{
    \begin{tabular}{@{}c|p{0.94\linewidth}@{}}
        \toprule
        \textbf{Trade-off ($\lambda$)} & \textbf{Model Ranking}/\textbf{Evaluation ($\mathcal{E}$) Mean} \\ \midrule
        \multirow{2}{*}{0.1} & DE, CreINN, EDL, LB-BNN, E-CNN, RS-NN\\  
        & [0.069, 0.117, 0.229, 0.259, 0.354, 0.399]\\ \midrule
        \multirow{2}{*}{0.2} & DE, CreINN, LB-BNN, RS-NN, EDL, E-CNN \\ 
        & [0.108, 0.177, 0.276, 0.309, 0.456, 0.515] \\ \midrule
        \multirow{2}{*}{0.3} & DE, CreINN, LB-BNN, RS-NN, E-CNN, EDL \\ 
        & [0.146, 0.237, 0.293, 0.309, 0.676, 0.682] \\ \midrule
        \multirow{2}{*}{0.4} & DE, CreINN, LB-BNN, RS-NN, E-CNN, EDL \\
        & [0.184, 0.296, 0.309, 0.402, 0.837, 0.909] \\ \midrule
        \multirow{2}{*}{0.5} & DE, LB-BNN, CreINN, RS-NN, E-CNN, EDL \\
        & [0.223, 0.326, 0.356, 0.403, 0.998, 1.136] \\ \midrule
        \multirow{2}{*}{0.6} & DE, LB-BNN, RS-NN, CreINN, E-CNN, EDL \\ 
        & [0.261, 0.342, 0.404, 0.415, 1.159, 1.363] \\ \midrule
        \multirow{2}{*}{0.7} & DE, LB-BNN, RS-NN, CreINN, E-CNN, EDL \\
        & [0.300, 0.359, 0.405, 0.475, 1.319, 1.589] \\ \midrule
        \multirow{2}{*}{0.8} & DE, LB-BNN, RS-NN, CreINN, E-CNN, EDL \\
        & [0.338, 0.376, 0.405, 0.535, 1.480, 1.816] \\ \midrule
        \multirow{2}{*}{0.9} & DE, LB-BNN, RS-NN, CreINN, E-CNN, EDL \\
        & [0.377, 0.392, 0.406, 0.594, 1.641, 2.043] \\ \midrule
        \multirow{2}{*}{1.0} & RS-NN, LB-BNN, DE, CreINN, E-CNN, EDL \\
        & [0.407, 0.409, 0.415, 0.654, 1.802, 2.270] \\ 
        \bottomrule
    \end{tabular} }
    \vspace{-6mm}
\end{table}

Tab. \ref{tab:model_rank} ranks all models on the basis of the mean value of the evaluation metric $\mathcal{E}$ 
on CIFAR-10, across different values of $\lambda$ (see Tabs. \ref{tab:model_selection_mnist}, \ref{tab:model_selection_cifar100} for MNIST, CIFAR-100). Lower values indicate better model performance. Here we have excluded models predicting point estimates 
to better show the effect of the trade-off parameter on $\mathcal{E}$.

DE consistently ranks first for most $\lambda$ values, followed by CreINN and LB-BNN. As $\lambda$ increases, RS-NN shows improved performance, ranking first at $\lambda = 1.0$. E-CNN and EDL perform worse overall, especially at higher $\lambda$ values, where their $\mathcal{E}$ values are significantly higher. Models like DE and CreINN rank higher at lower $\lambda$ values (i.e., leaning towards imprecision in low-risk tasks like crop disease classification), while models like RS-NN and LB-BNN perform better at higher $\lambda$ values (penalizing imprecision in high-risk scenarios, such as autonomous driving). An example of practical utility of the metric is given in \S{\ref{app:utility}}.

\vspace{-1mm}
\subsection{Limitations}\label{sec:limitations}
\vspace{-1mm}

Our metric is a function of various design choices: (i) the choice of mapping all predictions to credal sets using lower probabilities, (ii) the use of specific distance and non-specificity measures, and (iii) the metric being a linear combination of them. 
Regarding (i), other options may be possible but our choice has strong theoretical justification (\S{\ref{app:theory}}). Concerning (ii), we provided suitable ablation studies (\S{\ref{app:ablation-kl}}, \S{\ref{app:ablation-ns}}) to show why our choices of KL and NS are better.
As for (iii), we will explore alternative formulations in the future.
Also, the feasibility of computing the metric in practice depends on the specific model class. E.g., 
E-CNN could not be effectively trained on the CIFAR-100 dataset due to its inefficiency
(complexity or computational resources required).

\section{CONCLUSION}
\label{sec:conclusion}
\vspace{-1mm}

In this paper, we proposed a novel unified evaluation framework for assessing uncertainty-aware classifiers.
By combining 
measures of both distance to the ground truth and credal set size (non-specificity), the method can
offer a comprehensive understanding of the trade-off between predictive accuracy and precision (or imprecision). 
By evaluating predictions as they are, without reducing them to single values, our approach retains valuable information and provides a more nuanced assessment of model performance. 
This holistic evaluation strategy offers insights into the effectiveness and reliability of predictions across different paradigms (e.g., Bayesian vs evidential), paving the way for more informed model selection and deployment in real-world applications. Our future work will focus on studying the use of our evaluation metric as a loss function to drive the training of uncertainty-aware models.


\section*{Acknowledgement}
This work has received funding from the European Union’s Horizon 2020 Research and Innovation program under Grant Agreement No. 964505 (E-pi).

\bibliography{bibliography}

\begin{thebibliography}{}

\bibitem[Abell{\'a}n and Moral, 2000]{abellan2000non}
Abell{\'a}n, J. and Moral, S. (2000).
\newblock A non-specificity measure for convex sets of probability distributions.
\newblock {\em International journal of uncertainty, fuzziness and knowledge-based systems}, 8(03):357--367.

\bibitem[Abell{\'a}n and Moral, 2005]{abellan2005difference}
Abell{\'a}n, J. and Moral, S. (2005).
\newblock Difference of entropies as a non-specificity function on credal sets.
\newblock {\em International journal of general systems}, 34(3):201--214.

\bibitem[Antonucci and Cuzzolin, 2010a]{antonucci10-credal}
Antonucci, A. and Cuzzolin, F. (2010a).
\newblock Credal sets approximation by lower probabilities: {A}pplication to credal networks.
\newblock In Hüllermeier, E., Kruse, R., and Hoffmann, F., editors, {\em Computational Intelligence for Knowledge-Based Systems Design}, volume 6178 of {\em Lecture Notes in Computer Science}, pages 716--725. Springer, Berlin Heidelberg.

\bibitem[Antonucci and Cuzzolin, 2010b]{antonucci2010credal}
Antonucci, A. and Cuzzolin, F. (2010b).
\newblock Credal sets approximation by lower probabilities: application to credal networks.
\newblock In {\em Computational Intelligence for Knowledge-Based Systems Design: 13th International Conference on Information Processing and Management of Uncertainty, IPMU 2010, Dortmund, Germany, June 28-July 2, 2010. Proceedings 13}, pages 716--725. Springer.

\bibitem[Bernard, 2005]{bernard2005introduction}
Bernard, J.-M. (2005).
\newblock An introduction to the imprecise dirichlet model for multinomial data.
\newblock {\em International Journal of Approximate Reasoning}, 39(2-3):123--150.

\bibitem[Bronevich and Klir, 2008]{bronevich2008axioms}
Bronevich, A. and Klir, G.~J. (2008).
\newblock Axioms for uncertainty measures on belief functions and credal sets.
\newblock In {\em NAFIPS 2008-2008 Annual Meeting of the North American Fuzzy Information Processing Society}, pages 1--6. IEEE.

\bibitem[Buntine and Weigend, 1991]{Buntine1991BayesianB}
Buntine, W. and Weigend, A. (1991).
\newblock {Bayesian} back-propagation. {Technical Report FIA}-91-22.

\bibitem[Caprio et~al., 2023a]{caprio2023credal}
Caprio, M., Dutta, S., Jang, K., Lin, V., Ivanov, R., Sokolsky, O., and Lee, I. (2023a).
\newblock Credal bayesian deep learning.
\newblock {\em arXiv preprint arXiv:2302.09656}.

\bibitem[Caprio et~al., 2023b]{caprio2023imprecise}
Caprio, M., Dutta, S., Jang, K.~J., Lin, V., Ivanov, R., Sokolsky, O., and Lee, I. (2023b).
\newblock Imprecise {Bayesian} neural networks.
\newblock {\em arXiv preprint arXiv:2302.09656}.

\bibitem[Caprio et~al., 2024]{caprio2024credal}
Caprio, M., Sultana, M., Elia, E., and Cuzzolin, F. (2024).
\newblock Credal learning theory.
\newblock {\em arXiv preprint arXiv:2402.00957}.

\bibitem[Chateauneuf and Jaffray, 1989]{Chateauneuf89}
Chateauneuf, A. and Jaffray, J.-Y. (1989).
\newblock Some characterizations of lower probabilities and other monotone capacities through the use of {Möbius inversion}.
\newblock {\em Mathematical Social Sciences}, 17(3):263--283.

\bibitem[Cozman, 2000]{cozman2000credal}
Cozman, F.~G. (2000).
\newblock Credal networks.
\newblock {\em Artificial intelligence}, 120(2):199--233.

\bibitem[Cuzzolin, 2008a]{cuzzolin2008credal}
Cuzzolin, F. (2008a).
\newblock On the credal structure of consistent probabilities.
\newblock In {\em European Workshop on Logics in Artificial Intelligence}, pages 126--139. Springer.

\bibitem[Cuzzolin, 2008b]{cuzzolin08-credal}
Cuzzolin, F. (2008b).
\newblock On the credal structure of consistent probabilities.
\newblock In Hölldobler, S., Lutz, C., and Wansing, H., editors, {\em Logics in Artificial Intelligence}, volume 5293 of {\em Lecture Notes in Computer Science}, pages 126--139. Springer, Berlin Heidelberg.

\bibitem[Cuzzolin, 2010]{cuzzolin2010credal}
Cuzzolin, F. (2010).
\newblock {Credal semantics of {{Bayesian}} transformations in terms of probability intervals}.
\newblock {\em IEEE Transactions on Systems, Man, and Cybernetics, Part B: Cybernetics}, 40(2):421--432.

\bibitem[De~Campos et~al., 1994]{probability_interval_1994}
De~Campos, L.~M., Huete, J.~F., and Moral, S. (1994).
\newblock Probability intervals: A tool for uncertain reasoning.
\newblock {\em International Journal of Uncertainty, Fuzziness and Knowledge-Based Systems}, 02(02):167--196.

\bibitem[Denneberg and Grabisch, 1999]{denneberg99interaction}
Denneberg, D. and Grabisch, M. (1999).
\newblock Interaction transform of set functions over a finite set.
\newblock {\em Information Sciences}, 121(1-2):149--170.

\bibitem[Droguett and Mosleh, 2008]{droguett2008bayesian}
Droguett, E.~L. and Mosleh, A. (2008).
\newblock Bayesian methodology for model uncertainty using model performance data.
\newblock {\em Risk Analysis: An International Journal}, 28(5):1457--1476.

\bibitem[Dubois and Prade, 1987]{dubois1987properties}
Dubois, D. and Prade, H. (1987).
\newblock Properties of measures of information in evidence and possibility theories.
\newblock {\em Fuzzy sets and systems}, 24(2):161--182.

\bibitem[Dubois et~al., 1993]{dubois1993possibility}
Dubois, D., Prade, H., and Sandri, S. (1993).
\newblock On possibility/probability transformations.
\newblock In {\em Fuzzy logic: State of the art}, pages 103--112. Springer.

\bibitem[Goan and Fookes, 2020]{goan2020bayesian}
Goan, E. and Fookes, C. (2020).
\newblock Bayesian neural networks: An introduction and survey.
\newblock {\em Case Studies in Applied Bayesian Data Science: CIRM Jean-Morlet Chair, Fall 2018}, pages 45--87.

\bibitem[Graefe et~al., 2015]{graefe2015limitations}
Graefe, A., K{\"u}chenhoff, H., Stierle, V., and Riedl, B. (2015).
\newblock Limitations of ensemble {{Bayesian}} model averaging for forecasting social science problems.
\newblock {\em International Journal of Forecasting}, 31(3):943--951.

\bibitem[Hartley, 1928]{hartley1928transmission}
Hartley, R.~V. (1928).
\newblock Transmission of information 1.
\newblock {\em Bell System technical journal}, 7(3):535--563.

\bibitem[Hastings, 1970]{hastings1970monte}
Hastings, W.~K. (1970).
\newblock Monte carlo sampling methods using markov chains and their applications.

\bibitem[Hinne et~al., 2020]{hinne2020conceptual}
Hinne, M., Gronau, Q.~F., van~den Bergh, D., and Wagenmakers, E.-J. (2020).
\newblock A conceptual introduction to {{Bayesian}} model averaging.
\newblock {\em Advances in Methods and Practices in Psychological Science}, 3(2):200--215.

\bibitem[Hobbhahn et~al., 2022]{hobbhahn2022fast}
Hobbhahn, M., Kristiadi, A., and Hennig, P. (2022).
\newblock Fast predictive uncertainty for classification with {{Bayesian}} deep networks.
\newblock In {\em Uncertainty in Artificial Intelligence}, pages 822--832. PMLR.

\bibitem[Hoel et~al., 2023]{hoel2023ensemble}
Hoel, C.-J., Wolff, K., and Laine, L. (2023).
\newblock Ensemble quantile networks: Uncertainty-aware reinforcement learning with applications in autonomous driving.
\newblock {\em IEEE Transactions on Intelligent Transportation Systems}.

\bibitem[Huang et~al., 2014]{huang2014evidence}
Huang, A.~H., Zang, A.~Y., and Zheng, R. (2014).
\newblock Evidence on the information content of text in analyst reports.
\newblock {\em The Accounting Review}, 89(6):2151--2180.

\bibitem[H{\"u}llermeier et~al., 2022]{hullermeier2022quantification}
H{\"u}llermeier, E., Destercke, S., and Shaker, M.~H. (2022).
\newblock Quantification of credal uncertainty in machine learning: A critical analysis and empirical comparison.
\newblock In {\em Proceedings of the Uncertainty in Artificial Intelligence}, pages 548--557. PMLR.

\bibitem[H{\"{u}}llermeier and Waegeman, 2019]{DBLP:journals/corr/abs-1910-09457}
H{\"{u}}llermeier, E. and Waegeman, W. (2019).
\newblock Aleatoric and epistemic uncertainty in machine learning: {A} tutorial introduction.
\newblock {\em CoRR}, abs/1910.09457.

\bibitem[H{\"u}llermeier and Waegeman, 2021]{hullermeier2021aleatoric}
H{\"u}llermeier, E. and Waegeman, W. (2021).
\newblock Aleatoric and epistemic uncertainty in machine learning: An introduction to concepts and methods.
\newblock {\em Machine Learning}, 110(3):457--506.

\bibitem[Jospin et~al., 2022]{DBLP:journals/corr/abs-2007-06823}
Jospin, L.~V., Laga, H., Boussaid, F., Buntine, W., and Bennamoun, M. (2022).
\newblock Hands-on bayesian neural networks—a tutorial for deep learning users.
\newblock {\em IEEE Computational Intelligence Magazine}, 17(2):29--48.

\bibitem[Kendall and Gal, 2017]{kendall2017uncertainties}
Kendall, A. and Gal, Y. (2017).
\newblock What uncertainties do we need in {{Bayesian}} deep learning for computer vision?
\newblock {\em arXiv:1703.04977}.

\bibitem[Kingma and Ba, 2014]{kingma2014adam}
Kingma, D.~P. and Ba, J. (2014).
\newblock Adam: A method for stochastic optimization.
\newblock {\em arXiv preprint arXiv:1412.6980}.

\bibitem[Kingma and Welling, 2013]{https://doi.org/10.48550/arxiv.1312.6114}
Kingma, D.~P. and Welling, M. (2013).
\newblock Auto-encoding {Variational Bayes}.
\newblock {\em arXiv preprint arXiv:1312.6114}.

\bibitem[Klir, 1987]{klir1987we}
Klir, G.~J. (1987).
\newblock Where do we stand on measures of uncertainty, ambiguity, fuzziness, and the like?
\newblock {\em Fuzzy sets and systems}, 24(2):141--160.

\bibitem[Kolmogorov, 1965]{kolmogorov1965three}
Kolmogorov, A.~N. (1965).
\newblock Three approaches to the quantitative definition ofinformation’.
\newblock {\em Problems of information transmission}, 1(1):1--7.

\bibitem[K{\"o}rner and N{\"a}ther, 1995]{korner1995specificity}
K{\"o}rner, R. and N{\"a}ther, W. (1995).
\newblock On the specificity of evidences.
\newblock {\em Fuzzy sets and systems}, 71(2):183--196.

\bibitem[Kramosil, 1999]{kramosil1999nonspecificity}
Kramosil, I. (1999).
\newblock Nonspecificity degrees of basic probability assignments in {D}empster--{S}hafer theory.
\newblock {\em Computing and Informatics}, 18(6):559--574.

\bibitem[Krizhevsky, 2012]{krizhevsky2009learning}
Krizhevsky, A. (2012).
\newblock Learning multiple layers of features from tiny images.
\newblock {\em University of Toronto}.

\bibitem[Krizhevsky et~al., 2009]{CIFAR10}
Krizhevsky, A., Nair, V., and Hinton, G. (2009).
\newblock {CIFAR-10 (Canadian Institute For Advanced Research)}.
\newblock Technical report, CIFAR.

\bibitem[Krizhevsky et~al., 2012]{NIPS2012_c399862d}
Krizhevsky, A., Sutskever, I., and Hinton, G.~E. (2012).
\newblock Imagenet classification with deep convolutional neural networks.
\newblock In Pereira, F., Burges, C., Bottou, L., and Weinberger, K., editors, {\em Advances in Neural Information Processing Systems}, volume~25. Curran Associates, Inc.

\bibitem[Lakshminarayanan et~al., 2017]{lakshminarayanan2017simple}
Lakshminarayanan, B., Pritzel, A., and Blundell, C. (2017).
\newblock Simple and {Scalable Predictive Uncertainty Estimation using Deep Ensembles}.
\newblock {\em Advances in Neural Information Processing Systems}, 30.

\bibitem[Lampinen and Vehtari, 2001]{lampinen2001bayesian}
Lampinen, J. and Vehtari, A. (2001).
\newblock Bayesian approach for neural networks—review and case studies.
\newblock {\em Neural networks}, 14(3):257--274.

\bibitem[LeCun and Cortes, 2005]{LeCun2005TheMD}
LeCun, Y. and Cortes, C. (2005).
\newblock The {MNIST} database of handwritten digits.
\newblock In {\em Proceedings of the IEEE Conference on Computer Vision and Pattern Recognition (CVPR)}, pages 1--9.

\bibitem[Levi, 1980a]{levi1980enterprise}
Levi, I. (1980a).
\newblock {\em The enterprise of knowledge: An essay on knowledge, credal probability, and chance}.
\newblock MIT press.

\bibitem[Levi, 1980b]{levi80book}
Levi, I. (1980b).
\newblock {\em {The enterprise of knowledge: An essay on knowledge, credal probability, and chance}}.
\newblock The MIT Press, Cambridge, Massachusetts.

\bibitem[Manchingal and Cuzzolin, 2022]{manchingal2022epistemic}
Manchingal, S.~K. and Cuzzolin, F. (2022).
\newblock Epistemic deep learning.
\newblock {\em arXiv preprint arXiv:2206.07609}.

\bibitem[Manchingal et~al., 2025]{manchingal2025randomset}
Manchingal, S.~K., Mubashar, M., Wang, K., Shariatmadar, K., and Cuzzolin, F. (2025).
\newblock Random-set neural networks.
\newblock In {\em The Thirteenth International Conference on Learning Representations}.

\bibitem[Marcot, 2012]{MARCOT201250}
Marcot, B.~G. (2012).
\newblock Metrics for evaluating performance and uncertainty of bayesian network models.
\newblock {\em Ecological Modelling}, 230:50--62.

\bibitem[Men{\'e}ndez et~al., 1997]{menendez1997jensen}
Men{\'e}ndez, M.~L., Pardo, J., Pardo, L., and Pardo, M. (1997).
\newblock The jensen-shannon divergence.
\newblock {\em Journal of the Franklin Institute}, 334(2):307--318.

\bibitem[Miranda, 2008]{miranda2008survey}
Miranda, E. (2008).
\newblock A survey of the theory of coherent lower previsions.
\newblock {\em International Journal of Approximate Reasoning}, 48(2):628--658.

\bibitem[Miranda et~al., 2023]{miranda2023inner}
Miranda, E., Montes, I., and Presa, A. (2023).
\newblock Inner approximations of coherent lower probabilities and their application to decision making problems.
\newblock {\em Annals of Operations Research}, pages 1--39.

\bibitem[Miranda et~al., 2021]{miranda2021selection}
Miranda, E., Montes, I., and Vicig, P. (2021).
\newblock On the selection of an optimal outer approximation of a coherent lower probability.
\newblock {\em Fuzzy Sets and Systems}, 424:1--36.

\bibitem[Mukhoti et~al., 2021]{mukhoti2021deep}
Mukhoti, J., Kirsch, A., van Amersfoort, J., Torr, P.~H., and Gal, Y. (2021).
\newblock Deep deterministic uncertainty: A simple baseline.
\newblock {\em arXiv preprint arXiv:2102.11582}.

\bibitem[Neal, 2012]{neal2012bayesian}
Neal, R.~M. (2012).
\newblock {\em {Bayesian} learning for neural networks}, volume 118.
\newblock Springer Science \& Business Media.

\bibitem[Pal et~al., 1993]{pal1993uncertainty}
Pal, N.~R., Bezdek, J.~C., and Hemasinha, R. (1993).
\newblock Uncertainty measures for evidential reasoning ii: A new measure of total uncertainty.
\newblock {\em International Journal of Approximate Reasoning}, 8(1):1--16.

\bibitem[Park and Grandhi, 2011]{park2011quantifying}
Park, I. and Grandhi, R.~V. (2011).
\newblock Quantifying multiple types of uncertainty in physics-based simulation using bayesian model averaging.
\newblock {\em AIAA journal}, 49(5):1038--1045.

\bibitem[Pericchi and Walley, 1991]{pericchi1991robust}
Pericchi, L.~R. and Walley, P. (1991).
\newblock Robust bayesian credible intervals and prior ignorance.
\newblock {\em International Statistical Review/Revue Internationale de Statistique}, pages 1--23.

\bibitem[Sensoy et~al., 2018]{sensoy}
Sensoy, M., Kaplan, L., and Kandemir, M. (2018).
\newblock Evidential deep learning to quantify classification uncertainty.
\newblock In {\em Proceedings of the 32nd International Conference on Neural Information Processing Systems}, NIPS'18, page 3183–3193, Red Hook, NY, USA. Curran Associates Inc.

\bibitem[Shafer, 1976]{Shafer76}
Shafer, G. (1976).
\newblock {\em A mathematical theory of evidence}, volume~42.
\newblock Princeton university press.

\bibitem[Shlens, 2014]{shlens2014notes}
Shlens, J. (2014).
\newblock Notes on kullback-leibler divergence and likelihood.
\newblock {\em arXiv preprint arXiv:1404.2000}.

\bibitem[Smarandache et~al., 2011]{smarandache2011contradiction}
Smarandache, F., Martin, A., and Osswald, C. (2011).
\newblock Contradiction measures and specificity degrees of basic belief assignments.
\newblock In {\em 14th International Conference on Information Fusion}, pages 1--8. IEEE.

\bibitem[Smets, 1983]{smets1983information}
Smets, P. (1983).
\newblock Information content of an evidence.
\newblock {\em International Journal of Man-Machine Studies}, 19(1):33--43.

\bibitem[Smets, 2005]{SMETS2005133}
Smets, P. (2005).
\newblock Decision making in the {TBM}: the necessity of the pignistic transformation.
\newblock {\em International Journal of Approximate Reasoning}, 38(2):133--147.

\bibitem[Snowling and Kramer, 2001]{SNOWLING200117}
Snowling, S. and Kramer, J. (2001).
\newblock Evaluating modelling uncertainty for model selection.
\newblock {\em Ecological Modelling}, 138(1):17--30.

\bibitem[Song et~al., 2018]{song2018evidence}
Song, Y., Wang, X., Wu, W., Quan, W., and Huang, W. (2018).
\newblock Evidence combination based on credibility and non-specificity.
\newblock {\em Pattern Analysis and Applications}, 21:167--180.

\bibitem[Thogmartin, 2010]{thogmartin2010sensitivity}
Thogmartin, W.~E. (2010).
\newblock Sensitivity analysis of north american bird population estimates.
\newblock {\em Ecological Modelling}, 221(2):173--177.

\bibitem[Titterington, 2004]{titterington2004bayesian}
Titterington, D.~M. (2004).
\newblock Bayesian methods for neural networks and related models.
\newblock {\em Statistical science}, pages 128--139.

\bibitem[Tong et~al., 2021]{tong2021evidential}
Tong, Z., Xu, P., and Denoeux, T. (2021).
\newblock An evidential classifier based on {Dempster-Shafer} theory and deep learning.
\newblock {\em Neurocomputing}, 450:275--293.

\bibitem[Troffaes, 2007]{troffaes2007decision}
Troffaes, M.~C. (2007).
\newblock Decision making under uncertainty using imprecise probabilities.
\newblock {\em International journal of approximate reasoning}, 45(1):17--29.

\bibitem[Ullah, 1996]{ullah1996entropy}
Ullah, A. (1996).
\newblock Entropy, divergence and distance measures with econometric applications.
\newblock {\em Journal of Statistical Planning and Inference}, 49(1):137--162.

\bibitem[Vega and Todd, 2022]{vega2022variational}
Vega, M.~A. and Todd, M.~D. (2022).
\newblock {A variational Bayesian neural network for structural health monitoring and cost-informed decision-making in miter gates}.
\newblock {\em Structural Health Monitoring}, 21(1):4--18.

\bibitem[Virtanen et~al., 2020]{virtanen2020scipy}
Virtanen, P., Gommers, R., Oliphant, T.~E., Haberland, M., Reddy, T., Cournapeau, D., Burovski, E., Peterson, P., Weckesser, W., Bright, J., et~al. (2020).
\newblock Scipy 1.0: fundamental algorithms for scientific computing in python.
\newblock {\em Nature methods}, 17(3):261--272.

\bibitem[Walley, 1991]{walley1991statistical}
Walley, P. (1991).
\newblock Statistical reasoning with imprecise probabilities.

\bibitem[Walters et~al., 2023]{walters2023investor}
Walters, D.~J., {\"U}lk{\"u}men, G., Tannenbaum, D., Erner, C., and Fox, C.~R. (2023).
\newblock Investor behavior under epistemic vs. aleatory uncertainty.
\newblock {\em Management Science}, 69(5):2761--2777.

\bibitem[Wang et~al., 2024a]{wang2024credal}
Wang, K., Cuzzolin, F., Shariatmadar, K., Moens, D., and Hallez, H. (2024a).
\newblock Credal wrapper of model averaging for uncertainty estimation on out-of-distribution detection.
\newblock {\em arXiv preprint arXiv:2405.15047}.

\bibitem[Wang et~al., 2024b]{wang2024creinns}
Wang, K., Shariatmadar, K., Manchingal, S.~K., Cuzzolin, F., Moens, D., and Hallez, H. (2024b).
\newblock Creinns: Credal-set interval neural networks for uncertainty estimation in classification tasks.
\newblock {\em arXiv preprint arXiv:2401.05043}.

\bibitem[Yager, 2008]{yager2008entropy}
Yager, R.~R. (2008).
\newblock Entropy and specificity in a mathematical theory of evidence.
\newblock {\em Classic works of the Dempster-Shafer theory of belief functions}, pages 291--310.

\bibitem[Zaffalon, 2002]{Zaffalon}
Zaffalon, M. (2002).
\newblock The {Naive Credal Classifier}.
\newblock {\em Journal of Statistical Planning and Inference - J STATIST PLAN INFER}, 105:5--21.

\bibitem[Zaffalon and Fagiuoli, 2003]{zaffalon-treebased}
Zaffalon, M. and Fagiuoli, E. (2003).
\newblock Tree-based credal networks for classification.
\newblock {\em Reliable computing}, 9(6):487--509.

\bibitem[Zio and Apostolakis, 1996]{zio1996two}
Zio, E. and Apostolakis, G. (1996).
\newblock Two methods for the structured assessment of model uncertainty by experts in performance assessments of radioactive waste repositories.
\newblock {\em Reliability Engineering \& System Safety}, 54(2-3):225--241.

\end{thebibliography}
\bibliographystyle{apalike}

\section*{Checklist}



 \begin{enumerate}

 \item For all models and algorithms presented, check if you include:
 \begin{enumerate}
   \item A clear description of the mathematical setting, assumptions, algorithm, and/or model. [\textbf{Yes}/No/Not Applicable]
   \item An analysis of the properties and complexity (time, space, sample size) of any algorithm. [\textbf{Yes}/No/Not Applicable]
   \item (Optional) Anonymized source code, with specification of all dependencies, including external libraries. [\textbf{Yes}/No/Not Applicable]
 \end{enumerate}

(a) \textbf{Answer: Yes}. Mathematical setting and assumptions detailed in Secs. \ref{sec:intro}, \ref{sec:background} and \ref{sec:evaluation}. Models and training details mentioned in Sec. \ref{sec:experiments} under "Baselines and datasets", and further detailed in \S{\ref{sec:details}}, \S{\ref{sec:implementation-details}}.

(b) \textbf{Answer: Yes}. Time taken for implementation provided in Sec. \ref{sec:experiments}.

(c) \textbf{Answer: Yes}. Source code submitted as anonymized repository.

 \item For any theoretical claim, check if you include:
 \begin{enumerate}
   \item Statements of the full set of assumptions of all theoretical results. [Yes/No/\textbf{Not Applicable}]
   \item Complete proofs of all theoretical results. [Yes/No/\textbf{Not Applicable}]
   \item Clear explanations of any assumptions. [Yes/No/\textbf{Not Applicable}]     
 \end{enumerate}

(a) \textbf{Answer: Not Applicable}.

(b) \textbf{Answer: Not Applicable}. 

(c) \textbf{Answer: Not Applicable}. 

 \item For all figures and tables that present empirical results, check if you include:
 \begin{enumerate}
   \item The code, data, and instructions needed to reproduce the main experimental results (either in the supplemental material or as a URL). [\textbf{Yes}/No/Not Applicable]
   \item All the training details (e.g., data splits, hyperparameters, how they were chosen). [\textbf{Yes}/No/Not Applicable]
         \item A clear definition of the specific measure or statistics and error bars (e.g., with respect to the random seed after running experiments multiple times). [\textbf{Yes}/No/Not Applicable]
         \item A description of the computing infrastructure used. (e.g., type of GPUs, internal cluster, or cloud provider). [\textbf{Yes}/No/Not Applicable]
 \end{enumerate}

(a) \textbf{Answer: Yes}. Code for all experiments (including plots, tables, ablation studies) provided as anonymous Github repository.

(b) \textbf{Answer: Yes}. All training details provided in \S{\ref{sec:implementation-details}}.

(c) \textbf{Answer: Yes}. All plots have error bars reported (see Figs. \ref{fig:kl_ns_eval}, \ref{fig:app_kl_ns_eval_mnist} and \ref{fig:app_kl_ns_eval_cifar100}). The tables have mean and standard deviation reported (see Tabs. \ref{tab:kl_ns_e}, \ref{tab:cc_icc}, \ref{tab:lambda-1} and \ref{tab:approx_naive}). 

(d) \textbf{Answer: Yes}. All training details provided in \S{\ref{sec:implementation-details}}.

 \item If you are using existing assets (e.g., code, data, models) or curating/releasing new assets, check if you include:
 \begin{enumerate}
   \item Citations of the creator If your work uses existing assets. [\textbf{Yes}/No/Not Applicable]
   \item The license information of the assets, if applicable. [Yes/No/\textbf{Not Applicable}]
   \item New assets either in the supplemental material or as a URL, if applicable. [Yes/No/\textbf{Not Applicable}]
   \item Information about consent from data providers/curators. [Yes/No/\textbf{Not Applicable}]
   \item Discussion of sensible content if applicable, e.g., personally identifiable information or offensive content. [Yes/No/\textbf{Not Applicable}]
 \end{enumerate}

(a) \textbf{Answer: Yes}. All assets used are open-source, with citations provided for baseline models, datasets, etc. (Sec. \ref{sec:experiments}).

(b) \textbf{Answer: Not Applicable}. All assets used are open-source, with citations provided for baseline models, datasets, etc. (Sec. \ref{sec:experiments}).

(c) \textbf{Answer: Not Applicable}. 

(d) \textbf{Answer: Not Applicable}. All assets used are open-source, with citations provided for baseline models, datasets, etc. (Sec. \ref{sec:experiments}).

(e) \textbf{Answer: Not Applicable}.

 \item If you used crowdsourcing or conducted research with human subjects, check if you include:
 \begin{enumerate}
   \item The full text of instructions given to participants and screenshots. [Yes/No/\textbf{Not Applicable}]
   \item Descriptions of potential participant risks, with links to Institutional Review Board (IRB) approvals if applicable. [Yes/No/\textbf{Not Applicable}]
   \item The estimated hourly wage paid to participants and the total amount spent on participant compensation. [Yes/No/\textbf{Not Applicable}]
 \end{enumerate}

(a) \textbf{Answer: Not Applicable}. No crowdsourcing or research with human subjects.

(b) \textbf{Answer: Not Applicable}. 

(c) \textbf{Answer: Not Applicable}.

 \end{enumerate}


\newpage
\onecolumn
\appendix
\section*{
    \centering
    \Large \bfseries A Unified Evaluation Framework for Epistemic Predictions \\ \Large Appendix
}

\section{\MakeUppercase{Theoretical discussion of the proposed Evaluation Framework}}
\label{app:theory}

\subsection{Coherent lower probabilities} \label{app:coherent-theory}

A strong theoretical underpinning for reasoning with coherent lower probabilities (and, therefore, the corresponding credal sets) is that it allows us to comply with the coherence principle. In a Bayesian context, individual predictions (such as those of networks with specified weights) can be interpreted as subjective pieces of evidence about a fact (e.g., what is the true class of an input observation). Coherence ensures that one realises the full implications of such partial assessments.

For instance, assume that the available evidence is that on a binary classification problem is that $\underline{P}(0) = 1/4$, $\underline{P}(1) = 1/4$, $\underline{P}({0,1}) = 1/4$. In the behavioural interpretation of probability [Walley], in which lower probability of an event $A$ is the upper bound to the price one is willing to pay for betting on outcome $A$, this means that one is willing to pay up to 1/4 for betting on either 0 or 1. But then one should be willing to bet 1/4 + 1/4 = 1/2 for betting on ${0,1}$, which is over the specified lower probability for it. Hence, the above specifications are incoherent. Incoherence means that lower probabilities (specified prices) on some events effectively imply lower probabilities on other events which are incompatible.

Now, learning a coherent lower probability from sample predictions does ensure coherence. This is explained, for instance, in \citep{bernard2005introduction}. The concept was originally introduced by Walley in his general treatment of imprecise probabilities \citep{walley1991statistical}. Further, as observed e.g. in \citep{caprio2024credal}, working with credal sets allows us to hedge against model misspecification, when the set of probability distributions considered by a statistician does not include the distribution that generated the observed data.

In some cases, it may be useful to consider the conjugate function of a lower probability, referred to as the upper probability, $\overline{P}(A) = 1 - \underline{P}(A^c)$ for every $A \subseteq \mathbf{Y}$. The upper and lower probabilities provide upper and lower bounds respectively for the true distribution $\mathbf{P}(A) \in \mathbb{C}r$. Notably, for any coherent lower probability $\underline{P}$, its conjugate upper probability $\overline{P}$ satisfies:
\begin{equation}
    \overline{P}(A) = \max_{P \leq \overline{P}} P(A) \quad \forall \ A \subseteq \mathbf{Y}
\end{equation}
This equation underscores the equivalence between the probabilistic information conveyed by the lower and upper probabilities, indicating that working with either one suffices.

\subsection{Kullback-Leibler (KL) divergence} \label{app:kl-theory}

\textbf{Why KL over L2 and other measures?} KL divergence measures the distance between two probability distributions whereas L2 measures the Euclidean distance between points or vectors. KL divergence has a clear interpretation in information theory as the amount of information lost when one distribution approximates the other \citep{shlens2014notes}, i.e, KL can capture the relative differences between probabilities better. Other divergence measures, such as Jensen-Shannon divergence \citep{menendez1997jensen}, are symmetric and handles small probabilities well, but KL is often preferred as it penalizes cases where the approximate distribution assigns probability to regions where the true distribution assigns none. This property ensures that both the distributions closely match in significant regions \citep{ullah1996entropy}, avoiding substantial probability mass in unlikely areas. Therefore, KL divergence can be advantageous over JS divergence, particularly in its sensitivity to differences between distributions, especially when small probabilities are involved. Unlike JS, which is symmetric and bounded, KL divergence can grow unbounded when there is a significant mismatch between the predicted and true distributions. This sensitivity allows KL to emphasize differences that JS might smooth out.

\textbf{Why use KL Divergence to the nearest vertex?} Measuring KL divergence to the nearest vertex of the credal set avoids giving an unfair advantage to point predictions by deterministic models such as CNN. If we were to use the center of mass of a credal set formed by Bayesian predictions instead, there might be cases where the point prediction by CNN might align perfectly with this center. In such a case, the KL divergence would be identical for both the CNN prediction and the Bayesian prediction. This would unfairly favor the CNN prediction as their non-specificity is always zero, while the non-specificity of the Bayesian prediction depends on its credal set size. Hence, using the centroid would penalise imprecise predictions whose boundaries may be pretty close to the ground truth. By focusing on the nearest vertex, we ensure that the balance between KL and non-specificity is preserved. Now, if we were to consider the distance to farthest point in the credal set, the scoring would instead be on the basis of how wrong the wrongest prediction is. It acts as a dual \citep{troffaes2007decision} to the minimum distance route that we take.

\subsection{Non-Specificity (NS) measure} \label{app:nonspec-theory}

One of the earlier works by \citet{yager2008entropy} makes the distinction between two types of uncertainty within a credal set: conflict (also known as randomness or discord) and non-specificity. Non-specificity essentially varies with the size of the credal set \citep{kolmogorov1965three}. Since by definition, aleatoric uncertainty refers to the inherent randomness or variability in the data, while epistemic uncertainty relates to the lack of knowledge or information about the system \citep{DBLP:journals/corr/abs-1910-09457}, conflict and non-specificity directly parallel these concepts. These measures of uncertainty are axiomatically justified \citep{bronevich2008axioms}.

Over the years, several measures have been proposed to quantify the specificity of belief measures, representing the concentration of mass allocated to focal elements.
Dubios and Prade's non-specificity measure (Eq. \ref{eq:non_spec}) can be considered a generalisation of Hartley’s entropy ($H = \log(|\mathbf{Y}|)$) to belief functions \citep{hartley1928transmission}. When the BBA $m(\cdot)$ is a Bayesian BBA, meaning it only contains singleton focal elements, it reaches the minimum value of 0. When $m(\cdot)$ is a vacuous BBA, denoted by $m(\mathbf{Y}) = 1$, it reaches the maximum value of $\log_2(|\mathbf{Y}|)$.

Another measure based on commonality was proposed by Smets \citep{smets1983information}:
\vspace{-5pt}
\begin{equation}
NS_s[m] = \sum_{A \subseteq \mathbf{Y}} \log \left( \frac{1}{Q(A)} \right)
\vspace{-5pt}
\end{equation}

Korner's non-specificity \citep{korner1995specificity}, denoted as $KN(m)$, is defined as:
\vspace{-5pt}
\begin{equation}
NS_k[m] = \sum_{A \subseteq \mathbf{Y}} m(A) \cdot |A|
\vspace{-5pt}
\end{equation}
The maximum value occurs when the BBA represents a vacuous BBA, denoted by $|\mathbf{Y}|$, while the minimum value is attained for a Bayesian BBA, set to 1. Notably, the computation involves the mass assignments of singletons.

\textbf{Specificity measures. } Pal \citep{pal1993uncertainty} proposed a \emph{specificity} measure to evaluate the dispersion of evidence contributing to a belief function:
\vspace{-5pt}
\begin{equation}
S_p[m] = \sum_{A \in \mathbf{Y}} \frac{m(A)}{|A|}
\vspace{-5pt}
\end{equation}
The maximum value of Pal's specificity is 1 when the mass function is Bayesian, and the minimum value is $\frac{1}{|\mathbf{Y}|}$ when the mass is vacuous.

\subsection{Evaluation Framework or Uncertainty measure?}

The evaluation metric is \emph{not} a new uncertainty measure. The first term, the KL divergence, is unrelated to uncertainty and is more related to accuracy, whereas non-specificity \citep{yager2008entropy} is a measure of epistemic uncertainty as it is directly proportional to the size of the credal set \citep{hullermeier2021aleatoric}. Together, with the trade-off parameter $\lambda$, the evaluation metric provides a \textit{holistic assessment of the predictions}.

The value of the metric $\mathcal{E}$ is a function of the credal set. The distance measure $d(y,\hat{y})$ is the distance to a single one-hot probability vector, which has both 0 epistemic uncertainty (being a single point) and 0 aleatoric uncertainty (as it is one-hot). 
Non-specificity was originally introduced as a measure of imprecision in a random set framework in which masses are independently assigned to sets of outcomes. As discussed in \citep{hullermeier2021aleatoric}, Sec. 4.6.1, in the case of credal sets non-specificity can be considered a measure of \textit{epistemic}, rather than total uncertainty.

Therefore, the first term in the evaluation metric, KL divergence has no relation to uncertainty, while, the second term, non-specificity, can be considered a measure of epistemic uncertainty \citep{DBLP:journals/corr/abs-1910-09457}.

\section{\MakeUppercase{Practical utility of the Evaluation metric}}
\label{app:utility}

\textbf{How a practitioner determines the trade-off $\lambda$.} 

To select a suitable $\lambda$, the practitioner will consider the following task requirements:

\begin{itemize}[itemsep=0pt, topsep=2pt]
    \item Is abstention allowed?
    \item Is precision or accuracy more critical?
    \item What are the risks associated with incorrect predictions or abstention?
\end{itemize}

Here, we use the same two examples of \textbf{crop disease classification} \textit{(abstention from decision-making is allowed)}, and \textbf{autonomous driving} \textit{(abstention is not allowed)}. Consider the following toy problem involving four models (A, B, C, and D) and their respective performances under different $\lambda$ values, as shown in Tab. \ref{tab:model_comparison} below.

\begin{table}[!ht]
    \caption{Comparison of Evaluation Metric ($\mathcal{E}$) for Models A, B, C, and D under different values of $\lambda$.}
    \centering
    \resizebox{\linewidth}{!}{
    \begin{tabular}{@{}lccccc@{}}
        \toprule
        \textbf{Model} & \textbf{KL} & \textbf{NS} & \textbf{$\mathcal{E}$(Low $\lambda = 0.1$)} & \textbf{$\mathcal{E}$(Moderate $\lambda = 0.5$)} & \textbf{$\mathcal{E}$(High $\lambda = 2$)} \\ \midrule
        Model A & 0.243 & 0.166 & $0.243 + 0.1 \times 0.166 = 0.259$ & $0.243 + 0.5 \times 0.166 = 0.326$ & $0.243 + 2 \times 0.166 = 0.575$ \\ 
        Model B & 0.031 & 0.385 & $0.031 + 0.1 \times 0.385 = \textbf{0.069}$ & $0.031 + 0.5 \times 0.385 = \textbf{0.223}$ & $0.031 + 2 \times 0.385 = 0.801$ \\ 
        Model C & 0.002 & 2.267 & $0.002 + 0.1 \times 2.267 = 0.228$ & $0.002 + 0.5 \times 2.267 = 1.136$ & $0.002 + 2 \times 2.267 = 4.536$ \\ 
        Model D & 0.398 & 0.009 & $0.398 + 0.1 \times 0.009 = 0.398$ & $0.398 + 0.5 \times 0.009 = 0.402$ & $0.398 + 2 \times 0.009 = \textbf{0.416}$ \\ 
        \bottomrule
    \end{tabular}}
    \label{tab:model_comparison}
\end{table}

\textbf{Ranking:} The model selection algorithm returns the following ranking of models based on the values in Tab. \ref{tab:model_comparison}.

\begin{itemize}[itemsep=0pt, topsep=2pt]
    \item $\lambda$ = 0.1 (\textit{Low}): \textbf{B}, C, A, D
    \item $\lambda$ = 0.5 (\textit{Moderate}): \textbf{B}, A, D, C
    \item $\lambda$ = 2 (\textit{High}): \textbf{D}, A, B, C
\end{itemize}

\textbf{1. Crop disease classification}

Abstention is \textit{allowed}–can afford to abstain from making predictions for uncertain cases (e.g., sending ambiguous images for manual analysis).

\textbf{Why choose a low $\lambda$?} In this scenario, we have the option to abstain from making decisions for uncertain cases (e.g., we can send ambiguous images for manual analysis based on the model's predictions). Therefore, we do not give much weightage to non-specificity and prefer a model which has low KL, i.e., it can give us a prediction more closer to the ground truth.

Note that our metric is not used for test set evaluation, it is only used for model selection. After the model is selected, we do not scale either of these terms up or down using $\lambda$. The model operates with its inherent uncertainty predictions without any further adjustments.

\textbf{Why not choose a high $\lambda$?} If we choose a high $\lambda$, we penalize models with high uncertainty. As a result, we end up selecting a model that is very precise in its predictions. This prevents us from leveraging the advantage of the abstention allowance.

\textbf{2. Autonomous driving}

Abstention is \textit{not allowed}–decision making is necessary, as abstention can lead to critical failures, such as accidents, harm to pedestrians, or collisions with other vehicles.

\textbf{Why choose a high $\lambda$?} In this scenario, a decision must be made (e.g., when there is a shadow on the road, we cannot afford to be uncertain or abstain; we must decide whether to stop or not). Therefore, we heavily penalize non-specificity. Models with lower non-specificity will rank first using our model selection, because they are more decisive in their predictions.

\textbf{Why not choose a low $\lambda$?} If we choose a low $\lambda$, we prioritize the models with high uncertainty, therefore, end up selecting a model that is very imprecise about its predictions. When there is no option to abstain, we do not want an indecisive model.

\textbf{How does it reflect in the rankings?
}
In the rankings for low $\lambda$ (\textbf{$\lambda$ = 0.1}), Model \textbf{B} ranks first due to its low KL. Conversely, with high $\lambda$ (\textbf{$\lambda$ = 2}), the ranking shifts to prioritize models with low non-specificity. Model \textbf{D} ranks highest because of its minimal non-specificity. This ranking penalizes uncertain models like Models C and B, pushing it to the bottom.

For crop disease classification, choosing a high $\lambda$ in this scenario negates the advantage of abstention and leads to selecting a model that is overly precise at the cost of being potentially less reliable in uncertain cases. However, in autonomous driving, abstention is not an option because hesitation in decision-making can result in accidents. Here, a high $\lambda$ is crucial to prioritize models with low non-specificity, ensuring the model makes more precise and decisive predictions.

\section{\MakeUppercase{Uncertainty Estimation}}
\label{app:A}

\subsection{Bayesian Neural Networks}
\label{app:A_BNN}

Bayesian inference integrates over the posterior distribution $p(\theta \mid \mathbb{D})$
over model parameters $\theta$ given training data $\mathbb{D}$ to compute the predictive distribution $\hat{p}_{b}(y \mid \mathbf{x}, \mathbb{D})$, reflecting updated beliefs after observing the data:
\begin{equation}
    \hat{p}_{b}(y \mid \mathbf{x}, \mathbb{D}) = \int p(y \mid \mathbf{x}, \theta) p(\theta \mid \mathbb{D}) d\theta,
\end{equation}
where $p(y \mid \mathbf{x}, \theta)$ represents the likelihood function of observing label $y$ given $x$ and $\theta$. 
To overcome the infeasibility of this integral, direct sampling from $\hat{p}_{b}(y \mid \mathbf{x}, \mathbb{D})$ using methods such as Monte-Carlo \citep{hastings1970monte} are applied to obtain a large set of sample weight vectors, $\{ \theta_k, k\}$, from the posterior distribution. These sample weight vectors are then used to compute a set of possible outputs $y_k$,
namely: 
\begin{equation}
    \hat{p}_{b}(y_k\mid \mathbf{x}, \mathbb{D}) = \frac{1}{|\Theta|} \sum_{\theta_k \in \Theta} \Phi_{\theta_k}(\mathbf{x}),
\end{equation}
where $\Theta$ is the set of sampled weights, $\Phi_{\theta_k}(\mathbf{x})$ is the prediction made by the model with weights $\theta_k$ for input $\mathbf{x}$, and $\Phi$ is the function for the model. This process is called \emph{Bayesian Model Averaging (BMA)}.
Fig. \ref{fig:BMA_vs_nonBMA} illustrates two contrasting scenarios in which BMA proves advantageous in the first case (top), yet exhibits limitations as it discards all information in the second case (bottom). This can limit a BNN's ability to accurately represent complex uncertainty patterns, potentially undermining its effectiveness in scenarios requiring reliable uncertainty quantification.
BMA may inadvertently smooth out predictive distributions, diluting the inherent uncertainty present in individual models \citep{hinne2020conceptual, graefe2015limitations} as shown in Fig. \ref{fig:BMA_vs_nonBMA}. 
When applied to classification, BMA yields
point-wise predictions. For fair comparison and to overcome BMA's limitations, 
in this paper we also use sets of prediction samples obtained from the different posterior weights before averaging.

\begin{figure}[!h]
    \centering
    \begin{minipage}[t]{0.49\textwidth}
        \centering
        \includegraphics[width=\textwidth]{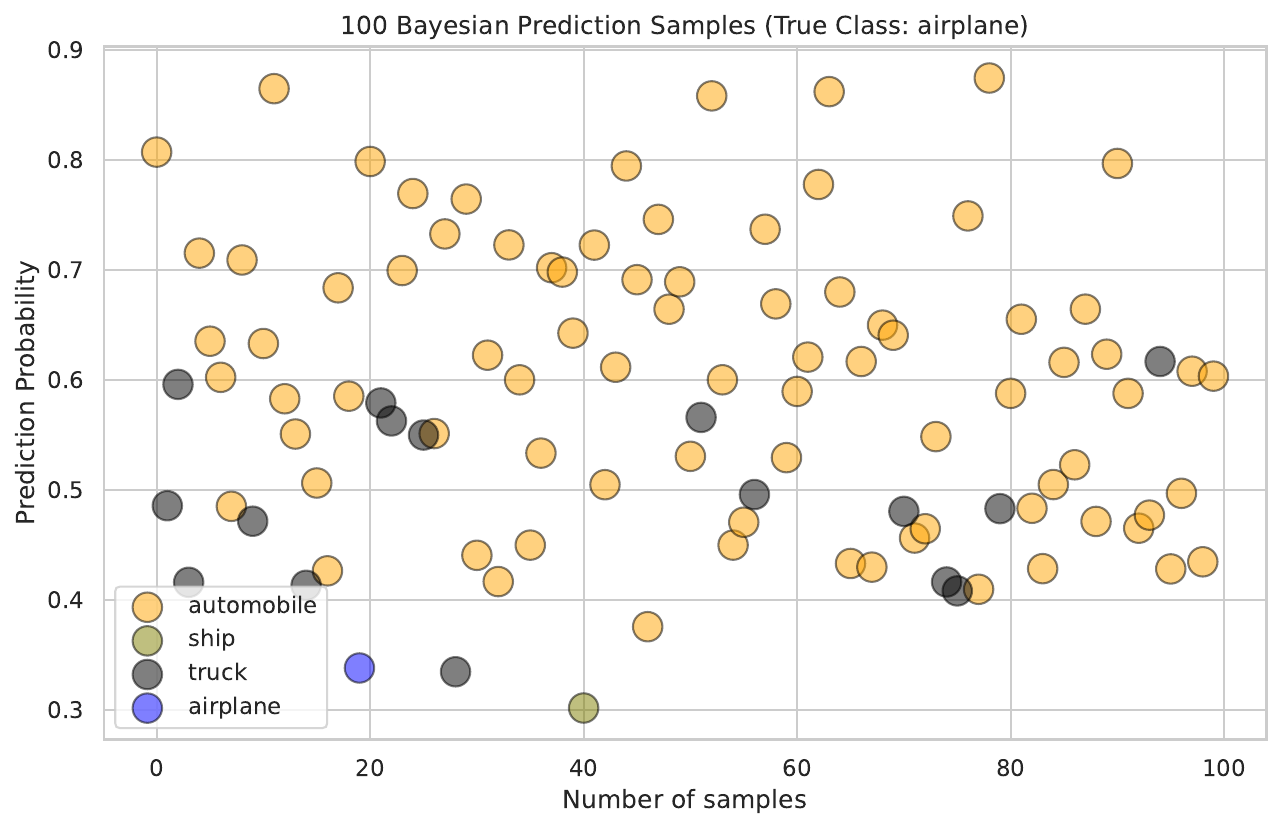}
    \end{minipage}\hspace{0.2cm}
    \begin{minipage}[t]{0.49\textwidth}
        \centering
        \includegraphics[width=\textwidth]{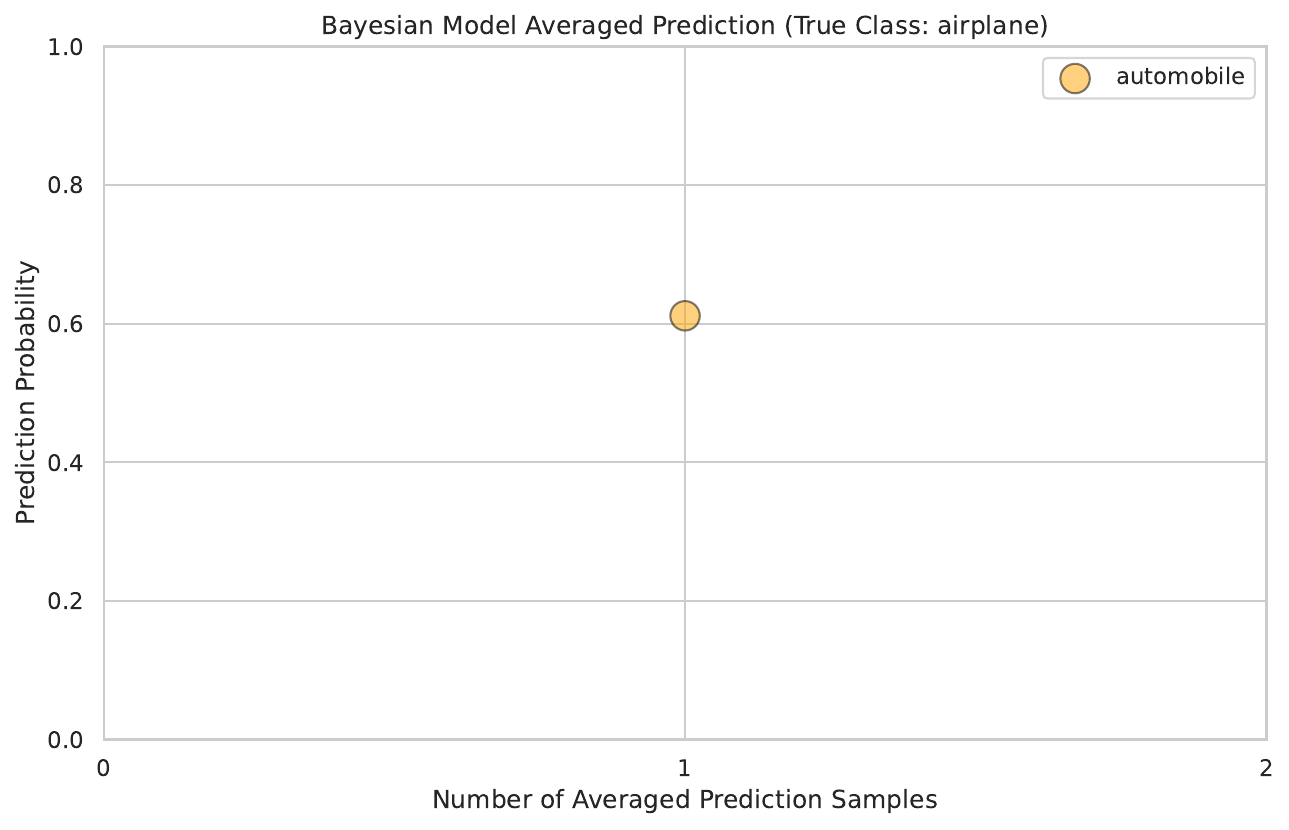}
    \end{minipage}\hfill
    \begin{minipage}[t]{0.49\textwidth}
        \centering
        \includegraphics[width=\textwidth]{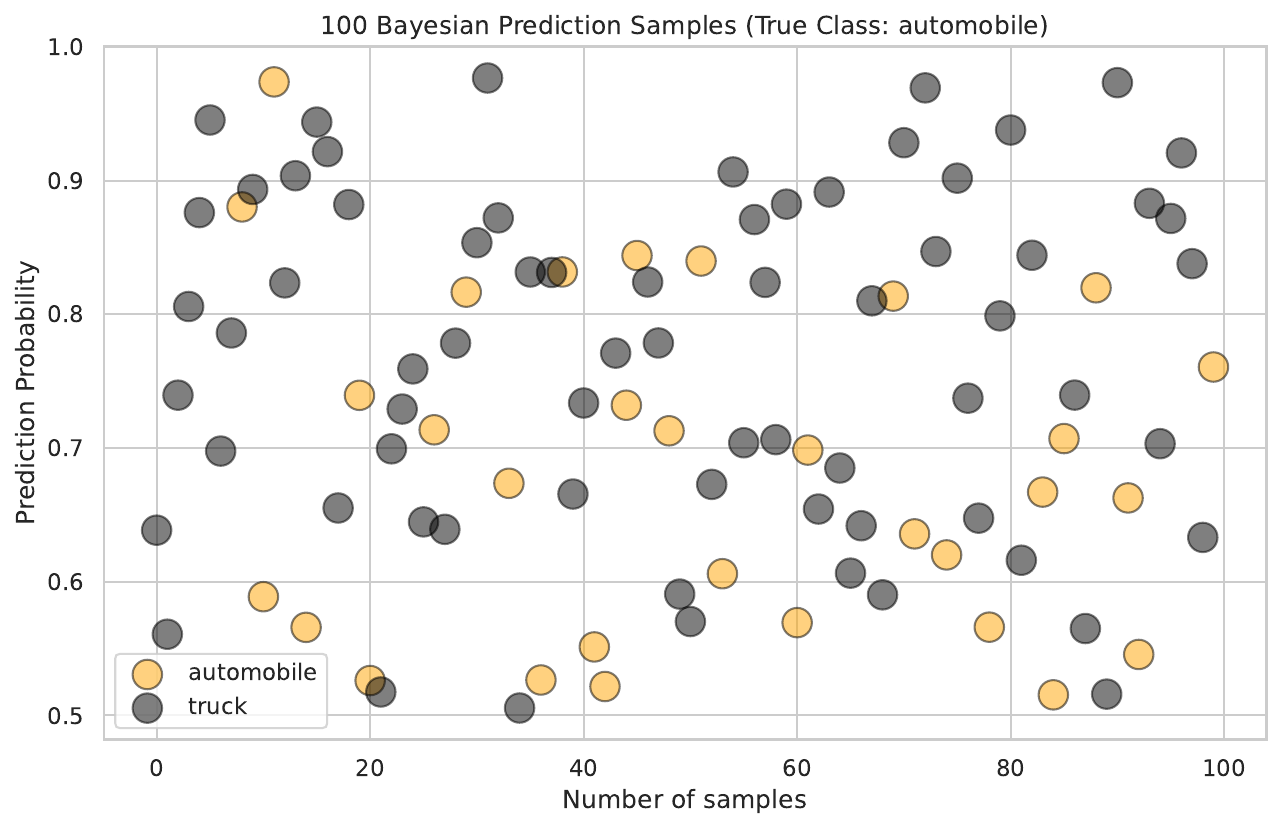}
    \end{minipage}\hspace{0.2cm}
    \begin{minipage}[t]{0.49\textwidth}
        \centering
        \includegraphics[width=\textwidth]{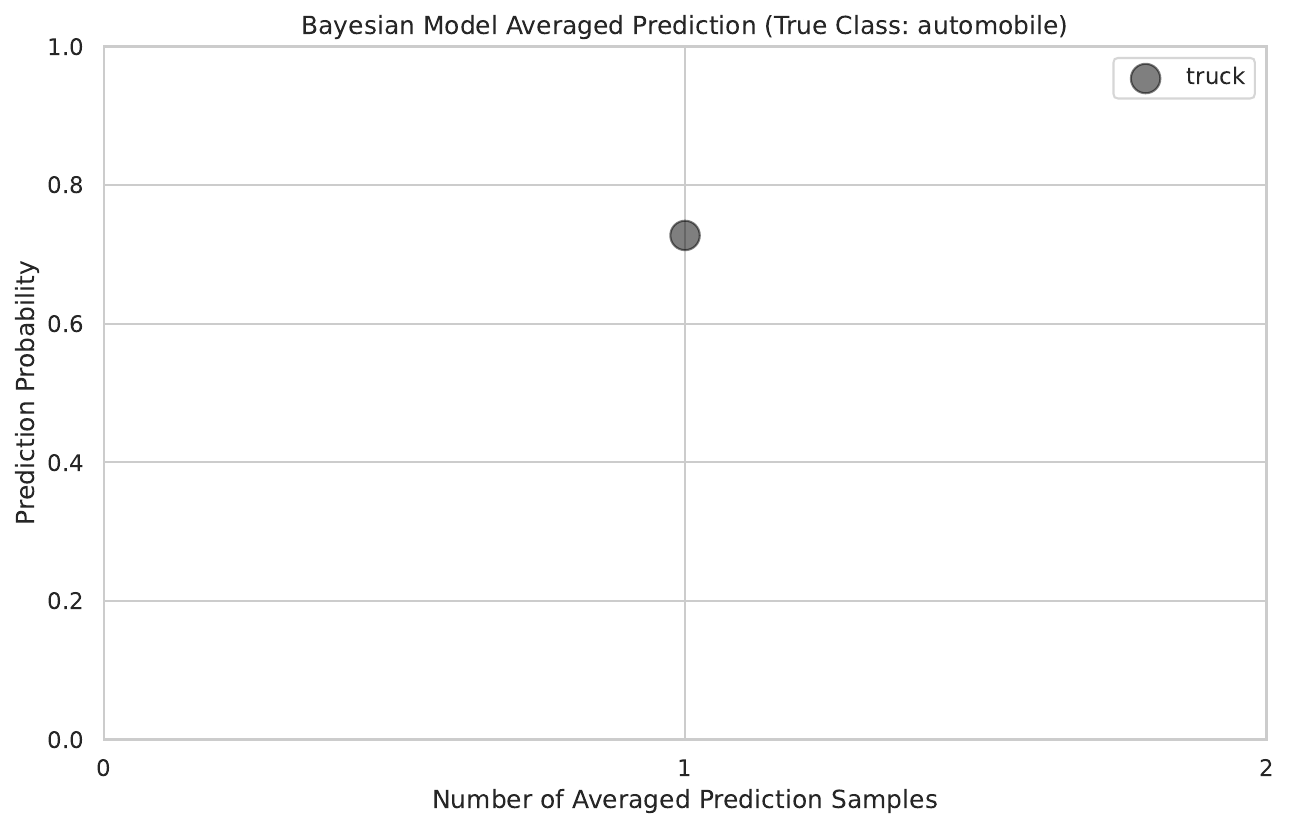}
    \end{minipage}
    \caption{Visualizations of 100 prediction samples obtained prior to Bayesian Model Averaging and corresponding Bayesian Model Averaged prediction in two real scenarios from CIFAR-10.}
    \label{fig:BMA_vs_nonBMA}
\end{figure}

In BNNs,
\emph{aleatoric uncertainty} is measured by the predictive entropy, while \emph{epistemic uncertainty} is represented by \emph{mutual information} \citep{hullermeier2021aleatoric, hullermeier2022quantification}, MI, which measures the difference between the entropy of the predictive distribution and the expected entropy of the individual predictions.
To compute both mutual information 
and predictive entropy 
in Bayesian Neural Networks (BNNs), one utilises the predictive distributions of the model. 
MI quantifies the amount of information gained about the label $y$ given the input $\mathbf{x}$ and the observed data $\mathbb{D}$, while the predictive entropy ($H$) measures the uncertainty associated with the predictions:
\begin{equation}
    \text{MI}(\hat{p}_{b}(y \mid \mathbf{x}, \mathbb{D})) = H(\hat{p}_{b}(y \mid \mathbf{x}, \mathbb{D})) - \mathbb{E}_{\mathbb{D}}[H(p(y \mid \mathbf{x},{\theta}))] ,
\end{equation}
where $H(\hat{p}_{b}(y \mid \mathbf{x}, \mathbb{D}))$ is the entropy of the predictive distribution obtained from BMA, and $\mathbb{E}_{\mathbb{D}}[H(p(y \mid \mathbf{x},{\theta}))]$ represents the expected entropy of the individual predictive distributions sampled from the posterior distribution of the parameters $p(\theta \mid \mathbb{D})$. $H(\cdot)$ denotes the Shannon entropy function. 

The predictive entropy can be calculated as: 
\begin{equation}
H(\hat{p}_{b}(y \mid \mathbf{x}, \mathbb{D})) = - \int \hat{p}_{b}(y \mid \mathbf{x}, \mathbb{D}) \log \hat{p}_{b}(y \mid \mathbf{x}, \mathbb{D}) dy ,
\end{equation}
where $\hat{p}(y \mid \mathbf{x}, \mathbb{D})$ is the predictive distribution. This equation represents the average uncertainty associated with the predictions across different possible values of $y$, considering the variability introduced by the parameter uncertainty captured in the posterior distribution $p(\theta \mid \mathbb{D})$.

\subsection{Deep Ensembles}
\label{app:A_DE}

In Deep Ensembles,
{aleatoric uncertainty} is assessed via the predictive entropy, averaged entropy of each ensemble's prediction, while {epistemic uncertainty} is encoded by the predictive variance, 
the difference between the entropy of all ensembles and the averaged entropy of each ensemble.

Let $\mathcal{M} = \{ M_1, M_2, \ldots, M_K \}$ denote the ensemble of $K$ neural network models for $k = 1, 2, \ldots, K$. Given an input $\mathbf{x}$, the prediction $y_{\mathcal{M}}$ is obtained by averaging the predictions of individual models.
The \emph{predictive entropy} 
represents the averaged entropy of each ensemble's prediction $y_{k}$ given the input 
$\mathbf{x}$ and the observed data $\mathbb{D}$:
\begin{equation}
H(\hat{p}_{de}(y \mid \mathbf{x}, \mathbb{D})) = \frac{1}{K} \sum_{k=1}^{K} H(\hat{p}_{de}(y_{k} \mid \mathbf{x}, \mathbb{D})),
\end{equation}
where $y_{k}$ represents the prediction of the $k$-th model $M_k$.

The \emph{predictive variance} 
is measured as the difference between the entropy of all the ensembles, $H(\hat{p}_{de}(y_{\mathcal{M}} \mid \mathbf{x}, \mathbb{D}))$, and the averaged entropy of each ensemble, $H(\hat{p}_{de}(y \mid \mathbf{x}, \mathbb{D}))$.
\begin{equation}
H(y_{\mathcal{M}}) = H(\hat{p}_{de}(y_{\mathcal{M}} \mid \mathbf{x}, \mathbb{D})) - H(\hat{p}_{de}(y \mid \mathbf{x}, \mathbb{D})).
\end{equation}

The predictive variance in DEs is considered an approximation of mutual information \citep{hullermeier2021aleatoric}. This formulation captures both the model uncertainty inherent in the ensemble predictions and the uncertainty due to the variance among individual model predictions. 

\begin{figure}[!ht]
    \centering
    \includegraphics[width=0.9\textwidth]{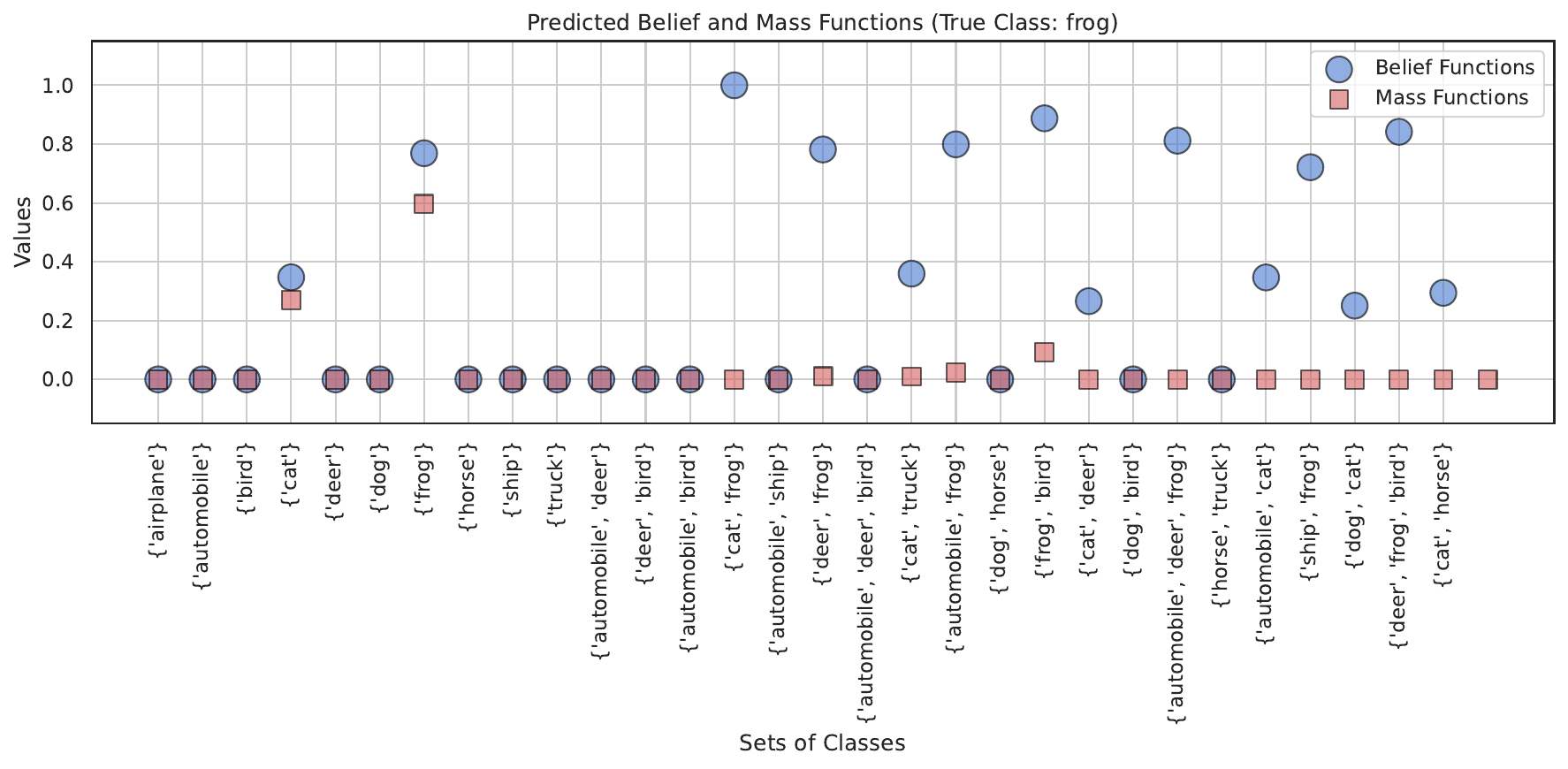}
    \centering
    \includegraphics[width=0.44\textwidth]{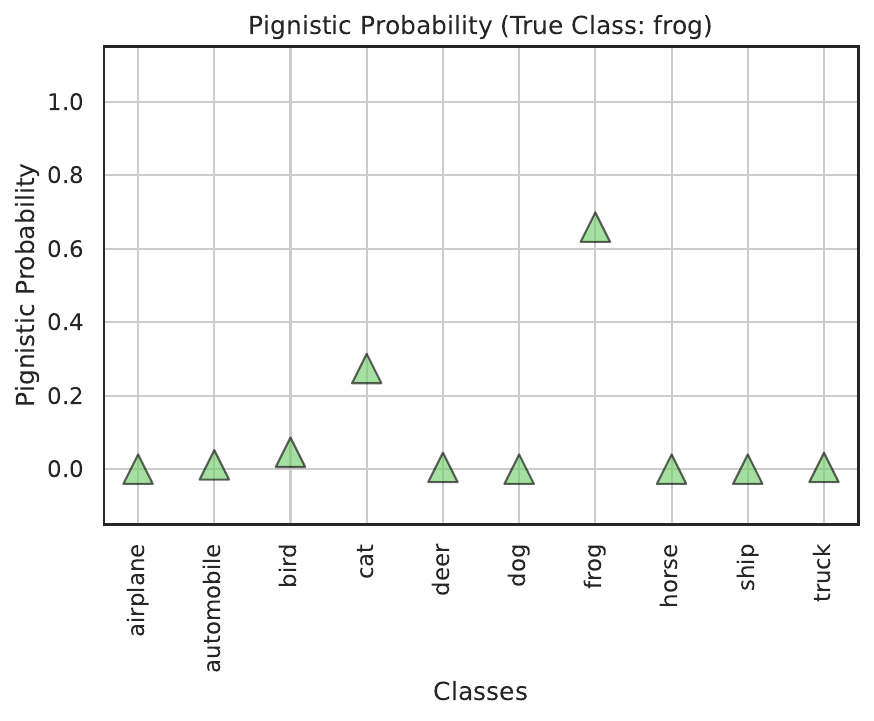}
    \caption{Visualizations of belief and mass predictions on the power-set space and its mapping to the label space $\mathbf{Y}$ using pignistic probabilities on the CIFAR-10 dataset. 
    }
    \label{fig:RS_SNN_pred}
\end{figure}

\subsection{Belief Function models}
\label{app:A_BF}

Belief functions can be derived from \emph{mass functions} through a normalization process, where the belief assigned to a hypothesis is the sum of the masses of all subsets of the frame of discernment that include the hypothesis.
A {mass function} 
\citep{Shafer76} is a set function \citep{denneberg99interaction} $m : 2^\Theta\rightarrow[0,1]$ such that $m(\emptyset)=0$ and
$\sum_{A\subset\Theta} m(A)=1$. In classification, $2^\Theta$ is the set of all subsets of classes $\mathcal{C}$, 
the powerset $\mathbb{P}(\mathcal{C})$. 
Subsets of $\Theta = \mathcal{C}$ whose mass values are non-zero are called \emph{focal elements} of $m$.
The \emph{belief function} associated with 
$m$ 
is given by:
$
Bel(A) = \sum_{B\subseteq A} m(B). 
$
The redistribution of mass values back to singletons from focal sets is achieved through the concept of \emph{pignistic probability} \citep{SMETS2005133}. Pignistic probability ($BetP$), also known as Smets' pignistic transform, is a method used to assign precise probability values to individual events based on the belief function's output. 

\textit{Aleatoric uncertainty} in such models is represented as the pignistic entropy of predictions $H_{BetP}$, whereas \textit{epistemic uncertainty} can be modelled by the `size' of the credal set (Eq. \ref{eq:consistent}). 

Fig. \ref{fig:RS_SNN_pred} shows belief function predictions for an input sample with the true class `\verb+frog+'. The belief function predictions and their mapping to mass functions and pignistic probabilities are illustrated.

\section{\MakeUppercase{Computing the credal set of predictions}}
\label{app:approx-credal}

We do not require the computation of all the vertices of a credal set. In this section, we provide a very efficient approximation to compute the credal set vertices that are most relevant, i.e, the outermost vertices of the credal set. With this approximation, the method can be applied to datasets with larger number of classes.

\textbf{Approximate computation}. 
Given the prohibitively high computational complexity associated with computing permutations as in (Eq. \ref{eq:prho}), especially for a large number of classes, we propose an approximation technique that offers comparable effectiveness without the need to compute all vertices.
For each class $c_i$, $i = 1, 2, \ldots, N$ in classes $\mathcal{C}$ where $N$ is the number of classes, we create two permutations. In the first permutation, $c_i$ is placed as the first element, while in the second permutation, $c_i$ is placed as the last element. By doing so, $c_i$ occupies both the first and last positions in the permutations. Due to the nature of extremal probabilities calculation (Eq. \ref{eq:prho}), when a class $c_i$ is placed as the first element in the permutation, it occupies the initial position in the subset, allowing it to be included in the maximal number of focal elements. As a result, the extremal probability for $c_i$ in this scenario is maximized because it contributes to the sum of masses for all the focal elements containing it. 

Conversely, when a class $c_i$ is placed as the last element in the permutation, 
it is excluded from the focal elements containing classes preceding it in the permutation order. Therefore, the extremal probability for $c_i$ is minimised, because it contributes to the sum of masses for fewer focal elements compared to when it is placed as the first element. This approach yields $2 \cdot N$ number of vertices, 
providing an effective approximate approach to computing all vertices.

The vertices of the credal set associated with each model are computed using mass functions as in Eq. \ref{eq:prho}, via the approximation technique detailed in Sec. \ref{sec:credal_set}. 

\textbf{The effect of approximating credal set vertices:} To demonstrate that the approximated vertices of the credal sets are sufficient (and often, even better), we consider a 4-class example of the CIFAR-10 dataset. In Tab. \ref{tab:approx_naive}, we show the evaluation metric $\mathcal{E}$ for computed using approximated credal vertices using the approach detailed above, and the naive credal set vertices computation given in Eq. \ref{eq:prho}. For the naive credal set, we use the complete 24 (4!) vertices and for the approximated credal set, we use a reduced number of 8 vertices. The values in bold are the clearly better results whereas other values are quite close to each other. For DEs, for instance, the approximated credal set is a better representation for our metric, as the approximation uses the most relevant permutations only.

\begin{table}[!ht]
\caption{Comparison of KL, Non-Specificity, Evaluation Metric ($\mathcal{E}$) calculated using approximated versus naive credal set vertices for LB-BNN and DE on the CIFAR-10 dataset.} 
\label{tab:approx_naive}
\centering
\resizebox{\textwidth}{!}{
\begin{tabular}{lccccccc}
\toprule
\multirow{2}{*}{Model} &
  \multirow{2}{*}{Credal Set Vertices} &
  \multicolumn{2}{c}{KL distance (KL)} &
  \multicolumn{2}{c}{Non-Specificity (NS)} &
  \multicolumn{2}{c}{Evaluation metric ($\mathcal{E}$)} \\
\cmidrule(lr){3-4} \cmidrule(lr){5-6} \cmidrule(lr){7-8}
&& CC ($\downarrow$) & ICC ($\downarrow$)  & CC ($\downarrow$) & ICC ($\uparrow$) & CC ($\downarrow$) & ICC ($\downarrow$)\\
\midrule
\multirow{2}{*}{LB-BNN} 
& Approximated & $0.0046 \pm 0.023$ & $0.612 \pm 0.3968$ & $0.564 \pm 0.099$ & $\mathbf{0.611} \pm \mathbf{0.140}$ & $0.569 \pm 0.104$ & $1.225 \pm 0.376$\\
& Naive & $0.005 \pm 0.023$ & $0.336 \pm 0.666$ & $\mathbf{0.033} \pm \mathbf{0.103}$ & $0.286 \pm 0.216$ & $\mathbf{0.039} \pm \mathbf{0.121}$ & $\mathbf{0.365} \pm \mathbf{0.594}$\\
\midrule
\multirow{2}{*}{DE} 
& Approximated & $0.0002 \pm 0.002$ & $\mathbf{0.142} \pm \mathbf{0.260}$ & $\mathbf{0.058} \pm \mathbf{0.125}$ & $\mathbf{0.801} \pm \mathbf{0.170}$ & $\mathbf{0.058} \pm \mathbf{0.125}$ & $\mathbf{0.944} \pm \mathbf{0.265}$\\
& Naive & $0.0001 \pm 0.002$ & $0.335 \pm 0.807$ & $0.080 \pm 0.204$ & $0.656 \pm 0.219$ & $0.080 \pm 0.205$ & $0.991 \pm 0.653$\\
\bottomrule
\end{tabular}
} 
\end{table}

\begin{figure}[!ht]
    \centering
    \begin{minipage}{0.33\textwidth}
        \centering
        \includegraphics[width=\textwidth]{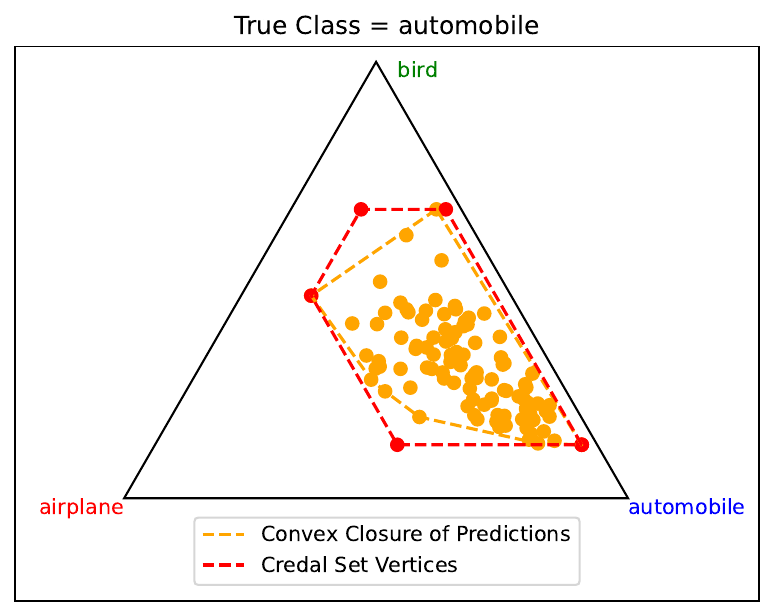}
    \end{minipage}%
    \begin{minipage}{0.33\textwidth}
        \centering
        \includegraphics[width=\textwidth]{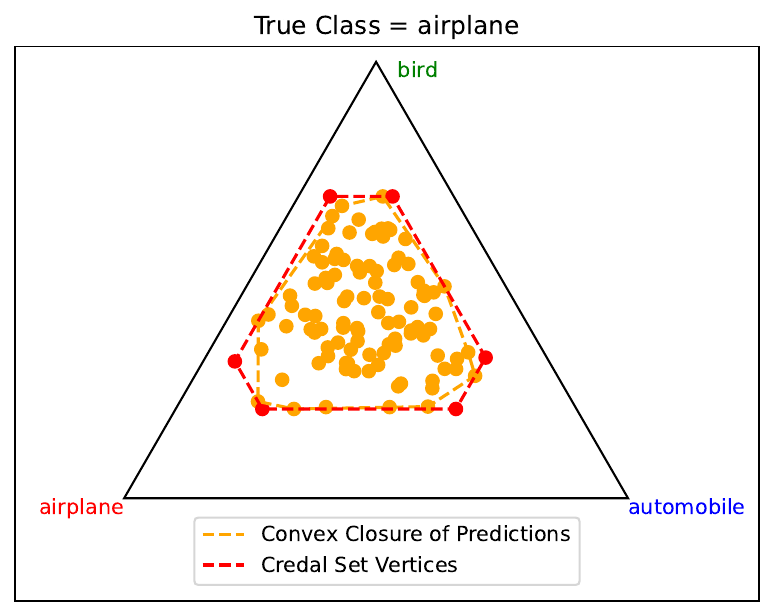}
    \end{minipage}%
    \begin{minipage}{0.33\textwidth}
        \centering
        \includegraphics[width=\textwidth]{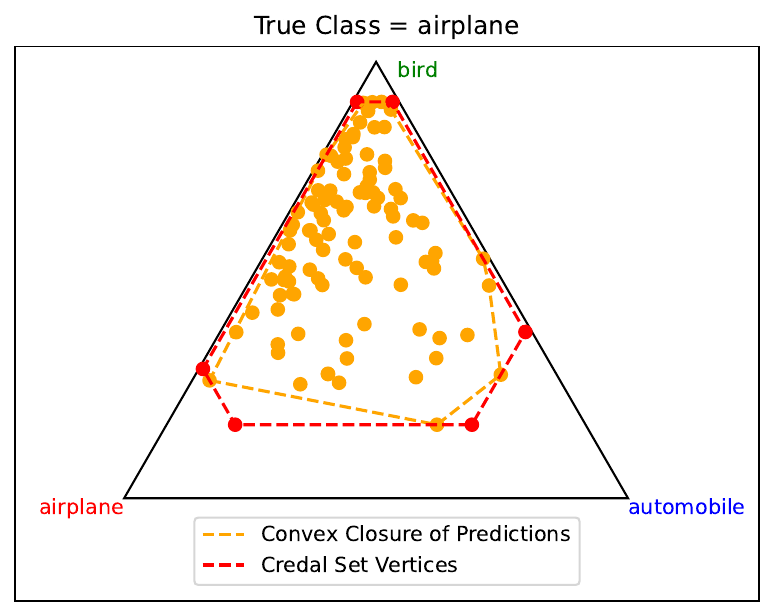} 
    \end{minipage}
    \caption{Probability simplices illustrating the convex closure of predictions and credal sets for the Bayesian model (LB-BNN) across three classes of the CIFAR-10 dataset. }
        \label{fig:convex-closure-vs-credal}
\end{figure}

\textbf{Credal set vs convex closure of predictions.} Since our metric uses the KL divergence between the ground truth and the closest vertex of the credal set (which is convex), the credal set representation does not affect its value as such vertices correspond to (some of) the original (e.g. Bayesian) predictions, as shown in Fig. \ref{fig:convex-closure-vs-credal}. Probability simplices illustrating the convex closure of predictions and credal sets for LB-BNN (Bayesian) model are shown in Fig. \ref{fig:convex-closure-vs-credal}, over three classes of CIFAR-10. The convex closure of predictions often coincides with the vertices of the credal set, indicating that some Bayesian predictions are situated exactly at the extremities of the credal set.
There is a slight difference between the convex convex closure of predictions and the credal set, which is not due to vertex approximations. \citep{cozman2000credal} details how there is more than a single way to specify a set of distributions inducing a given lower envelope. The credal set in Fig. \ref{fig:convex-closure-vs-credal} has been approximated using coherent lower probabilities and have boundries parallel to that of the simplex.

\section{\MakeUppercase{Experimental details}} \label{sec:details}

\subsection{Baselines}

\textbf{LB-BNN}\citep{hobbhahn2022fast} is a Bayesian approximation model which uses the Laplace Bridge to 
efficiently link
Gaussian and Dirichlet distributions. In LB-BNN, the Laplace Bridge combined with a last layer Laplace approximation for Bayesian Neural Networks is used to compute the output Dirichlet parameters. 
\textbf{DEs} \citep{lakshminarayanan2017simple} are an ensemble-based method for uncertainty estimation where multiple neural networks (each initialised with different random seeds) are trained independently on the same task. At test time, predictions from these diverse models are aggregated and averaged to obtain a point estimates for ensemble predictions. 
\textbf{EDL} \citep{sensoy} predictions are obtained by sampling from the posterior Dirichlet distribution, and \textbf{DDU} \citep{mukhoti2021deep} predictions are softmax probabilities over classes, similar to \textbf{SNN}.
\textbf{CreINNs} \citep{wang2024creinns} retain the structure of traditional Interval Neural Networks to generate interval probabilities, lower and upper probabilities for each prediction, and formulate credal sets from these interval predictions. \textbf{E-CNN} is an evidential uncertainty classifier which predicts mass functions for sets of outcomes. Credal sets can directly be generated from these mass functions. \textbf{RS-NN} \citep{manchingal2025randomset} predicts belief functions and their corresponding mass functions for sets of outcomes, forming a convex set of probability distributions, associated with a credal set.

\subsection{Model architecture and training}
\label{sec:implementation-details}

All models are trained on ResNet50 (using 3 NVIDIA A100 GPUs) with a learning rate scheduler, starting at 1e-3 and decreasing by 0.1 at epochs 80, 120, 160, and 180. Standard data augmentation \citep{NIPS2012_c399862d}, including random horizontal/vertical shifts with a magnitude of 0.1 and horizontal flips, is applied to all models. The image size used for all models is $224 \times 224$, except for CreINNs, which uses $32 \times 32$. The optimizer used for all models is Adam \citep{kingma2014adam}. The data is split into 40000:10000:10000 samples for training, testing, and validation respectively for CIFAR-10 and CIFAR-100, 50000:10000:10000 samples for MNIST. We only require a CPU for evaluation of predictions on CIFAR-10 and MNIST. For CIFAR-100, we use an Nvidia A100 40GB GPU. The entire code runs in approximately 70 seconds for CIFAR-10 dataset. The credal uncertainty estimation in ablation study (\S{\ref{app:ablation-ns}}) takes an additional 90 seconds.

\section{\MakeUppercase{Additional experiments}}
\label{app:additional_exp_main}

\subsection{Ablation study on different distance measures} \label{app:ablation-kl}

In this section, we present an ablation study comparing different distance measures, specifically Kullback-Leibler (KL) divergence and Jensen-Shannon (JS) \citep{menendez1997jensen} divergence. The goal of this analysis is to assess how these divergence measures influence the evaluation of uncertainty in our framework. While KL divergence has been chosen for its superior ability, Jensen-Shannon divergence offers a symmetric alternative that handles small probabilities more gracefully. By experimenting with both measures, we provide insights into their respective advantages and determine the most suitable metric for capturing distributional differences in uncertainty-aware model predictions.

\begin{figure}[!ht]
    \begin{minipage}[t]{0.48\textwidth}
    \includegraphics[width=\textwidth]{images/CIFAR10/kl_cifar10.pdf}
    \caption*{(a)}
    \end{minipage} \hspace{0.02\textwidth}
    \begin{minipage}[t]{0.48\textwidth}
    \includegraphics[width=\textwidth]{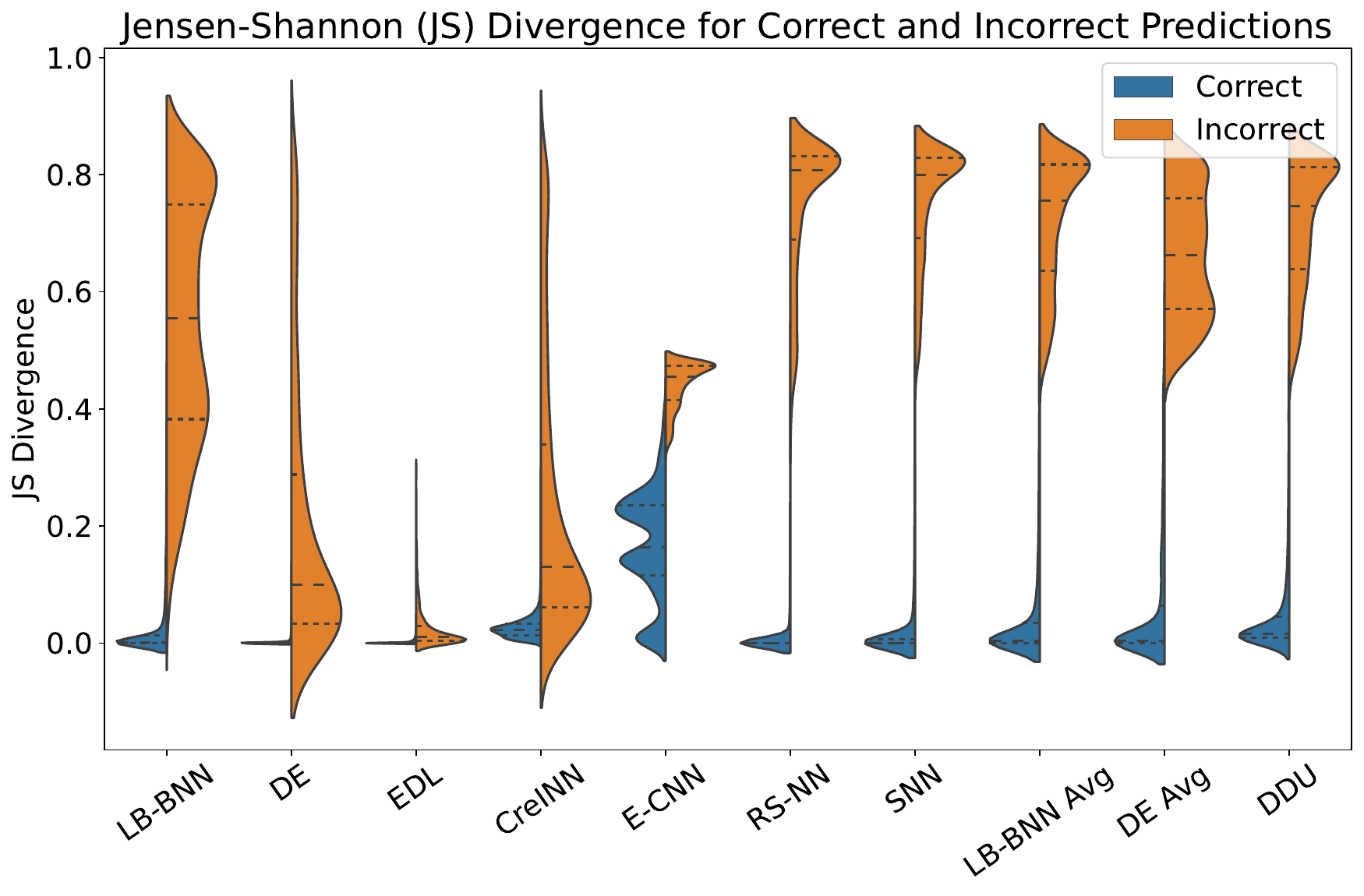}
    \caption*{(b)}
    \end{minipage}
    \caption{Comparison of (a) Kullback-Leibler (KL) divergence, and (b) Jensen-Shannon (JS) divergence for Correctly Classified (CC) and Incorrectly Classified (ICC) samples from the CIFAR-10 dataset, for all models considered here. Notably, the scales of these two measures differ significantly (Y-axis).
    }\label{fig:kl-js}
\end{figure}

Fig. \ref{fig:kl-js} illustrates the Kullback-Leibler (KL) divergence (\ref{fig:kl-js}a) and Jensen-Shannon (JS) divergence (\ref{fig:kl-js}b) between the ground truth probabilities and the vertices of the credal set on the CIFAR-10 dataset, over the entire test set. The violin plots depict distributions of the divergences for correct (blue) and incorrect (orange) predictions. Ideally, we expect the KL divergence to be larger for incorrect predictions, but not to an extent that it approaches a completely different class on the simplex of classes. This outcome indicates a preference for models that do not make severe misclassifications, which would be reflected in excessively predicting a wrong class.
The plots reveal how the KL divergence for incorrect predictions often spans a wider range, indicating greater uncertainty in these predictions compared to correct ones. 
The JS divergence is presented on a smaller scale compared to the KL divergence. 

\begin{figure}[!h]
    \includegraphics[width=\textwidth]{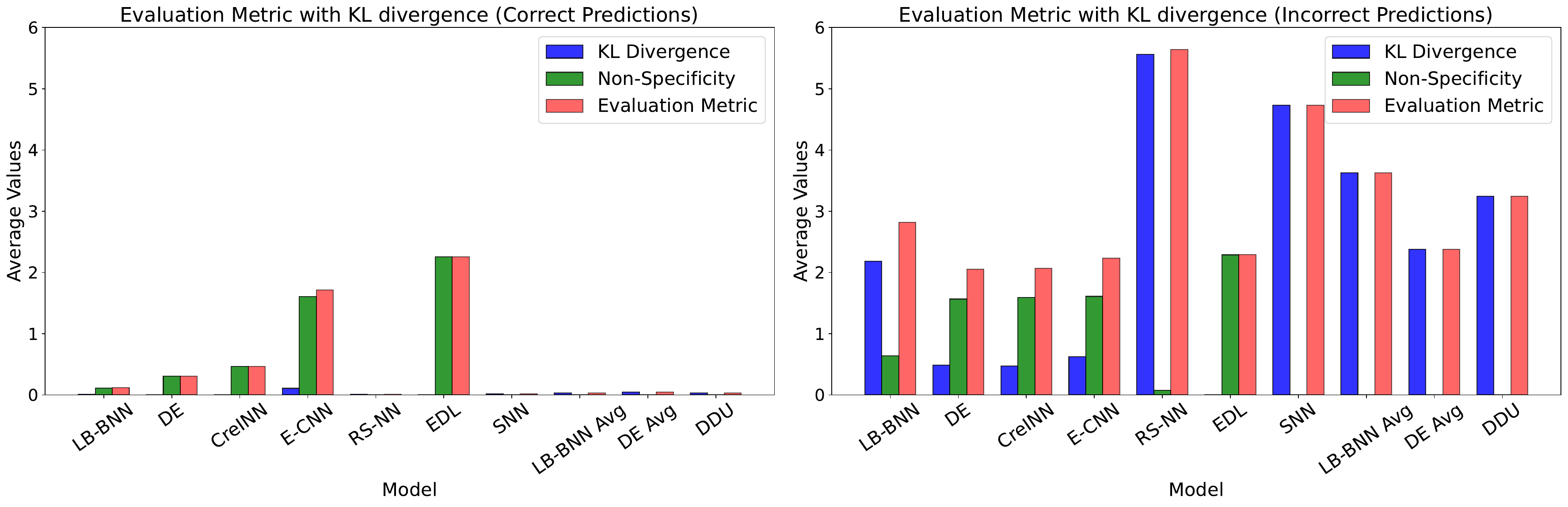}
    \includegraphics[width=\textwidth]{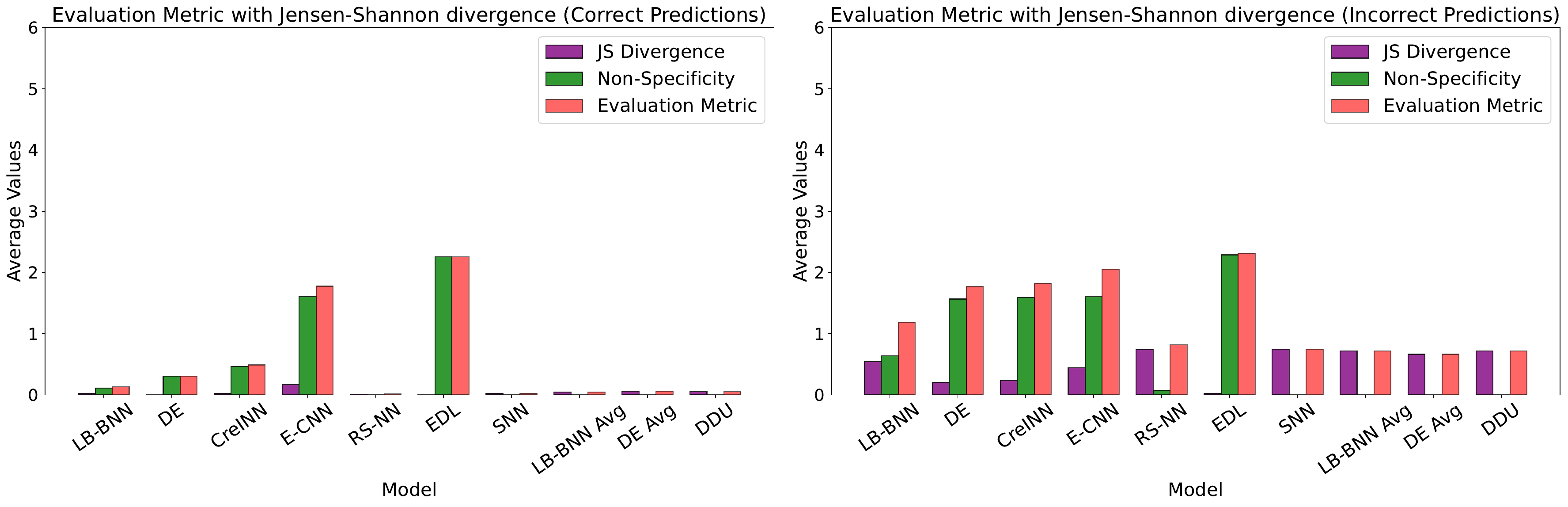}
    \caption{Comparison of mean Evaluation Metric $\mathcal{E}$ using mean \textit{Kullback Leibler} (KL) divergence (top) and mean \textit{Jensen-Shannon} (JS) divergence (bottom) for Correct (left) and Incorrect (right) predictions of the CIFAR-10 dataset.
    }\label{fig:kl-js-eval}
\end{figure}

Fig. \ref{fig:kl-js-eval} shows bar plots comparing the KL and JS divergences along with their influence on the overall evaluation metric and therefore, corresponding model rankings. All values are shown as means. Again, the differing ranges of KL and JS divergences are evident, with the Y-axis scaled equally for straightforward comparison. This discrepancy affects the overall evaluation metric $\mathcal{E}$. According to Algorithm \ref{alg:model-selection} for model selection, the model with the smallest evaluation metric should be chosen.

\textbf{Model Selection.} In Tab. \ref{tab:model_rank_ablation}, we show the ranking of all models (including the models that predict point estimates) arranged in ascending order of the mean of the evaluation metric, $\mathcal{E}$. 
As $\lambda$ increases, DE, which consistently ranked highly in KL divergence across lower $\lambda$ values, is replaced by point prediction model, DE Avg (since it has zero non-specificity). LB-BNN and RS-NN show mixed performance, typically appearing in the top half but with varying positions.

In terms of JS divergence, RS-NN stands out as a top performer, frequently occupying the top spots across most $\lambda$ values, while SNN also maintains a strong position, often ranking within the top three or four models. E-CNN and EDL tend to rank lower for both KL and JS measures. If we exclude the point prediction models from the ranking, both KL divergence and JS divergence consistently identify \textbf{RS-NN} and \textbf{DE} as the top two models in the model selection procedure.

\begin{table}[!ht]
    \centering
    \caption{Model Rankings Based on KL and JS Divergence on the CIFAR-10 dataset. Model selection is based on the mean of Evaluation Metric ($\mathcal{E}$) with the models with the lowest $\mathcal{E}$ ranking first.}
    \label{tab:model_rank_ablation}
    \resizebox{\textwidth}{!}{
    \begin{tabular}{@{}cc|l|l@{}}
        \toprule
        \textbf{Lambda} & \textbf{Metric} & \textbf{Model Ranking} & \textbf{Evaluation Metric ($\mathcal{E}$) Mean} \\ \midrule
        0.1 & KL & DE, CreINN, DE Avg, EDL, LB-BNN, DDU, E-CNN, RS-NN, LB-BNN Avg, SNN & [0.069, 0.117, 0.195, 0.229, 0.259, 0.309, 0.354, 0.399, 0.420, 0.481] \\ 
             & JS & DE, RS-NN, LB-BNN, SNN, DE Avg, DDU, CreINN, LB-BNN Avg, EDL, E-CNN & [0.053, 0.065, 0.095, 0.098, 0.099, 0.109, 0.109, 0.119, 0.238, 0.373] \\ \midrule
        0.2 & KL & DE, CreINN, DE Avg, LB-BNN, DDU, RS-NN, LB-BNN Avg, EDL, SNN, E-CNN & [0.108, 0.177, 0.195, 0.276, 0.309, 0.400, 0.420, 0.456, 0.481, 0.515] \\ 
             & JS & RS-NN, DE, SNN, DE Avg, DDU, LB-BNN, LB-BNN Avg, CreINN, EDL, E-CNN & [0.066, 0.091, 0.098, 0.099, 0.109, 0.111, 0.119, 0.169, 0.464, 0.534] \\ \midrule
        0.3 & KL & DE, DE Avg, CreINN, LB-BNN, DDU, RS-NN, LB-BNN Avg, SNN, E-CNN, EDL & [0.146, 0.195, 0.237, 0.293, 0.309, 0.401, 0.420, 0.481, 0.676, 0.682] \\ 
             & JS & RS-NN, SNN, DE Avg, DDU, LB-BNN Avg, LB-BNN, DE, CreINN, EDL, E-CNN & [0.067, 0.098, 0.099, 0.109, 0.119, 0.128, 0.130, 0.229, 0.691, 0.695] \\ \midrule
        0.4 & KL & DE, DE Avg, CreINN, DDU, LB-BNN, RS-NN, LB-BNN Avg, SNN, E-CNN, EDL & [0.184, 0.195, 0.296, 0.309, 0.309, 0.402, 0.420, 0.481, 0.837, 0.909] \\ 
             & JS & RS-NN, SNN, DE Avg, DDU, LB-BNN Avg, LB-BNN, DE, CreINN, E-CNN, EDL & [0.068, 0.098, 0.099, 0.109, 0.119, 0.145, 0.168, 0.288, 0.856, 0.918] \\ \midrule
        0.5 & KL & DE Avg, DE, DDU, LB-BNN, CreINN, RS-NN, LB-BNN Avg, SNN, E-CNN, EDL & [0.195, 0.223, 0.309, 0.326, 0.356, 0.403, 0.420, 0.481, 0.998, 1.136] \\ 
             & JS & RS-NN, SNN, DE Avg, DDU, LB-BNN Avg, LB-BNN, DE, CreINN, E-CNN, EDL & [0.069, 0.098, 0.099, 0.109, 0.119, 0.161, 0.206, 0.348, 1.017, 1.145] \\ \midrule
        0.6 & KL & DE Avg, DE, DDU, LB-BNN, RS-NN, CreINN, LB-BNN Avg, SNN, E-CNN, EDL & [0.195, 0.261, 0.309, 0.342, 0.404, 0.415, 0.420, 0.481, 1.159, 1.363] \\ 
             & JS & RS-NN, SNN, DE Avg, DDU, LB-BNN Avg, LB-BNN, DE, CreINN, E-CNN, EDL & [0.070, 0.098, 0.099, 0.109, 0.119, 0.178, 0.245, 0.407, 1.178, 1.371] \\ \midrule
        0.7 & KL & DE Avg, DE, DDU, LB-BNN, RS-NN, LB-BNN Avg, CreINN, SNN, E-CNN, EDL & [0.195, 0.300, 0.309, 0.359, 0.405, 0.420, 0.475, 0.481, 1.319, 1.589] \\ 
             & JS & RS-NN, SNN, DE Avg, DDU, LB-BNN Avg, LB-BNN, DE, CreINN, E-CNN, EDL & [0.071, 0.098, 0.099, 0.109, 0.119, 0.194, 0.283, 0.467, 1.338, 1.598] \\ \midrule
        0.8 & KL & DE Avg, DDU, DE, LB-BNN, RS-NN, LB-BNN Avg, SNN, CreINN, E-CNN, EDL & [0.195, 0.309, 0.338, 0.376, 0.405, 0.420, 0.481, 0.535, 1.480, 1.816] \\ 
             & JS & RS-NN, SNN, DE Avg, DDU, LB-BNN Avg, LB-BNN, DE, CreINN, E-CNN, EDL & [0.072, 0.098, 0.099, 0.109, 0.119, 0.211, 0.322, 0.527, 1.499, 1.825] \\ \midrule
        0.9 & KL & DE Avg, DDU, DE, LB-BNN, RS-NN, LB-BNN Avg, SNN, CreINN, E-CNN, EDL & [0.195, 0.309, 0.377, 0.392, 0.406, 0.420, 0.481, 0.594, 1.641, 2.043] \\ 
             & JS & RS-NN, SNN, DE Avg, DDU, LB-BNN Avg, LB-BNN, DE, CreINN, E-CNN, EDL & [0.072, 0.098, 0.099, 0.109, 0.119, 0.228, 0.360, 0.586, 1.660, 2.052] \\ \midrule
        1.0 & KL & DE Avg, DDU, RS-NN, LB-BNN, DE, LB-BNN Avg, SNN, CreINN, E-CNN, EDL & [0.195, 0.309, 0.407, 0.409, 0.415, 0.420, 0.481, 0.654, 1.802, 2.270] \\ 
             & JS & RS-NN, SNN, DE Avg, DDU, LB-BNN Avg, LB-BNN, DE, CreINN, E-CNN, EDL & [0.073, 0.098, 0.099, 0.109, 0.119, 0.250, 0.392, 0.681, 1.810, 2.280] \\ 
        \bottomrule
    \end{tabular} }
\end{table}

\begin{figure}[!ht]
\centering
    \includegraphics[width=0.8\textwidth]{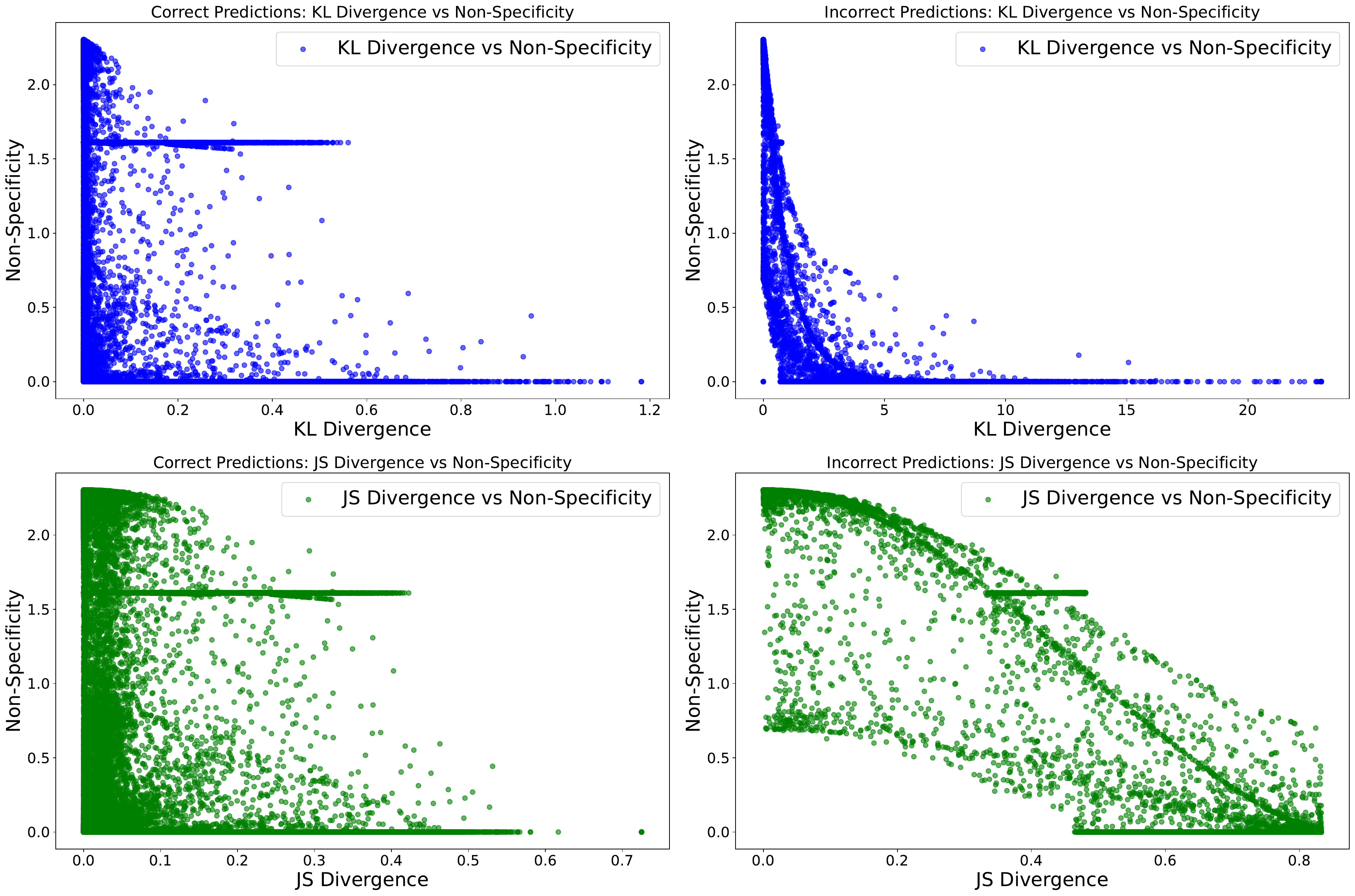}
    \caption{Scatter plots showing the relationship between uncertainty (KL and JS divergences) and non-specificity for correct and incorrect predictions across models. 
    }\label{fig:kl-vs-js-vs-ns}
\end{figure}

It is worth noting that the rankings for different values of the trade-off parameter $\lambda$ vary. This variation arises from the lack of adjustment for the range of $\lambda$ across the different metrics. For instance, if the KL divergence ranges from 0 to 25, while the JS divergence spans from 0 to 1, we should consider scaling down the range of $\lambda$ for the JS divergence accordingly.
To ensure that the influence of the JS divergence is properly scaled in the overall evaluation metric, we can define a scaling factor based on their maximum values:
$\text{Scaling Factor} = \frac{\max(\text{KL})}{\max(\text{JS})} = \frac{25}{1} = 25.
$
Given that KL divergence will generally dominate due to its larger range, we should scale down $\lambda$ for the JS divergence. A suitable adjusted range for $\lambda$ in the context of JS divergence can be expressed as: $
\lambda_{JS} = \lambda \times \left(\frac{1}{\text{Scaling Factor}}\right) = \lambda \times \frac{1}{25}.
$
Thus, while $\lambda$ initially ranges from 0 to 1, the effective range for $\lambda_{JS}$ will be from 0 to $\frac{1}{25}$. This adjustment ensures that both KL and JS divergences contribute appropriately to the overall evaluation metric $\mathcal{E}$, reflecting their respective scales during model selection.


In Fig. \ref{fig:kl-vs-js-vs-ns}, the set of scatter plots visualizes the relationship between distance measures (KL and JS divergences) and non-specificity for correct and incorrect predictions. The top row focuses on KL divergence, showing how it correlates with non-specificity for both correct and incorrect predictions. The bottom row highlights JS divergence with the same comparisons. In general, a positive trend between divergence and non-specificity indicates that as the model becomes more uncertain (higher non-specificity), its predictions deviate more from the true distribution (higher divergence). 

\textbf{Why choose KL divergence over JS divergence?} We selected KL divergence for our measure because, as discussed in \S{\ref{app:kl-theory}} and shown in Fig. \ref{fig:kl-js-eval}, JS is symmetric and bounded while KL divergence can grow unbounded when there is a significant mismatch between the predicted and true distributions, especially in small probabilities, making it more effective in distinguishing between correct and incorrect predictions. KL's greater sensitivity can provide a clearer signal when small deviations in probabilities matter, offering sharper differentiation in model evaluation. As Fig. \ref{fig:kl-js-eval} shows, KL offers a clearer separation between the two, with the plot showing the mean KL and JS values. This trend is further emphasized in Fig. \ref{fig:kl-js} where the full distributions of KL and JS are shown. Note that in Fig. \ref{fig:kl-js}, the ranges on the Y-axis differ significantly, with KL values spanning from 0 to 25, whereas JS values range only from 0 to 1.

\subsection{Ablation study on different non-specificity measures}  \label{app:ablation-ns}

In this section, we present an ablation study on the non-specificity (NS) measure in our evaluation metric $\mathcal{E}$. As we explained in Sec. \ref{app:nonspec-theory}, non-specificity is in fact an epistemic uncertainty measure of the credal set, and can be replaced with other suitable uncertainty measures related to credal sets. In this ablation, we present an alternative to the \citet{dubois1987properties} equation (\ref{eq:non_spec}) that we use, in the form of a different measure of credal uncertainty (CU) \citep{wang2024credal}.

This credal uncertainty (CU) measure can be computed as the difference between the credal upper bound $\overline{H}(\hat{\mathbb{C}re})$ and lower bounds $\underline{H}(\hat{\mathbb{C}re})$ computed as follows.
Given a set of $K$ individual predictive distributions from Bayesian Neural Networks, evidential models or deep ensembles, we can obtain an upper and lower probability bound for the $c_i$-th class element, denoted as $\overline{P}_{c_i}$ and $\underline{P}_{c_i}$, respectively. Recall that $\mathbf{Y} = \{ c_i, i \}$
where $i = 1, 2, \ldots, N$, and $N$ is the number of classes in $\mathcal{C}$.

These bounds can be computed as follows: $\overline{P}_{c_i} = \max \limits_{k=1,..,K} p_{k,c_i}, \quad \underline{P}_{c_i} = \min \limits_{k=1,..,K} p_{k,c_i}$,
where $p_{k,c_i}$ denotes the $c_i$-th class element of the $k$-th single probability vector $\mathbf{p}_k$. Such probability intervals over $\mathcal{C}$ classes define a non-empty credal set $\hat{\mathbb{C}re}$:
\begin{equation}
\hat{\mathbb{C}re} = \{ \mathbf{p} \mid \underline{P}_{c_i} \leq p_{c_i} \leq \overline{P}_{c_i}, \, \forall c_i \in \mathcal{C}, i = 1, 2, \ldots, N \}
\end{equation}
with the normalization condition: $\sum_{i=1}^{N} \underline{P}_{c_i} \leq 1 \leq \sum_{i=1}^{N} \overline{P}_{c_i}$.

It can be readily shown that the probability intervals given in the previous equations satisfy this condition, as follows:
\begin{equation}
\sum_{i=1}^{N} \underline{P}_{c_i} = \sum_{i=1}^{N} \min \limits_{k=1,..,K} p_{k,{c_i}} \leq \sum_{i=1}^{N} p_{k^*,{c_i}} = 1 \leq \sum_{i=1}^{N} \max \limits_{k=1,..,K} p_{k,{c_i}} = \sum_{i=1}^{N} \overline{P}_{c_i},
\end{equation}
where $k^*$ is any index in $1, \ldots, K$ and $c_i \in \mathcal{C}$. To estimate the uncertainty using $H(\hat{\mathbb{C}re})$ and $H(\hat{\mathbb{C}re})$, we need to solve the following optimization problems:

\begin{equation}
    \overline{H}(\hat{\mathbb{C}re}) = \operatorname{maximize} \sum_{i=1}^{N} -p_{c_i} \log_2 p_{c_i}, \quad
\underline{H}(\hat{\mathbb{C}re}) = \operatorname{minimize} \sum_{i=1}^{N} -p_{c_i} \log_2 p_{c_i},
\end{equation}
subject to the constraints: $
\sum_{i=1}^{N} p_{c_i} = 1, \quad p_{c_i} \in [\underline{P}_{c_i}, \overline{P}_{c_i}], \, \forall {c_i}$.

These optimization problems can be addressed using standard solvers, such as the SciPy optimization package \citep{virtanen2020scipy}.

\begin{figure}[!ht]
    \begin{minipage}[t]{0.48\textwidth}
    \includegraphics[width=\textwidth]{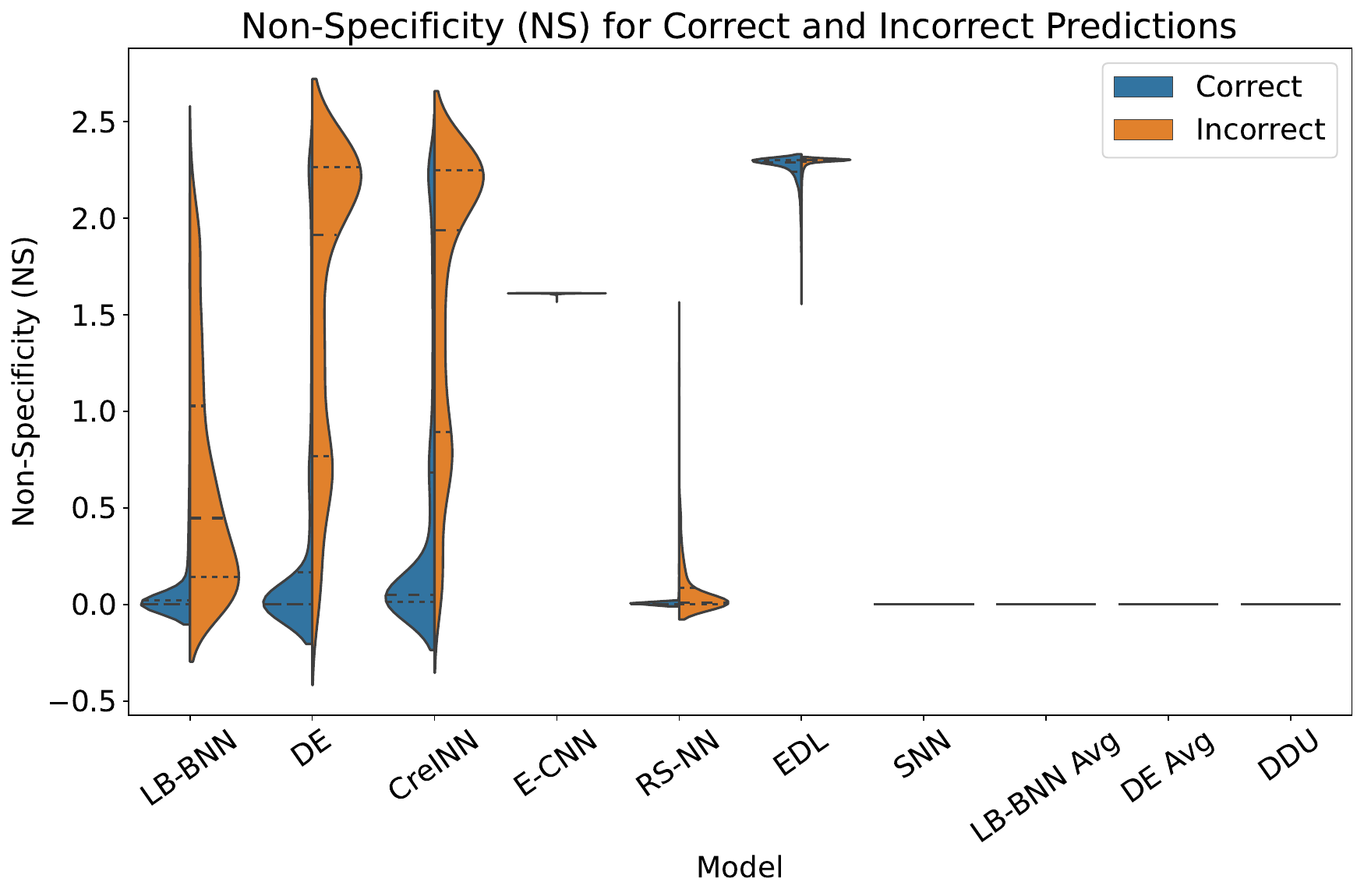}
    \caption*{(a)}
    \end{minipage} \hspace{0.02\textwidth}
    \begin{minipage}[t]{0.48\textwidth}
    \includegraphics[width=\textwidth]{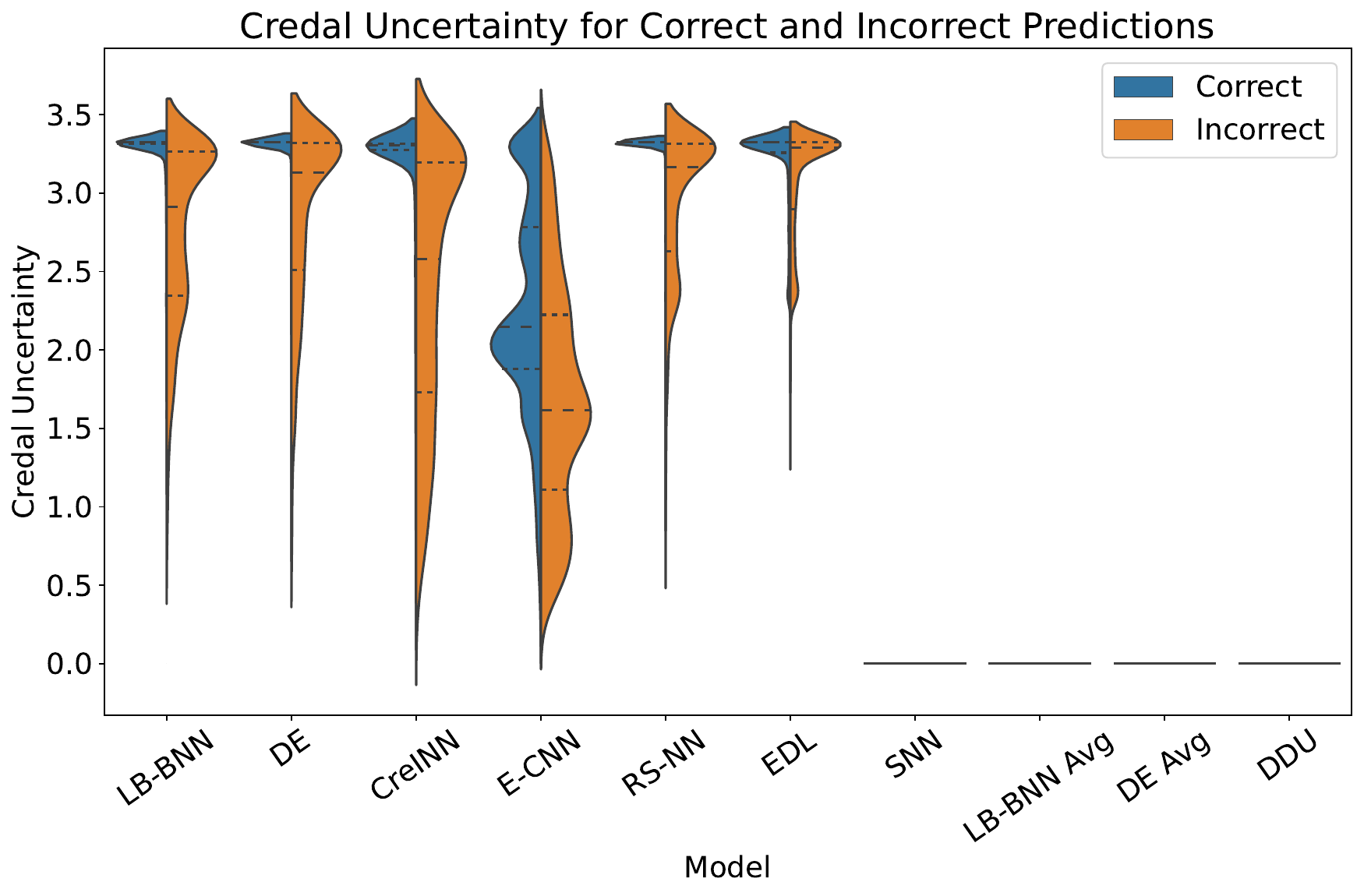}
    \caption*{(b)}
    \end{minipage}
    \caption{Comparison of (a) Non-Specificity (NS), and (b) Credal Uncertainty (CU) for Correctly Classified (CC) and Incorrectly Classified (ICC) samples from the CIFAR-10 dataset, for all models considered here. Notably, the scales of these two measures differ (Y-axis).
    }\label{fig:ns-cu}
\end{figure}

\begin{figure}[!h]
    \includegraphics[width=\textwidth]{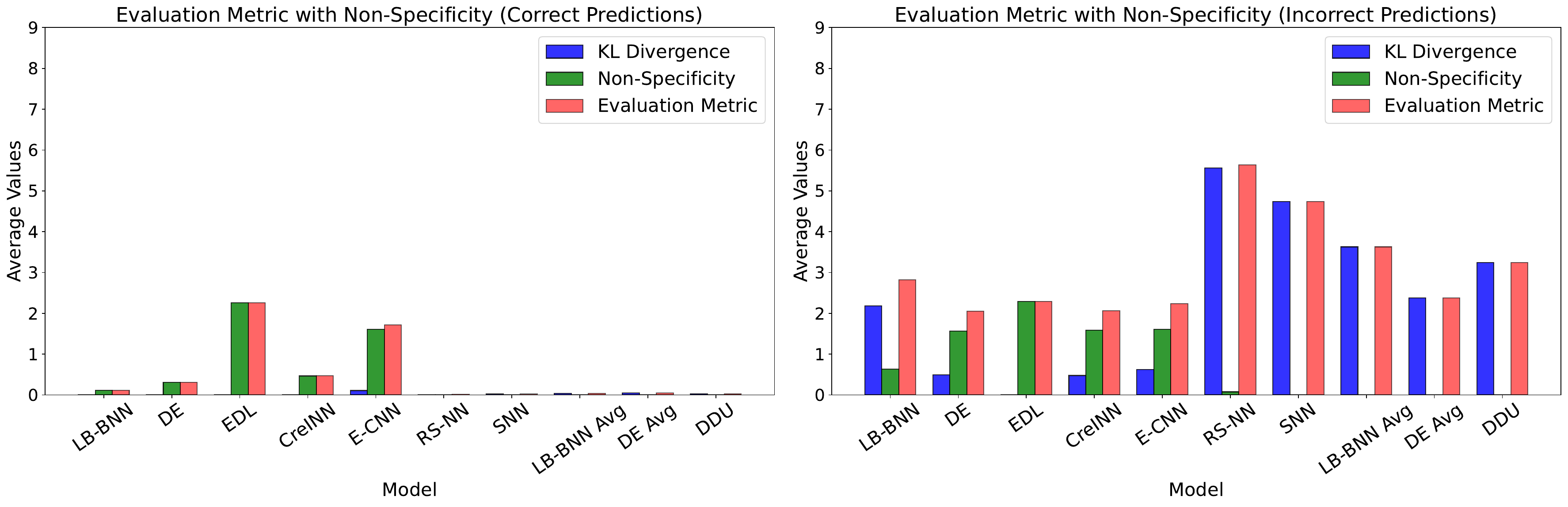}
    \includegraphics[width=\textwidth]{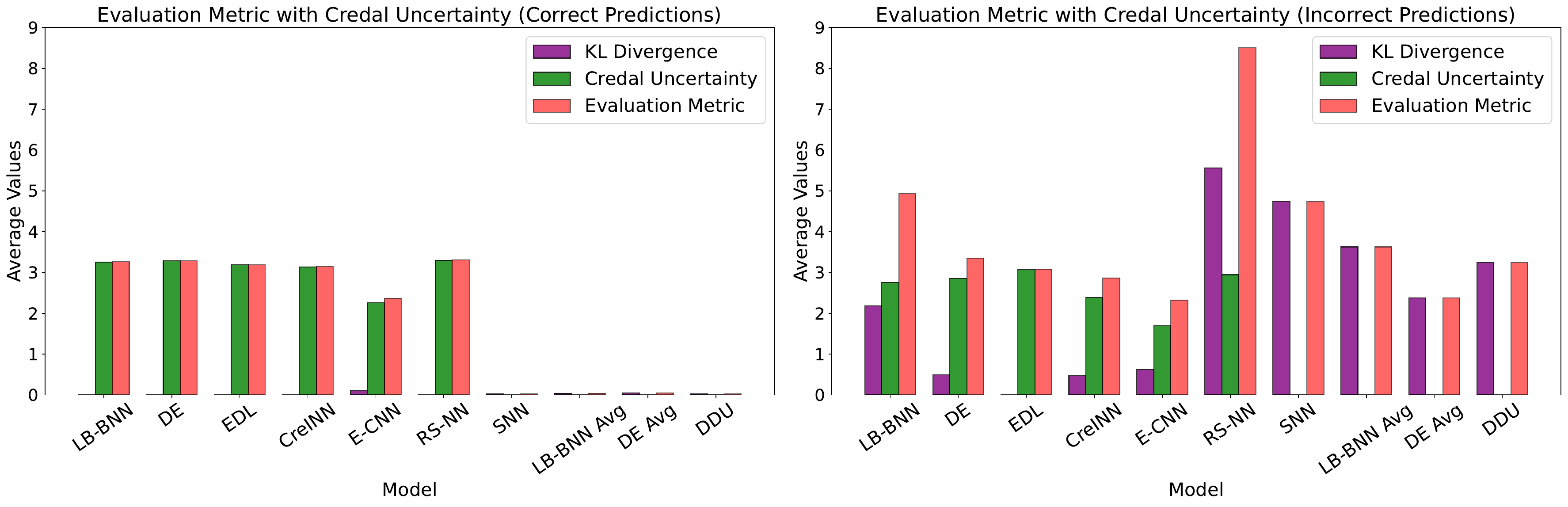}
    \caption{Comparison of mean Evaluation Metric $\mathcal{E}$ using mean Non-Specificity (top) and mean Credal Uncertainty (bottom) for Correct (left) and Incorrect (right) predictions of the CIFAR-10 dataset.
    }\label{fig:kl-ns-cu-bar}
\end{figure}

Fig. \ref{fig:ns-cu} shows the Non-Specificity (NS) (\ref{fig:ns-cu}a) and Credal Uncertainty (CU) (\ref{fig:ns-cu}b) over the entire test set. The violin plots depict distributions of the NS and CU for correct (blue) and incorrect (orange) predictions. We expect the both the measures to be larger for incorrect predictions and smaller for correct predictions. The plots show how the range of both values differ, but the trend remains the same, except for E-CNN. Non-specificity for E-CNN is high for both correct and incorrect predictions. 

Fig. \ref{fig:kl-ns-cu-bar} shows bar plots comparing Non-Specificity (NS) and Credal Uncertainty (CU) along with their influence on the overall evaluation metric. All values are shown as means. As discussed earlier, having larger upper bounds even for correct predictions often cause Credal Uncertainty to make the overall Evaluation Metric $\mathcal{E}$ larger. 

\begin{table}[!ht]
    \centering
    \caption{Model Rankings Based on Non-Specificity (NS) and Credal Uncertainty (CU) on the CIFAR-10 dataset. Model selection is based on the mean of Evaluation Metric ($\mathcal{E}$) with the models with the lowest $\mathcal{E}$ ranking first. The distance metric used here is KL divergence.}
    \label{tab:model_rank_ablation_ns}
    \resizebox{\textwidth}{!}{
    \begin{tabular}{@{}cc|l|l@{}}
        \toprule
        \textbf{Lambda} & \textbf{Metric} & \textbf{Model Ranking} & \textbf{Evaluation Metric ($\mathcal{E}$) Mean} \\ \midrule
        0.1 & NS & DE, CreINN, DE Avg, EDL, LB-BNN, DDU, E-CNN, RS-NN, LB-BNN Avg, SNN & [0.069, 0.117, 0.195, 0.229, 0.259, 0.309, 0.354, 0.399, 0.420, 0.481] \\ 
             & CU & DE Avg, DDU, EDL, DE, CreINN, E-CNN, LB-BNN Avg, SNN, LB-BNN, RS-NN &
[0.195, 0.309, 0.316, 0.357, 0.363, 0.410, 0.420, 0.481, 0.563, 0.726] \\ \midrule
        0.2 & NS & DE, CreINN, DE Avg, LB-BNN, DDU, RS-NN, LB-BNN Avg, EDL, SNN, E-CNN & [0.108, 0.177, 0.195, 0.276, 0.309, 0.400, 0.420, 0.456, 0.481, 0.515] \\ 
             & CU & DE Avg, DDU, LB-BNN Avg, SNN, E-CNN, EDL, CreINN, DE, LB-BNN, RS-NN & [0.195, 0.309, 0.420, 0.481, 0.627, 0.631, 0.668, 0.683, 0.883, 1.053]
\\ \midrule
        0.3 & NS & DE, DE Avg, CreINN, LB-BNN, DDU, RS-NN, LB-BNN Avg, SNN, E-CNN, EDL & [0.146, 0.195, 0.237, 0.293, 0.309, 0.401, 0.420, 0.481, 0.676, 0.682] \\ 
             & CU & DE Avg, DDU, LB-BNN Avg, SNN, E-CNN, EDL, CreINN, DE, LB-BNN, RS-NN &
[0.195, 0.309, 0.420, 0.481, 0.844, 0.945, 0.973, 1.009, 1.202, 1.380] \\ \midrule
        0.4 & NS & DE, DE Avg, CreINN, DDU, LB-BNN, RS-NN, LB-BNN Avg, SNN, E-CNN, EDL & [0.184, 0.195, 0.296, 0.309, 0.309, 0.402, 0.420, 0.481, 0.837, 0.909] \\ 
             & CU & DE Avg, DDU, LB-BNN Avg, SNN, E-CNN, EDL, CreINN, DE, LB-BNN, RS-NN &
[0.195, 0.309, 0.420, 0.481, 1.061, 1.259, 1.279, 1.335, 1.522, 1.707] \\ \midrule
        0.5 & NS & DE Avg, DE, DDU, LB-BNN, CreINN, RS-NN, LB-BNN Avg, SNN, E-CNN, EDL & [0.195, 0.223, 0.309, 0.326, 0.356, 0.403, 0.420, 0.481, 0.998, 1.136] \\ 
             & CU & DE Avg, DDU, LB-BNN Avg, SNN, E-CNN, EDL, CreINN, DE, LB-BNN, RS-NN & [0.195, 0.309, 0.420, 0.481, 1.277, 1.573, 1.584, 1.661, 1.842, 2.034]
\\ \midrule
        0.6 & NS & DE Avg, DE, DDU, LB-BNN, RS-NN, CreINN, LB-BNN Avg, SNN, E-CNN, EDL & [0.195, 0.261, 0.309, 0.342, 0.404, 0.415, 0.420, 0.481, 1.159, 1.363] \\ 
             & CU & DE Avg, DDU, LB-BNN Avg, SNN, E-CNN, EDL, CreINN, DE, LB-BNN, RS-NN &
[0.195, 0.309, 0.420, 0.481, 1.494, 1.888, 1.889, 1.987, 2.162, 2.362] \\ \midrule
        0.7 & NS & DE Avg, DE, DDU, LB-BNN, RS-NN, LB-BNN Avg, CreINN, SNN, E-CNN, EDL & [0.195, 0.300, 0.309, 0.359, 0.405, 0.420, 0.475, 0.481, 1.319, 1.589] \\ 
             & CU & DE Avg, DDU, LB-BNN Avg, SNN, E-CNN, CreINN, EDL, DE, LB-BNN, RS-NN &
[0.195, 0.309, 0.420, 0.481, 1.711, 2.194, 2.202, 2.313, 2.482, 2.689] \\ \midrule
        0.8 & NS & DE Avg, DDU, DE, LB-BNN, RS-NN, LB-BNN Avg, SNN, CreINN, E-CNN, EDL & [0.195, 0.309, 0.338, 0.376, 0.405, 0.420, 0.481, 0.535, 1.480, 1.816] \\ 
             & CU & DE Avg, DDU, LB-BNN Avg, SNN, E-CNN, CreINN, EDL, DE, LB-BNN, RS-NN &
[0.195, 0.309, 0.420, 0.481, 1.928, 2.499, 2.516, 2.639, 2.802, 3.016] \\ \midrule
        0.9 & NS & DE Avg, DDU, DE, LB-BNN, RS-NN, LB-BNN Avg, SNN, CreINN, E-CNN, EDL & [0.195, 0.309, 0.377, 0.392, 0.406, 0.420, 0.481, 0.594, 1.641, 2.043] \\ 
             & CU & DE Avg, DDU, LB-BNN Avg, SNN, E-CNN, CreINN, EDL, DE, LB-BNN, RS-NN &
[0.195, 0.309, 0.420, 0.481, 2.145, 2.805, 2.830, 2.965, 3.122, 3.343] \\ \midrule
        1.0 & NS & DE Avg, DDU, RS-NN, LB-BNN, DE, LB-BNN Avg, SNN, CreINN, E-CNN, EDL & [0.195, 0.309, 0.407, 0.409, 0.415, 0.420, 0.481, 0.654, 1.802, 2.270] \\ 
             & CU & DE Avg, DDU, LB-BNN Avg, SNN, E-CNN, CreINN, EDL, DE, LB-BNN, RS-NN &
[0.195, 0.309, 0.420, 0.481, 2.362, 3.110, 3.144, 3.292, 3.442, 3.671]
 \\ 
        \bottomrule
    \end{tabular} }
\end{table}

\textbf{Model Selection.} Tab. \ref{tab:model_rank_ablation_ns}, presents the rankings of various models based on their performance in terms of Non-Specificity (NS) and Credal Uncertainty (CU) on the CIFAR-10 dataset. The rankings are derived from the Evaluation Metric ($\mathcal{E}$), where models are ordered by their mean $\mathcal{E}$ values. The models are assessed under different values of the parameter lambda, with lower $\mathcal{E}$ values indicating better performance. For instance, at a lambda value of 0.1, the top-ranked models for Non-Specificity are DE, CreINN, and DE Avg. Similarly, for Credal Uncertainty, DE Avg and DDU lead the rankings, emphasizing their effectiveness in managing credal uncertainty within the predictions.

As lambda increases, the rankings reveal variations in model performance across both metrics. Notably, while DE Avg and DDU consistently ranks highly across different lambda values, other models exhibit fluctuating positions. If we do not consider models that predict point predictions, since their NS and CU are zero and $\mathcal{E}$ is the same as KL, DE and E-CNN are the best ranked models for CU and DE and RS-NN are the best ranked models for NS.





\textbf{Why choose Non-Specificity (NS) over Credal Uncertainty?} We chose Non-Specificity (NS) over Credal Uncertainty (CU) for our measure because, as demonstrated in Fig. \ref{fig:ns-cu}, CU struggles to differentiate the uncertainty between correct and incorrect predictions across all models. The mean CU remains high for both cases, as highlighted in Fig. \ref{fig:kl-ns-cu-bar}. In contrast, NS offers a clearer distinction, with lower values for correct predictions and higher values for incorrect ones. CU, on the other hand, tends to exhibit a high upper bound for both correct and incorrect predictions, and remains consistently elevated for correct predictions regardless of the model's performance. This does not accurately reflect the true state of uncertainty within the credal set.

\subsection{Evaluation on Additional Datasets (MNIST and CIFAR-100)} \label{app:addional_dataset_eval}

\begin{figure}[!ht]
\centering
    \begin{minipage}[t]{0.49\textwidth}
    \includegraphics[width=\textwidth]{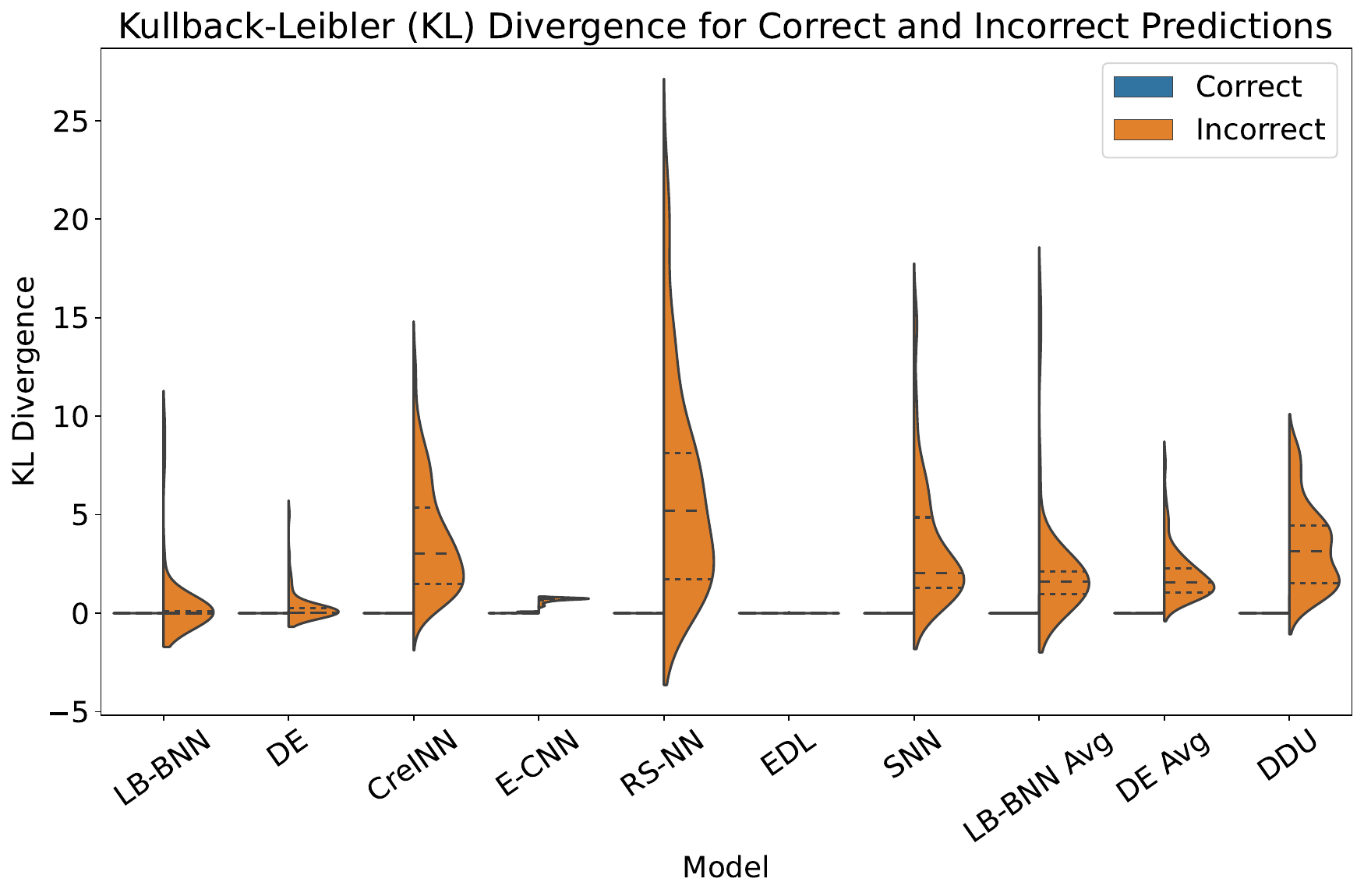}
    \caption*{(a)} \label{fig:mnist_kl}
    \end{minipage}
    \begin{minipage}[t]{0.49\textwidth}
    \includegraphics[width=\textwidth]{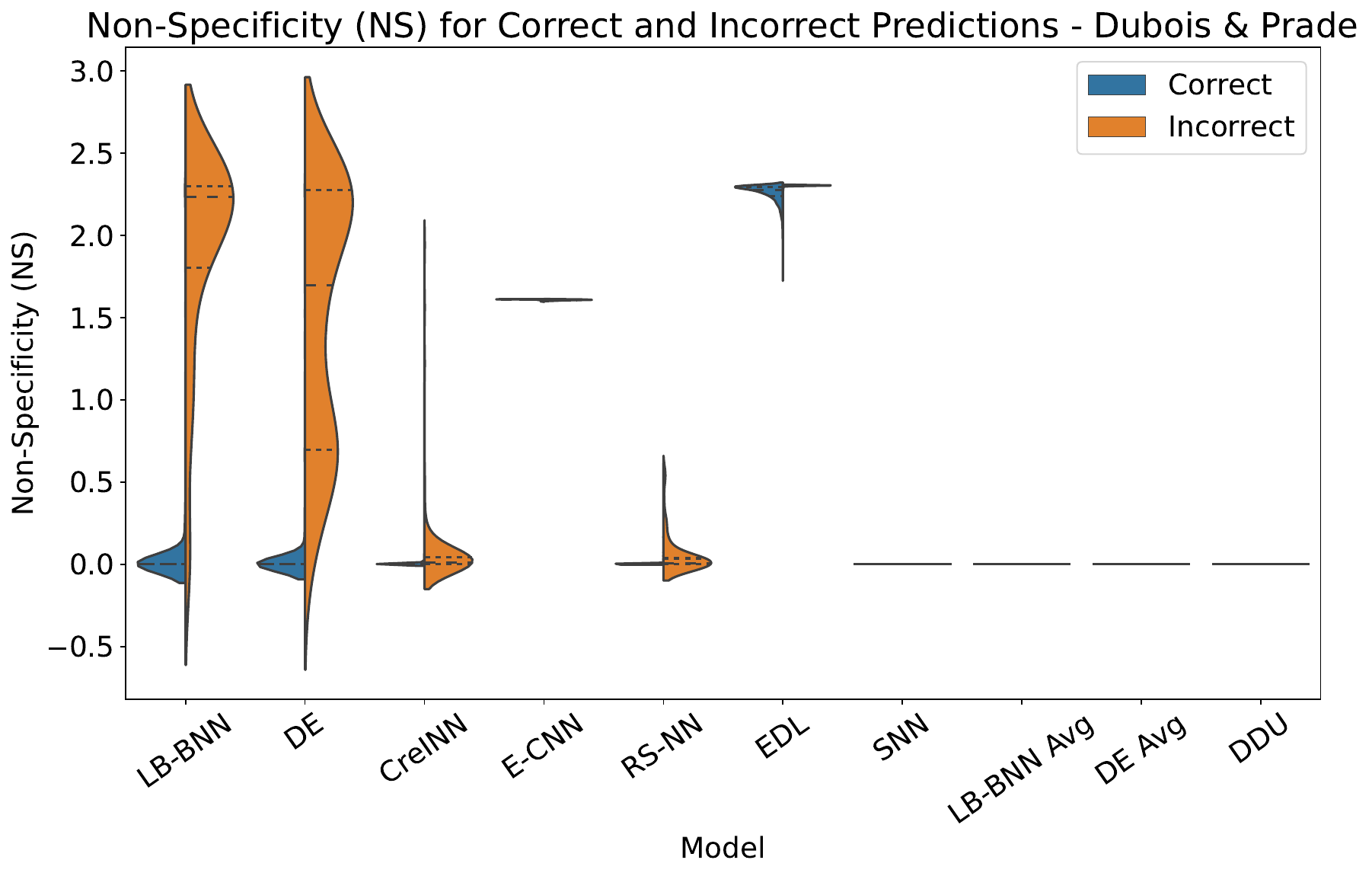}
    \caption*{(b)}\label{fig:mnist_ns}
    \end{minipage}
        \begin{minipage}[t]{0.47\textwidth}
    \includegraphics[width=\textwidth]{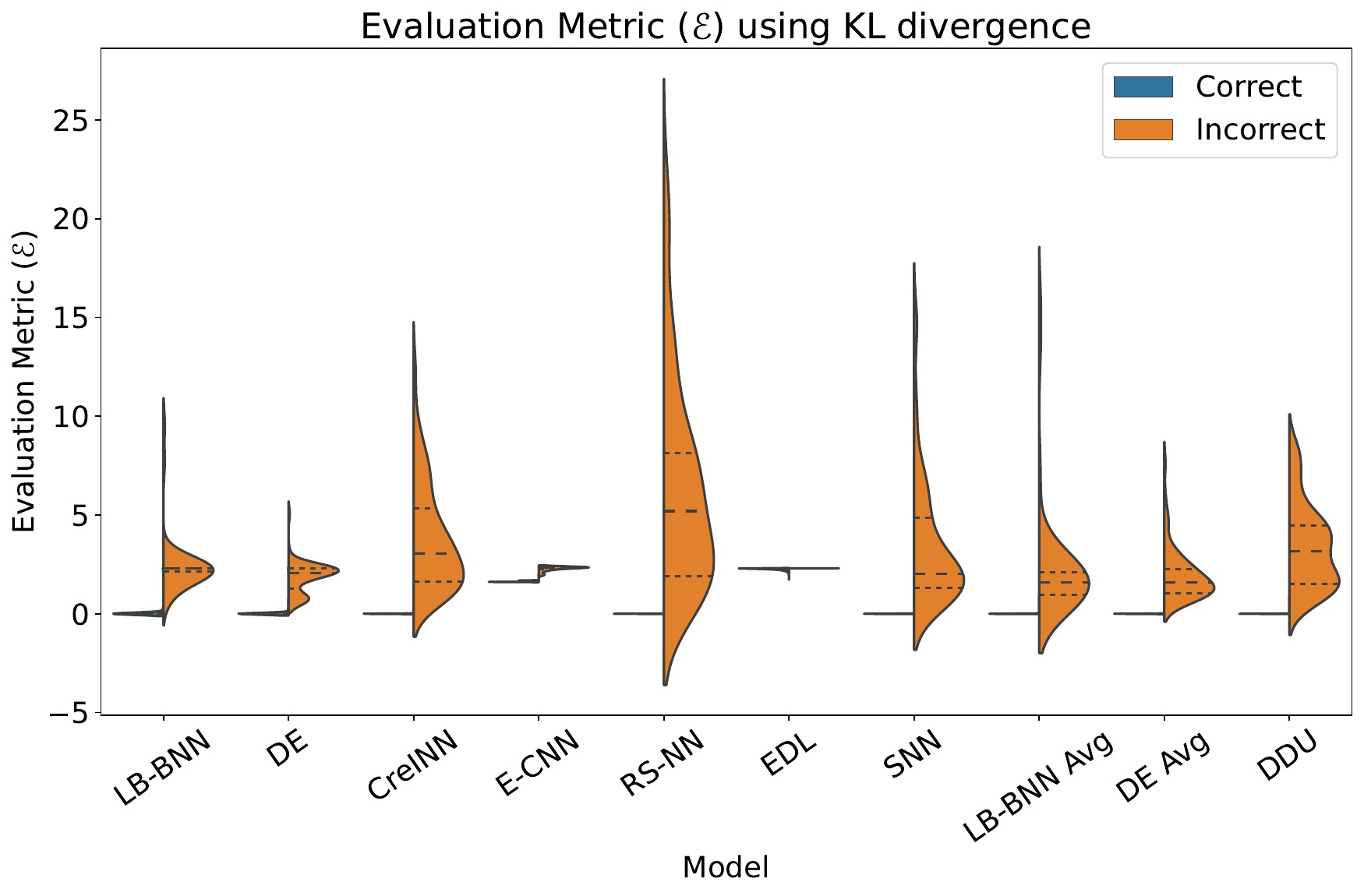}
    \caption*{(c)} \label{fig:mnist_e}
    \end{minipage}
    \begin{minipage}[b]{0.52\textwidth}
    \includegraphics[width=\textwidth]{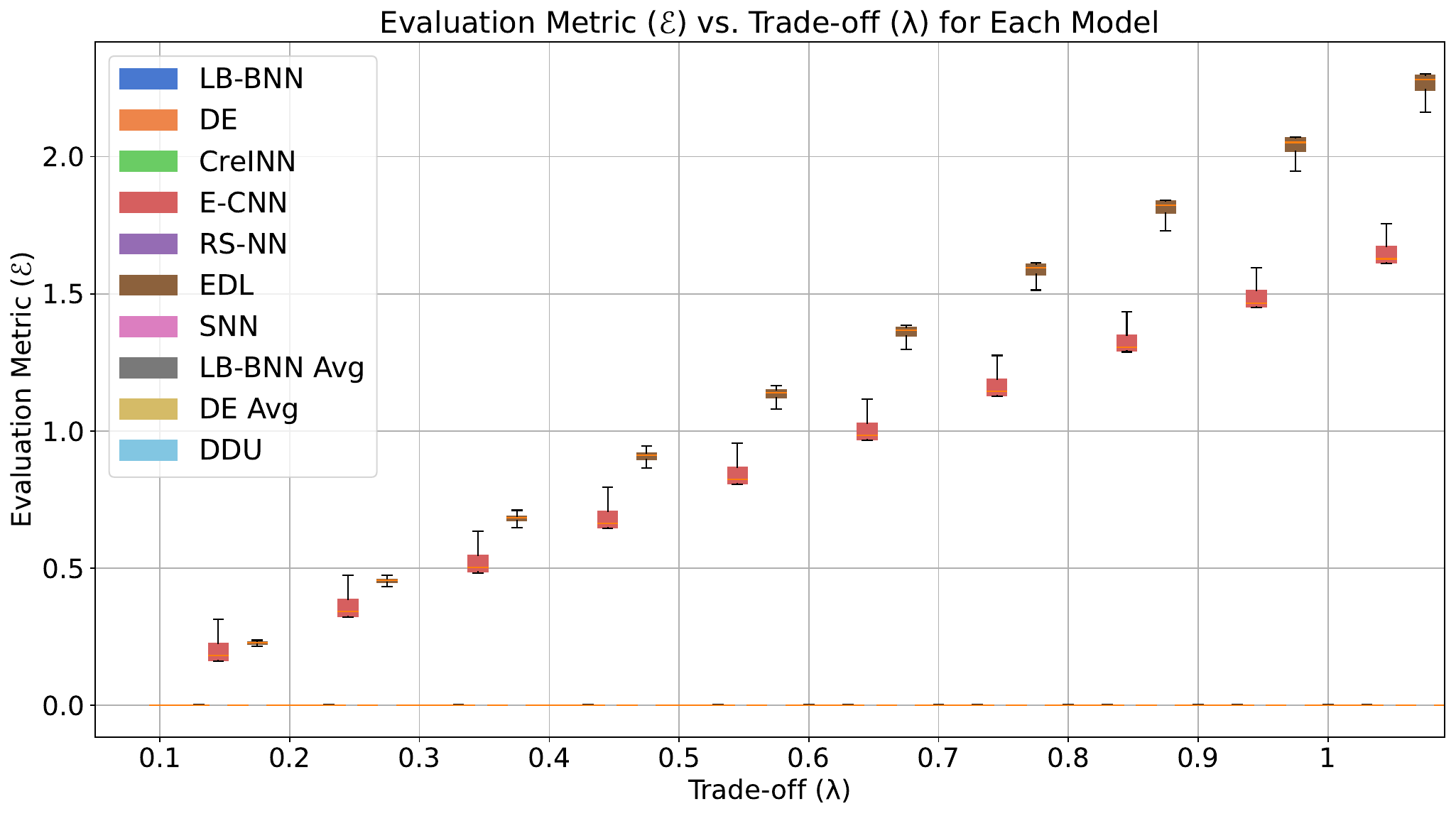}
    \caption*{(d)}  \label{fig:mnist_lambda}
\end{minipage} 
    \caption{Measures of (a) Kullback-Leibler (KL) divergence, (b) Non-specificity (NS), (c) Evaluation Metric ($\mathcal{E}$) for Correctly Classified (CC) and Incorrectly Classified (ICC) samples (d) Evaluation metric ($\mathcal{E}$) estimates vs trade-off ($\lambda$) for all models on the MNIST dataset.
    }\label{fig:app_kl_ns_eval_mnist}
\end{figure}

The distributions of KL-divergence, non-specificity and evaluation metric $\mathcal{E}$ for correct and incorrect samples on MNIST and CIFAR-100 datasets for all models are illustrated in Figs. \ref{fig:app_kl_ns_eval_mnist} and \ref{fig:app_kl_ns_eval_cifar100}, respectively. KL-divergence is computed as per usual (Figs. \ref{fig:app_kl_ns_eval_mnist}(a), \ref{fig:app_kl_ns_eval_cifar100}(a)), 
while non-specificity is obtained using Eq. \ref{eq:non_spec} (Figs. \ref{fig:app_kl_ns_eval_mnist}(b), \ref{fig:app_kl_ns_eval_cifar100}(b)). Finally, the Evaluation Metric ($\mathcal{E}$) is calculated using Eq. \ref{eq:metric} (Figs. \ref{fig:app_kl_ns_eval_mnist}(c), \ref{fig:app_kl_ns_eval_cifar100}(c)). 

For point predictions, however, the non-specificity is 0, thus the Evaluation Metric ($\mathcal{E}$) reduces to just the KL divergence between ground truth and the prediction. Fig. \ref{fig:app_kl_ns_eval_mnist}(c) and \ref{fig:app_kl_ns_eval_cifar100}(c) shows the Evaluation Metric ($\mathcal{E}$) for all models with the trade-off parameter $\lambda = 1$. All the models exhibit very concentrated distributions for KL-divergence for correct samples for both datasets, while non-specificity, and consquently, evaluation metric $\mathcal{E}$ appears more distributed. Incorrect samples, on the other hand, show more smooth distributions for all. 

In Tab. \ref{tab:cc_icc}, a comparison of KL divergence, Non-Specificity, and Evaluation Metric ($\mathcal{E}$) with trade-off $\lambda$ = 1 is presented for Correctly Classified (CC) and Incorrectly Classified (ICC) samples across each model. Incorrect samples demonstrate a higher mean value than correct samples for KL divergence, Non-Specificity, and Evaluation Metric ($\mathcal{E}$) indicating that the models are behaving appropriately, except ECNN which gives high non-specificity for both correct and incorrect predictions. We can form a tentative ranking of the models with Tab. \ref{tab:cc_icc}. For correct predictions of CIFAR-10, MNIST, and CIFAR-100, we desire low KL and low non-specificity, as exhibited by DE and RS-CNN respectively. For incorrect predictions, a low KL and high non-specificity is optimal. Finally, evaluation metric $\mathcal{E}$ ($\lambda = 1$) should reflect a low value for both correct and incorrect predictions (RS-NN and DE).

\begin{table}[!ht]
\caption{Comparison of KL divergence (KL), Non-Specificity (NS), and Evaluation Metric ($\mathcal{E}$) (trade-off $\lambda = 1$) for Correctly Classified (CC) and Incorrectly Classified (ICC) samples for each model on three datasets: CIFAR-10, MNIST, and CIFAR-100.}
\label{tab:cc_icc}
\centering
\resizebox{\textwidth}{!}{
\begin{tabular}{lccccccc}
\toprule
\multirow{2}{*}{Dataset} &
  \multirow{2}{*}{Model} &
  \multicolumn{2}{c}{KL distance (KL)} &
  \multicolumn{2}{c}{Non-Specificity (NS)} &
  \multicolumn{2}{c}{Evaluation metric ($\mathcal{E}$)} \\
\cmidrule(lr){3-4} \cmidrule(lr){5-6} \cmidrule(lr){7-8}
&& CC ($\downarrow$) & ICC ($\downarrow$)  & CC ($\downarrow$) & ICC ($\uparrow$) & CC ($\downarrow$) & ICC ($\downarrow$)\\
\midrule
\multirow{6}{*}{CIFAR-10} 
& LB-BNN & $0.009 \pm 0.038$ & $2.181 \pm 3.444$ & $0.109 \pm 0.323$ & $0.637 \pm 0.600$ & $0.118 \pm 0.341$ & $2.818 \pm 3.203$\\
& DE & $0.0001 \pm 0.001$ & $0.489 \pm 1.395$ & $0.306 \pm 0.639$ & $1.566 \pm 0.756$ & $0.306 \pm 0.639$ & $2.055 \pm 1.183$\\
& EDL & $\mathbf{0.00005} \pm \mathbf{0.0002}$ & $\mathbf{0.0051} \pm \mathbf{0.017}$ & $2.253 \pm	0.080$ & $\mathbf{2.288} \pm \mathbf{0.034}$ & $ 2.253 \pm 0.08$ & $2.293 \pm 0.023$ \\
& CreINN & $0.003 \pm 0.006$ & $0.476 \pm 1.001$ & $0.465 \pm 0.727$ & $1.590 \pm 0.731$ & $0.468 \pm 0.728$ & $\mathbf{2.066} \pm \mathbf{0.744}$\\
& E-CNN & $0.108 \pm 0.093$ & $0.625 \pm 0.115$ & $1.609 \pm 0.003$ & $1.610 \pm 0.001$ & $1.717 \pm 0.092$ & $2.235 \pm 0.115$\\
& RS-NN & $0.009 \pm 0.055$ & $5.562 \pm 4.747$ & $\mathbf{0.004} \pm \mathbf{0.031}$ & $0.076 \pm 0.145$ & $\mathbf{0.013} \pm \mathbf{0.077}$ & $5.638 \pm 4.691$\\
  \cmidrule{2-8}
& SNN & $0.021 \pm 0.090$ & $4.735 \pm 3.604$ & $0.000 \pm 0.000$ & $0.000 \pm 0.000$ & $0.021 \pm 0.090$ & $4.735 \pm 3.604$\\
& LB-BNN Avg & $0.034 \pm 0.114$ & $3.628 \pm 3.145$ & $0.000 \pm 0.000$ & $0.000 \pm 0.000$ & $0.034 \pm 0.114$ & $3.628 \pm 3.145$\\
& DE Avg & $0.050 \pm 0.135$ & $2.378 \pm 2.000$ & $0.000 \pm 0.000$ & $0.000 \pm 0.000$ & $0.050 \pm 0.135$ & $2.378 \pm 2.000$\\
& DDU & $0.031 \pm	0.104$ & $3.243 \pm	2.196$ & $0.000 \pm 0.000$ & $0.000 \pm 0.000$ & $0.031 \pm	0.104$ & $3.243 \pm	2.196$\\
\midrule
\multirow{6}{*}{MNIST} 
& LB-BNN & $0.00002 \pm 0.001$ & $0.490 \pm 1.832$ & $0.083 \pm 0.359$ & $\mathbf{1.883} \pm \mathbf{0.656}$ & $0.083 \pm 0.359$ & $2.373 \pm 1.449$\\
& DE & $0.00005 \pm 0.001$ & $0.340 \pm 0.805$ & $0.058 \pm 0.291$ & $1.503 \pm 0.766$ & $0.058 \pm 0.291$ & $\mathbf{1.843} \pm \mathbf{0.758}$\\
& EDL & $\mathbf{0.00006} \pm \mathbf{0.00002}$ & $\mathbf{0.00062} \pm \mathbf{0.005}$ & $2.255 \pm 0.055$ & $2.301 \pm 0.010$ & $2.255 \pm 0.055$ &  $2.302\pm 0.007$\\
& CreINN & $0.006 \pm 0.037$ & $3.720 \pm 2.714$ & $0.004 \pm 0.032$ & $0.061 \pm 0.212$ & $0.010 \pm 0.053$ & $3.781 \pm 2.664$\\
& E-CNN & $0.032 \pm 0.033$ & $0.675 \pm 0.115$ & $1.608 \pm 0.004$ & $1.607 \pm 0.003$ & $1.641 \pm 0.032$ & $2.282 \pm 0.116$\\
& RS-NN & $0.001 \pm 0.023$ & $6.211 \pm 5.290$ & $\mathbf{0.0004} \pm \mathbf{0.010}$ & $0.056 \pm 0.120$ & $\mathbf{0.002} \pm \mathbf{0.031}$ & $6.267 \pm 5.242$\\
  \cmidrule{2-8}
& SNN & $0.004 \pm 0.040$ & $3.484 \pm 3.234$ & $0.000 \pm 0.000$ & $0.000 \pm 0.000$ & $0.004 \pm 0.040$ & $3.484 \pm 3.234$\\
& LB-BNN Avg & $0.006 \pm 0.044$ & $2.295 \pm 2.921$ & $0.000 \pm 0.000$ & $0.000 \pm 0.000$ & $0.006 \pm 0.044$ & $2.295 \pm 2.921$\\
& DE Avg & $0.007 \pm 0.048$ & $1.956 \pm 1.287$ & $0.000 \pm 0.000$ & $0.000 \pm 0.000$ & $0.007 \pm 0.048$ & $1.956 \pm 1.287$\\
& DDU & $0.003	\pm 0.032$ & $3.367 \pm	2.092$ & $0.000 \pm 0.000$ & $0.000 \pm 0.000$ & $0.003 \pm 0.032$ & $3.367 \pm	2.092$\\
\midrule
\multirow{6}{*}{CIFAR-100} 
& LB-BNN & $0.006 \pm 0.021$ & $0.399 \pm 0.783$ & $1.790 \pm 1.748$ & $3.353 \pm 1.309$ & $1.796 \pm 1.752$ & $3.752 \pm 0.947$\\
& DE & $\mathbf{0.0001} \pm \mathbf{0.001}$ & $0.074 \pm 0.477$ & $2.780 \pm 2.013$ & $\mathbf{4.331} \pm \mathbf{0.836}$ & $2.780 \pm 2.013$ & $4.405 \pm 0.690$\\
& EDL & $0.005 \pm 0.190$ & $\mathbf{0.015} \pm \mathbf{0.194}$ & $2.629 \pm 2.086$ & $4.114	\pm 1.258$ & $2.634 \pm 2.090$ & $4.129 \pm 1.238$\\
& CreINN & $0.310 \pm 0.321$ & $0.857 \pm 0.668$ & $1.483 \pm	1.132$ & $2.232 \pm	1.147$ & $1.792 \pm 1.131$ & $\mathbf{3.089} \pm \mathbf{0.599}$\\
& E-CNN & - & - & - & - & - & -\\
& RS-NN & $0.042 \pm 0.130$ & $4.539 \pm 5.857$ & $\mathbf{0.242} \pm \mathbf{0.760}$ & $1.239 \pm 1.510$ & $\mathbf{0.284} \pm \mathbf{0.783}$ & $5.777 \pm 5.276$\\
  \cmidrule{2-8}
& SNN & $0.063 \pm 0.165$ & $6.528 \pm 4.868$ & $0.000 \pm 0.000$ & $0.000 \pm 0.000$ & $\mathbf{0.063} \pm \mathbf{0.165}$ & $6.528 \pm 4.868$\\
& LB-BNN Avg & $0.167 \pm 0.255$ & $4.235 \pm 3.548$ & $0.000 \pm 0.000$ & $0.000 \pm 0.000$ & $0.167 \pm 0.255$ & $4.235 \pm 3.548$\\
& DE Avg & $0.284 \pm 0.349$ & $3.285 \pm 2.696$ & $0.000 \pm 0.000$ & $0.000 \pm 0.000$ & $0.284 \pm 0.349$ & $\mathbf{3.285} \pm \mathbf{2.696}$\\
& DDU & $0.102 \pm	0.228$ & $4.161 \pm	2.646$ & $0.000 \pm 0.000$ & $0.000 \pm 0.000$ & $0.102 \pm	0.228$ & $4.161 \pm	2.646$\\
\bottomrule
\end{tabular}
} 
\vspace{-4mm}
\end{table}

\subsection{Additional ablation on trade-off parameter} 
\label{app:trade_off_parameter}

Figs. \ref{fig:app_kl_ns_eval_mnist}(d) and \ref{fig:app_kl_ns_eval_cifar100}(d) demonstrates the variation in Evaluation Metric ($\mathcal{E}$) values across different trade-off ($\lambda$) settings for each model on MNIST and CIFAR-100 datasets respectively. Each box in the plot represents the spread of $\mathcal{E}$ values for a specific model. The bottom and top edges of the box correspond to the 25\% (Q1) and 75\% (Q3) percentiles, respectively, while the line inside the box indicates the median $\mathcal{E}$ value. 
The whiskers extend from the edges of the box to illustrate the range of non-outlier values. 
For MNIST (Fig. \ref{fig:app_kl_ns_eval_mnist}(d)), all the models except E-CNN show relatively consistent values across all values $\lambda$, indicating that the models are very precise for this dataset. Whereas, for CIFAR-100 (Figs. \ref{fig:app_kl_ns_eval_cifar100}(d)), there is a noticeable trend of increasing $\mathcal{E}$ values as $\lambda$ grows showcasing the potential of trade-off parameter for model selection. These results are also shown in Tab. \ref{tab:lambda-1}.

In Tab. \ref{tab:lambda-1}, the MNIST results show that DE has the lowest $\mathcal{E}$ for $\lambda$ values 0.1 to 0.7. therefore, the model selection will return DE as the best model. As $\lambda$ increases from 0.7 to 1, RS-NN has the lowest scores and the model selection procedure returns RS-NN as the best model. Overall, these are models with low $\mathcal{E}$ scores. These results are also reflected in Tab. \ref{tab:model_selection_mnist}.

In the CIFAR-100 dataset, the results are even more pronounced. The evaluation metric $\mathcal{E}$ start at a relatively high baseline for most models. The standard deviations also illustrate substantial variability among models, particularly at higher $\lambda$ values, where some models demonstrate higher susceptibility to the increase in non-specificity.
\newpage
\begin{figure}[!ht]
\centering
    \begin{minipage}[t]{0.49\textwidth}
    \includegraphics[width=\textwidth]{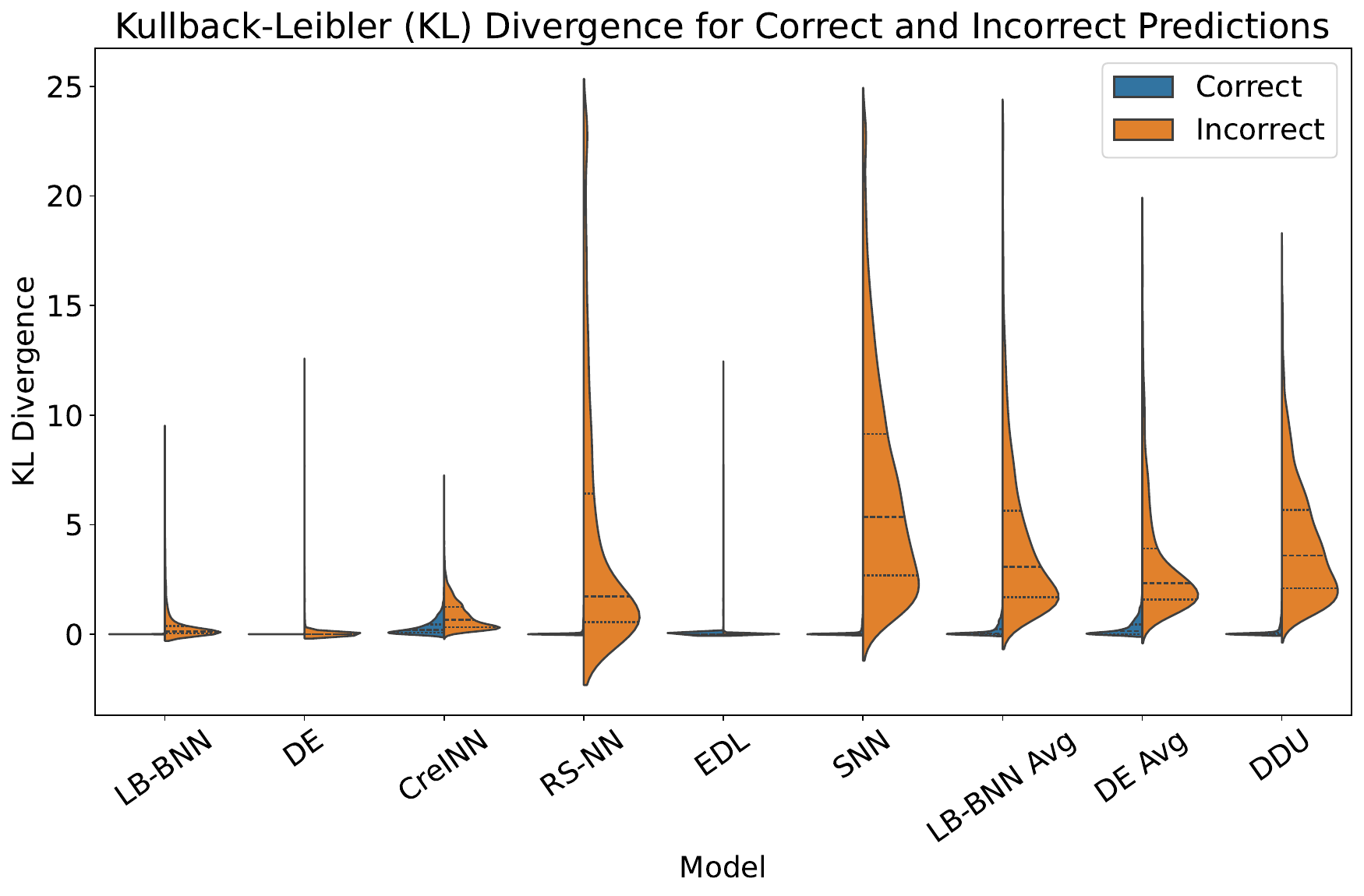}
    \caption*{(a)} \label{fig:cifar100_kl}
    \end{minipage}
    \begin{minipage}[t]{0.49\textwidth}
    \includegraphics[width=\textwidth]{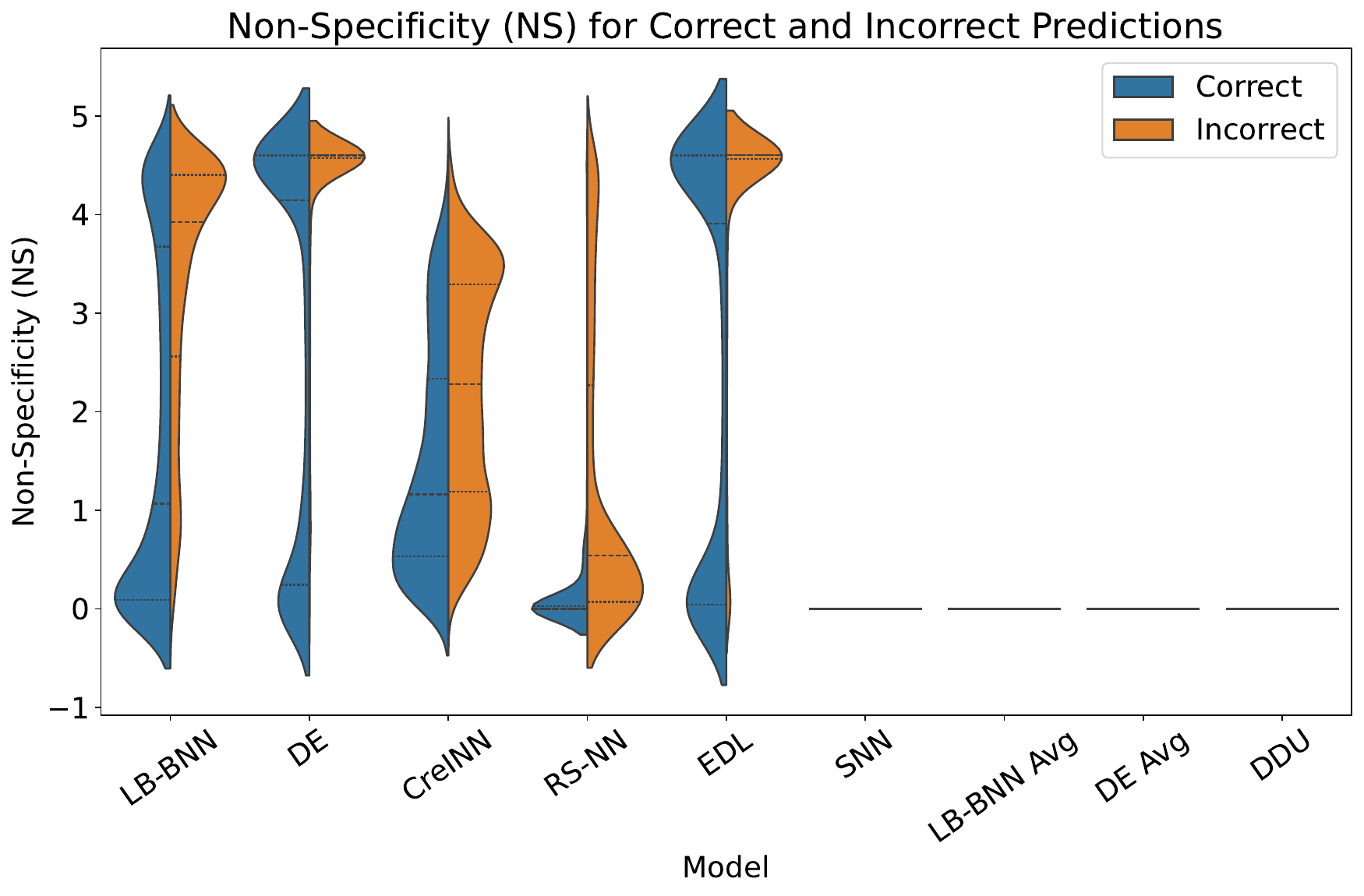}
    \caption*{(b)}\label{fig:cifar100_ns}
    \end{minipage}
        \begin{minipage}[t]{0.59\textwidth}
    \includegraphics[width=\textwidth]{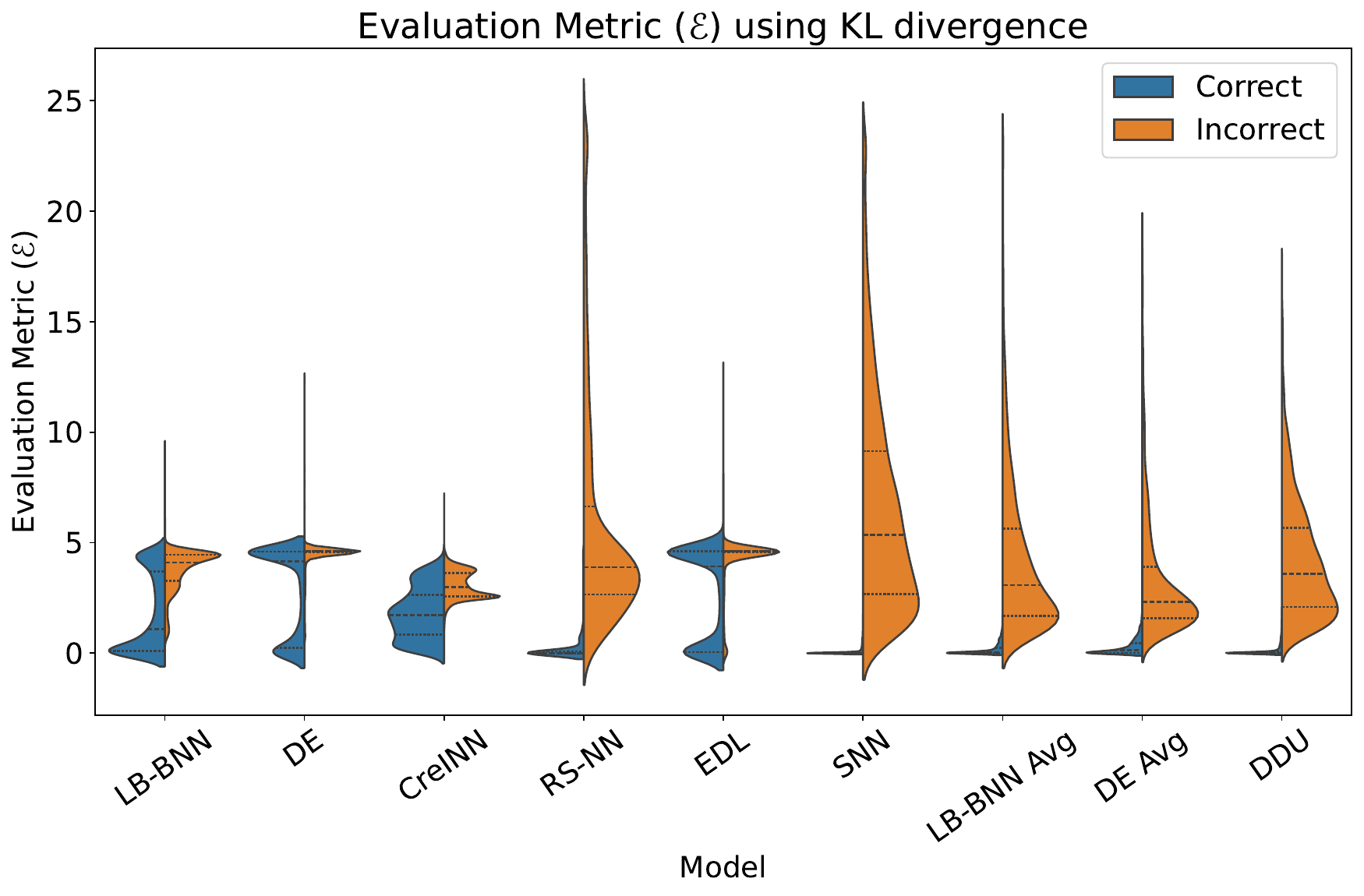}
    \caption*{(c)} \label{fig:cifar100_e}
    \end{minipage} 
    \begin{minipage}[b]{\textwidth}
    \includegraphics[width=\textwidth]{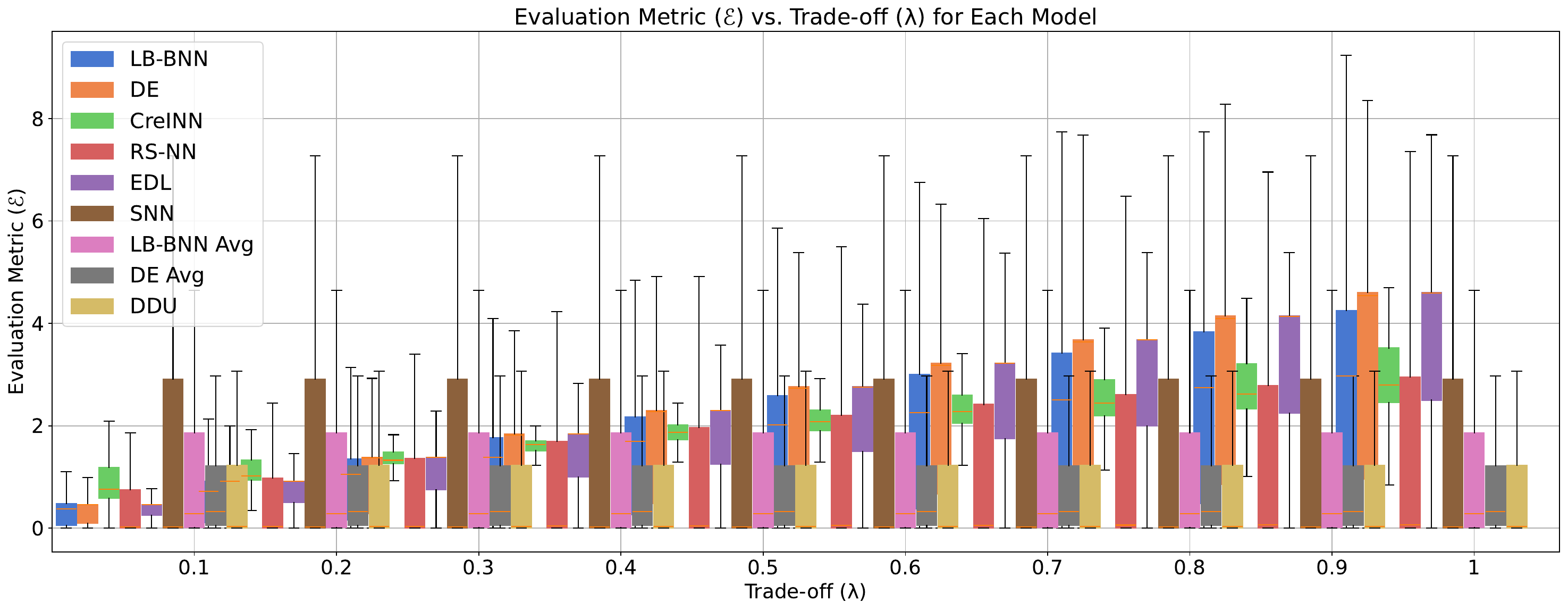}
    \caption*{(d)}  \label{fig:cifar100_lambda}
\end{minipage} 
    \caption{Measures of (a) Kullback-Leibler (KL) divergence, (b) Non-specificity (NS), (c) Evaluation Metric ($\mathcal{E}$) for Correctly Classified (CC) and Incorrectly Classified (ICC) samples (d) Evaluation metric ($\mathcal{E}$) estimates vs trade-off ($\lambda$) for all models on the CIFAR-100 dataset.
    }\label{fig:app_kl_ns_eval_cifar100}
\end{figure}

\begin{table}[!h]
\caption{ 
Trade-off ($\lambda$) vs Evaluation Metric ($\mathcal{E}$) for different values of $\lambda$ for the CIFAR-10 dataset. \label{tab:lambda-1}}
\centering
\resizebox{\textwidth}{!}{
\begin{tabular}{lccccccc}
\toprule
\multirow{2}{*}{Dataset} &
 \multirow{2}{*}{\begin{tabular}[c]{@{}c@{}}Trade-off\\ parameter ($\lambda$)\end{tabular}} 
    & \multicolumn{5}{c}{Evaluation Metric ($\mathcal{E}$)} \\
  \cmidrule{3-8}
& & LB-BNN  & DE & EDL & CreINN  & E-CNN & RS-NN \\
  \midrule
  \multicolumn{1}{c}{\multirow{6}{*}{CIFAR-10}}
& 0.1 & $0.259 \pm 1.316$ & $\mathbf{0.069} \pm \mathbf{0.375}$ & $0.229 \pm 0.012$ & $0.117 \pm 0.382$ & $0.354 \pm 0.215$ & $0.399 \pm 1.895$\\
& 0.2 & $0.276 \pm 1.319$ & $\mathbf{0.108} \pm \mathbf{0.395}$ & $0.455 \pm 0.016$ & $0.177 \pm 0.407$ & $0.515 \pm 0.215$ & $0.400 \pm 1.896$\\
& 0.3 & $0.293 \pm 1.323$ & $\mathbf{0.146} \pm \mathbf{0.426}$ &  $0.682 \pm 0.022$& $0.237 \pm 0.445$ & $0.676 \pm 0.215$ & $0.401 \pm 1.896$\\
& 0.4 & $0.309 \pm 1.327$ & $\mathbf{0.184} \pm \mathbf{0.467}$ & $0.909 \pm 0.029$  & $0.296 \pm 0.494$ & $0.837 \pm 0.215$ & $0.402 \pm 1.897$\\
& 0.5 & $0.326 \pm 1.334$ & $\mathbf{0.223} \pm \mathbf{0.514}$ & $1.135 \pm 0.035$ & $0.356 \pm 0.550$ & $0.998 \pm 0.215$ & $0.403 \pm 1.897$\\
& 0.6 & $0.342 \pm 1.341$ & $\mathbf{0.261} \pm \mathbf{0.566}$ &  $1.362 \pm 0.042$ & $0.415 \pm 0.612$ & $1.159 \pm 0.215$ & $0.404 \pm 1.898$\\
& 0.7 & $0.359 \pm 1.349$ & $\mathbf{0.300} \pm \mathbf{0.622}$ &  $1.588 \pm 0.049$  &  $0.475 \pm 0.679$ & $1.319 \pm 0.215$ & $0.405 \pm 1.898$\\
& 0.8 & $0.376 \pm 1.359$ & $\mathbf{0.338} \pm \mathbf{0.681}$ &  $1.815 \pm 0.056$   &  $0.535 \pm 0.748$ & $1.480 \pm 0.215$ & $0.405 \pm 1.899$\\
& 0.9 & $0.392 \pm 1.369$ & $\mathbf{0.377} \pm \mathbf{0.742}$ &  $2.042 \pm 0.063$  & $0.594 \pm 0.819$ & $1.641 \pm 0.215$ & $0.406 \pm 1.899$\\
& 1 & $0.409 \pm 1.381$ & $0.415 \pm 0.805$ &  $2.268 \pm 0.070$  &$0.654 \pm 0.892$ & $1.802 \pm 0.215$ & $\mathbf{0.407} \pm \mathbf{1.900}$\\
\midrule 
\multicolumn{1}{c}{\multirow{6}{*}{MNIST}}
& 0.1 & $0.011 \pm 0.132$ & $\mathbf{0.009} \pm \mathbf{0.080}$ &  $0.226 \pm 0.005$  & $0.072 \pm 0.609$ & $0.198 \pm 0.065$ & $0.053 \pm 0.740$\\
& 0.2 & $0.020 \pm 0.147$ & $\mathbf{0.016} \pm \mathbf{0.098}$ & $0.452 \pm 0.011$   & $0.072 \pm 0.609$ & $0.359 \pm 0.065$ & $0.053 \pm 0.740$\\
& 0.3 & $0.030 \pm 0.170$ & $\mathbf{0.023} \pm \mathbf{0.123}$ &  $0.678 \pm 0.016$  & $0.073 \pm 0.610$ & $0.519 \pm 0.065$ & $0.053 \pm 0.740$\\
& 0.4 & $0.039 \pm 0.198$ & $\mathbf{0.029} \pm \mathbf{0.150}$ &  $0.905 \pm 0.021$  & $0.074 \pm 0.610$ & $0.680 \pm 0.064$ & $0.053 \pm 0.740$\\
& 0.5 & $0.048 \pm 0.229$ & $\mathbf{0.036} \pm \mathbf{0.179}$ & $1.131 \pm 0.027$   & $0.074 \pm 0.610$ & $0.841 \pm 0.064$ & $0.053 \pm 0.740$\\
& 0.6 & $0.057 \pm 0.261$ & $\mathbf{0.043} \pm \mathbf{0.208}$ & $1.357 \pm 0.032$   & $0.075 \pm 0.611$ & $1.002 \pm 0.064$ & $0.053 \pm 0.741$\\
& 0.7 & $0.066 \pm 0.295$ & $\mathbf{0.050} \pm \mathbf{0.239}$ & $1.583 \pm 0.037$   & $0.075 \pm 0.611$ & $1.163 \pm 0.064$ & $0.054 \pm 0.741$\\
& 0.8 & $0.075 \pm 0.330$ & $0.056 \pm 0.269$ & $1.809 \pm 0.043$   & $0.076 \pm 0.611$ & $1.324 \pm 0.064$ & $\mathbf{0.054} \pm \mathbf{0.741}$\\
& 0.9 & $0.084 \pm 0.366$ & $0.063 \pm 0.300$ &  $2.035 \pm 0.048$  & $0.076 \pm 0.612$ & $1.484 \pm 0.064$ & $\mathbf{0.054} \pm \mathbf{0.741}$\\
& 1 & $0.093 \pm 0.401$ & $0.070 \pm 0.331$ & $2.261 \pm 0.053$   & $0.077 \pm 0.612$ & $1.645 \pm 0.064$ & $\mathbf{0.054} \pm \mathbf{0.741}$\\
\midrule
\multicolumn{1}{c}{\multirow{6}{*}{CIFAR-100}}
& 0.1 & $0.381 \pm 0.514$ & $\mathbf{0.337} \pm \mathbf{0.299}$ &  $0.354 \pm 0.257$  & $0.929 \pm 0.582$ & - & $1.575 \pm 3.957$\\
& 0.2 & $\mathbf{0.616} \pm \mathbf{0.580}$ & $0.656 \pm 0.437$ &  $0.697 \pm 0.404$  & $1.134 \pm 0.536$ & - & $1.632 \pm 3.951$\\
& 0.3 & $\mathbf{0.850} \pm \mathbf{0.686}$ & $0.974 \pm 0.605$ & $1.041 \pm 0.573$   & $1.339 \pm 0.514$ & - & $1.689 \pm 3.948$\\
& 0.4 & $\mathbf{1.085} \pm \mathbf{0.818}$ & $1.292 \pm 0.784$ &  $1.384 \pm 0.749$ & $1.544 \pm 0.519$ & - & $1.746 \pm 3.949$\\
& 0.5 & $\mathbf{1.320} \pm \mathbf{0.964}$ & $1.610 \pm 0.967$ & $1.727 \pm 0.929$  &  $1.749 \pm 0.550$ & - & $1.803 \pm 3.953$\\
& 0.6 & $\mathbf{1.555} \pm \mathbf{1.119}$ & $1.928 \pm 1.153$ &  $2.071 \pm 1.110$  & $1.954 \pm 0.603$ & - & $1.860 \pm 3.961$\\
& 0.7 & $\mathbf{1.789} \pm \mathbf{1.280}$ & $2.247 \pm 1.340$ & $2.414 \pm 1.291$   & $2.159 \pm 0.673$ & - & $1.917 \pm 3.972$\\
& 0.8 & $2.024 \pm 1.444$ & $2.565 \pm 1.528$ & $2.758 \pm 1.474$   & $2.364 \pm 0.756$ & - & $\mathbf{1.974\mathbf} \pm \mathbf{3.986}$\\
& 0.9 & $2.259 \pm 1.612$ & $2.883 \pm 1.717$ & $3.101 \pm 1.657$   & $2.569 \pm 0.847$ & - & $\mathbf{2.031} \pm \mathbf{4.004}$\\
& 1 & $2.494 \pm 1.781$ & $3.201 \pm 1.906$ &  $3.445 \pm 1.840$  & $2.774 \pm 0.945$ & - & $\mathbf{2.088} \pm \mathbf{4.025}$\\
\bottomrule
\end{tabular}
} 
\end{table}

\begin{table}[!ht]
    \centering
    \caption{Model Selection Based on Evaluation Metric using KL Divergence and Non-Specificity on the MNIST dataset.}
\resizebox{\textwidth}{!}{
    \begin{tabular}{@{}cccc@{}}
        \toprule
        \textbf{Trade-off ($\lambda$)} & \textbf{Models} & \textbf{Evaluation ($\mathcal{E}$) Mean} \\ \midrule
        0.1 & DE, LB-BNN, RS-NN, CreINN, E-CNN, EDL & [0.009, 0.011, 0.053, 0.072, 0.198, 0.226] \\
        0.2 & DE, LB-BNN, RS-NN, CreINN, E-CNN, EDL & [0.016, 0.020, 0.053, 0.072, 0.359, 0.452] \\
        0.3 & DE, LB-BNN, RS-NN, CreINN, E-CNN, EDL & [0.023, 0.030, 0.053, 0.073, 0.519, 0.678] \\
        0.4 & DE, LB-BNN, RS-NN, CreINN, E-CNN, EDL & [0.029, 0.039, 0.053, 0.074, 0.680, 0.904] \\
        0.5 & DE, LB-BNN, RS-NN, CreINN, E-CNN, EDL & [0.036, 0.048, 0.053, 0.074, 0.841, 1.130] \\
        0.6 & DE, RS-NN, LB-BNN, CreINN, E-CNN, EDL & [0.043, 0.053, 0.057, 0.075, 1.002, 1.356] \\
        0.7 & DE, RS-NN, LB-BNN, CreINN, E-CNN, EDL & [0.050, 0.054, 0.066, 0.075, 1.163, 1.582] \\
        0.8 & RS-NN, DE, LB-BNN, CreINN, E-CNN, EDL & [0.054, 0.056, 0.075, 0.076, 1.324, 1.808] \\
        0.9 & RS-NN, DE, CreINN, LB-BNN, E-CNN, EDL & [0.054, 0.063, 0.076, 0.084, 1.484, 2.034] \\
        1.0 & RS-NN, DE, CreINN, LB-BNN, E-CNN, EDL & [0.054, 0.070, 0.077, 0.093, 1.645, 2.260] \\ \bottomrule
    \end{tabular}
    }
    \label{tab:model_selection_mnist}
\end{table}

\begin{table}[ht]
    \centering
    \small
    \caption{Model Selection Based on Evaluation Metric using KL Divergence and Non-Specificity on the CIFAR-100 dataset.}
    \resizebox{0.9\textwidth}{!}{
    \begin{tabular}{@{}cccc@{}}
        \toprule
        \textbf{Trade-off ($\lambda$)} & \textbf{Models} & \textbf{Evaluation ($\mathcal{E}$) Mean} \\ \midrule
        0.1 & DE, EDL, LB-BNN, CreINN, RS-NN &
[0.337, 0.354, 0.381, 0.929, 1.575] \\
        0.2 &  LB-BNN, DE, EDL, CreINN, RS-NN &
[0.616, 0.656, 0.697, 1.134, 1.632] \\
        0.3 & LB-BNN, DE, EDL, CreINN, RS-NN &
[0.850, 0.974, 1.041, 1.339, 1.689] \\
        0.4 & LB-BNN, DE, EDL, CreINN, RS-NN &
[1.085, 1.292, 1.384, 1.544, 1.746] \\
        0.5 & LB-BNN, DE, EDL, CreINN, RS-NN &
[1.320, 1.610, 1.727, 1.749, 1.803] \\
        0.6 & LB-BNN, RS-NN, DE, CreINN, EDL &
[1.555, 1.860, 1.928, 1.954, 2.071] \\
        0.7 & LB-BNN, RS-NN, CreINN, DE, EDL &
[1.789, 1.917, 2.159, 2.247, 2.414] \\
        0.8 & RS-NN, LB-BNN, CreINN, DE, EDL &
[1.974, 2.024, 2.364, 2.565, 2.758] \\
        0.9 & RS-NN, LB-BNN, CreINN, DE, EDL &
[2.031, 2.259, 2.569, 2.883, 3.101] \\
        1.0 & RS-NN, LB-BNN, CreINN, DE, EDL &
[2.088, 2.494, 2.774, 3.201, 3.445] \\ \bottomrule
    \end{tabular}
    }
    \label{tab:model_selection_cifar100}
\end{table}

Among the models tested, the LB-BNN consistently achieves the best performance across almost all $\lambda$ values, indicative of its robust ability to manage the trade-off between specificity and non-specificity. The model's performance surpasses others, particularly in terms of lower variability, which is evident as it maintains tighter standard deviations relative to its means. 
As $\lambda$ approaches $1$, RS-NN again is selected as the best model as shown in Tabs. \ref{tab:lambda-1} and \ref{tab:model_selection_cifar100}.

The trends observed in the CIFAR-100 results further reinforce the need for careful tuning of $\lambda$ to optimize model performance. As $\lambda$ increases, the complexity of classification tasks also rises, necessitating models that can adaptively learn from both accurate and non-specific features.


\subsection{Ablation on number of prediction samples}
\label{sec:exp_ablation_samples}

In the ablation study on the number of prediction samples of LB-BNN and the number of ensembles of DE with Evaluation Metric ($\mathcal{E}$) shown in Fig. \ref{fig:ablation_samples}, we note that as the number of samples increases, the value of $\mathcal{E}$ also increases, with a more pronounced effect seen in DE compared to LB-BNN. Results are shown for LB-BNN with 50 to 500 samples and DE with 5 to 30 ensembles. The increase in the number of samples leads to a corresponding increase in the size of the credal set, 
without apparently being compensated by lower KL divergence values.

\begin{figure}[!h]
    \centering
    \includegraphics[width = 0.9\linewidth]{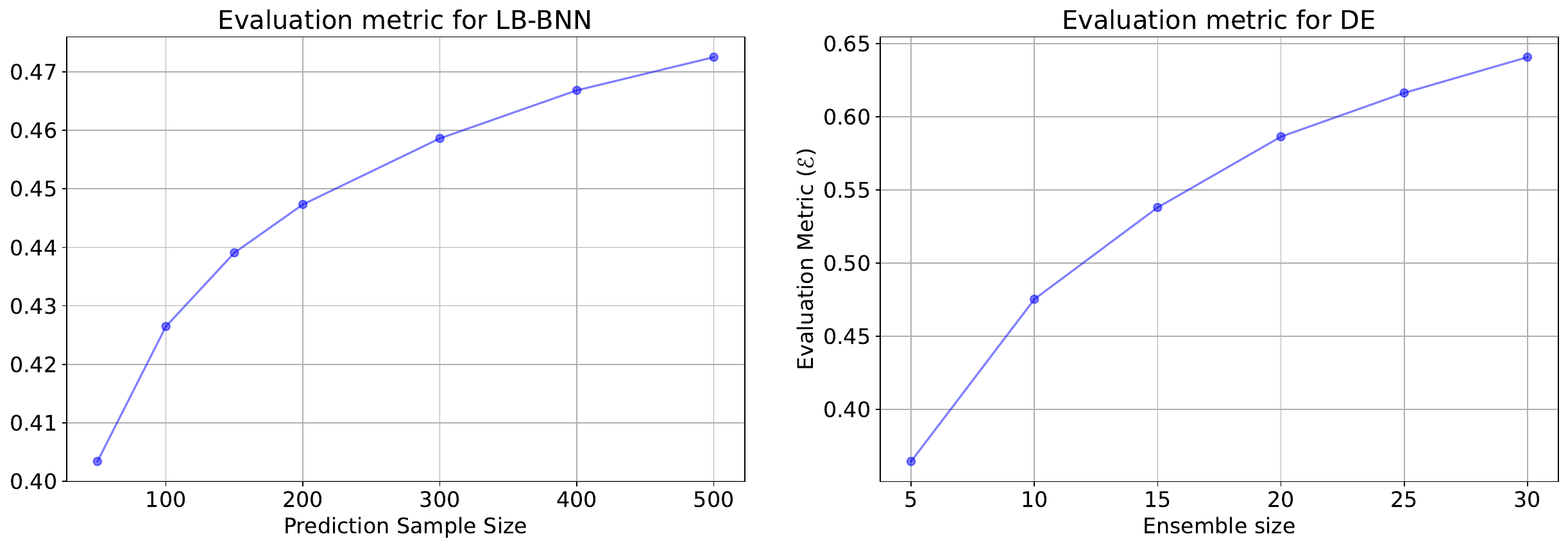} 
    \caption{Ablation study on the number of prediction samples of LB-BNN and the number of ensembles of DE with Evaluation Metric ($\mathcal{E}$). 
    } 
    \label{fig:ablation_samples}
\end{figure}

\subsubsection{Computation of the credal set vertices} \label{app:credal_set_size}

Fig. \ref{fig:credal_set_size} shows the sizes of the credal sets predicted by all models, for 50 prediction samples of the CIFAR-10 dataset. This is obtained by taking the difference between maximal and minimal extremal probability Eq. \ref{eq:prho} for the predicted class.  
\begin{figure}[!ht]
    \centering
    \includegraphics[width=0.9\textwidth]{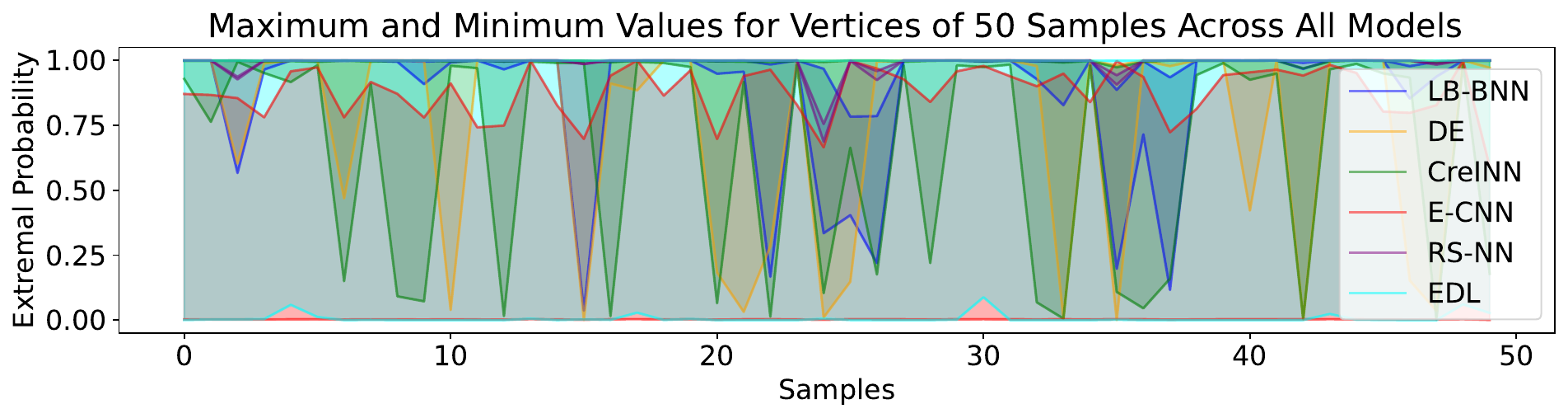}
    \caption{Credal set sizes for all models for 50 prediction samples of the CIFAR-10 dataset. Larger credal set sizes indicate a more imprecise prediction. 
    }
    \label{fig:credal_set_size}
\end{figure}
\begin{figure}[!h]
    \centering
    \includegraphics[width=0.9\textwidth]{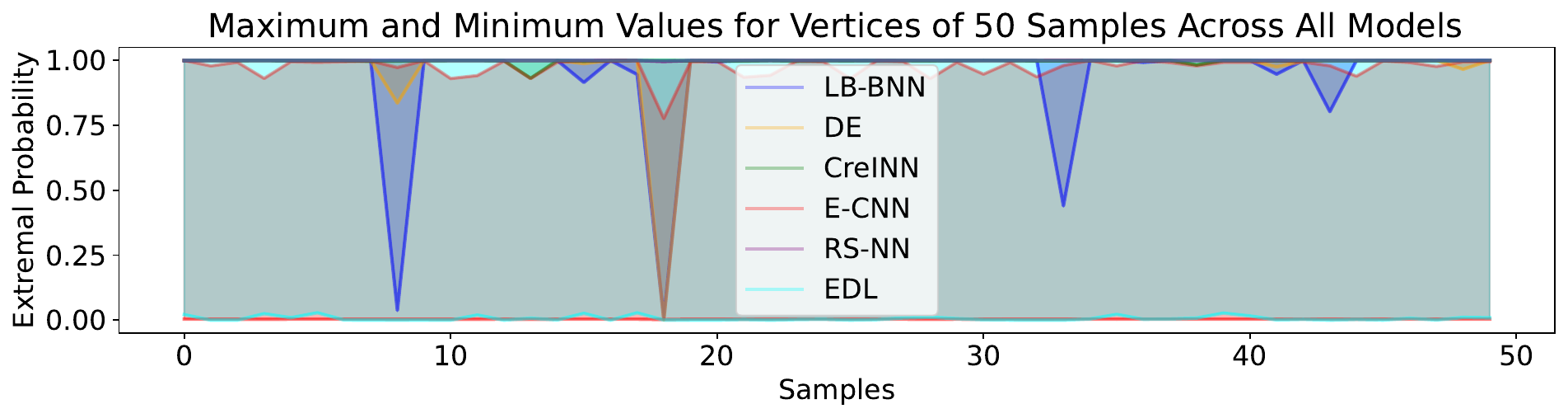}
    \caption{Credal set sizes for all models for 50 prediction samples of the MNIST dataset. Larger credal set sizes indicate a more imprecise prediction}
    \label{fig:app_credal_set_size_mnist}
\end{figure}
\begin{figure}[!htbp]
    \centering
    \includegraphics[width=0.9\textwidth]{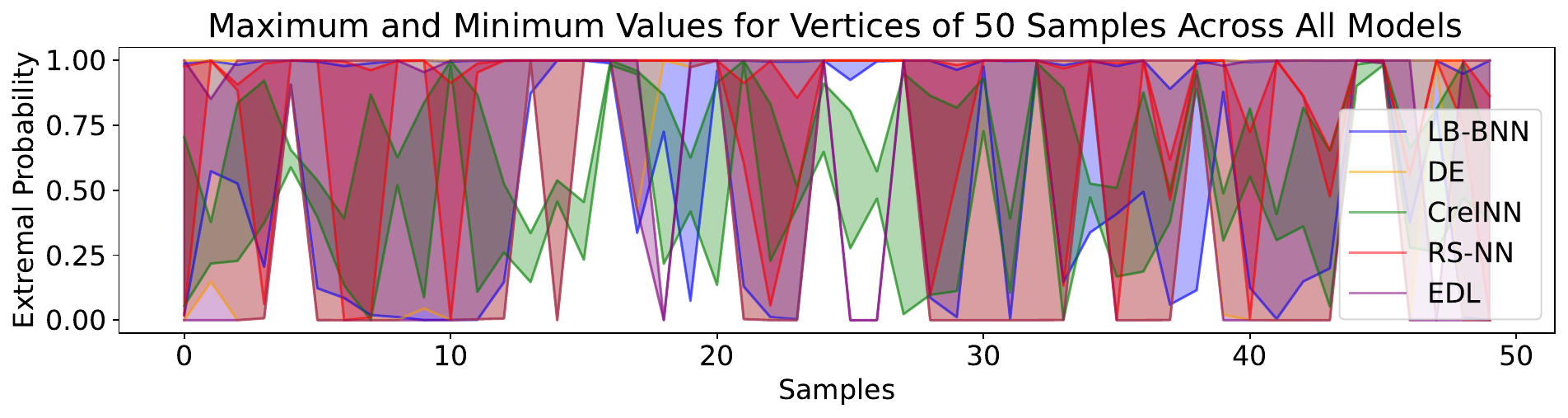}
    \caption{Credal set sizes for all models for 50 prediction samples of the CIFAR-100 dataset. Larger credal set sizes indicate a more imprecise prediction}
    \label{fig:app_credal_set_size_cifar100}
\end{figure}

Similarly, Figs. \ref{fig:app_credal_set_size_mnist} and \ref{fig:app_credal_set_size_cifar100} demonstrate the sizes of the credal
sets of the MNIST and CIFAR-100 datasets, respectively.
A larger credal set size signifies that the model is highly uncertain about its predictions. E-CNN exhibits the widest credal set for all predictions, indicating 
significant imprecision
regardless of whether the model predicts the correct or incorrect class.  This is undesirable, as it suggests a general lack of confidence in the predicted outcomes. 
Models with smaller credal set sizes demonstrate higher confidence in their predictions, implying more reliable and precise outcomes. 
Lower non-specificity values align with smaller credal widths, indicating smaller (epistemic) uncertainty and greater model confidence. 

\subsection{Distance measure vs Confidence score}

The choice of using distance instead of confidence in the proposed metric is motivated by the goal to capture the true nature of the prediction within the simplex and to properly characterize the credal set.

A key issue with using confidence (the maximum predicted probability) is that many models, particularly standard neural networks, exhibit a problem of overconfidence even for wrong predictions. In contrast, models like Deep Ensembles are designed to mitigate overconfidence by averaging predictions across different models. By using a distance metric, we can better capture the spread or diversity in the predicted probabilities across the output space instead of focusing solely on the confidence of predictions.

While predictive entropy is a well-known and effective method for quantifying total uncertainty, it is not useful here as we are mainly concerned with epistemic uncertainty. For instance, if infinite amounts of data were available, epistemic uncertainty (and credal set size) would reduce to zero, whereas entropy will not necessarily be zero unless we have a fully confident prediction (100\% confidence).

Furthermore, as illustrated in Fig. \ref{fig:kl_vs_confidence} below, entropy is closely tied to confidence, while non-specificity has little correlation with confidence. High confidence predictions lead to low entropy, while low confidence predictions result in higher entropy. However, when considering epistemic uncertainty, entropy alone may not provide a complete picture. It is also detailed in \citet{song2018evidence} (Sec. 3, page 5) that the entropy-based uncertainty measure mainly depends on the total distribution of probabilities without concern about the largest probability, whereas the non-specificity is related to the difference between the largest probability and other ones.

\begin{figure}[!ht]
    \centering
    \begin{minipage}[t]{0.3\textwidth}
    \includegraphics[width=\textwidth]{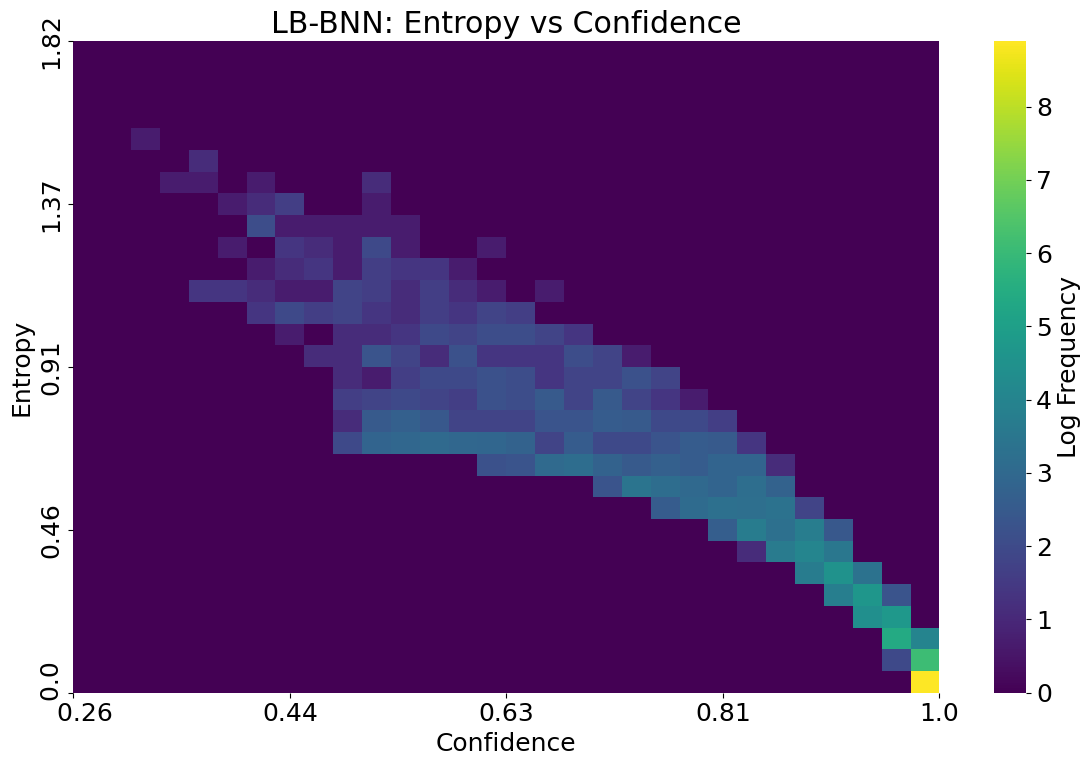}
    \end{minipage} \hspace{0.01\textwidth}
    \begin{minipage}[t]{0.3\textwidth}
    \includegraphics[width=\textwidth]{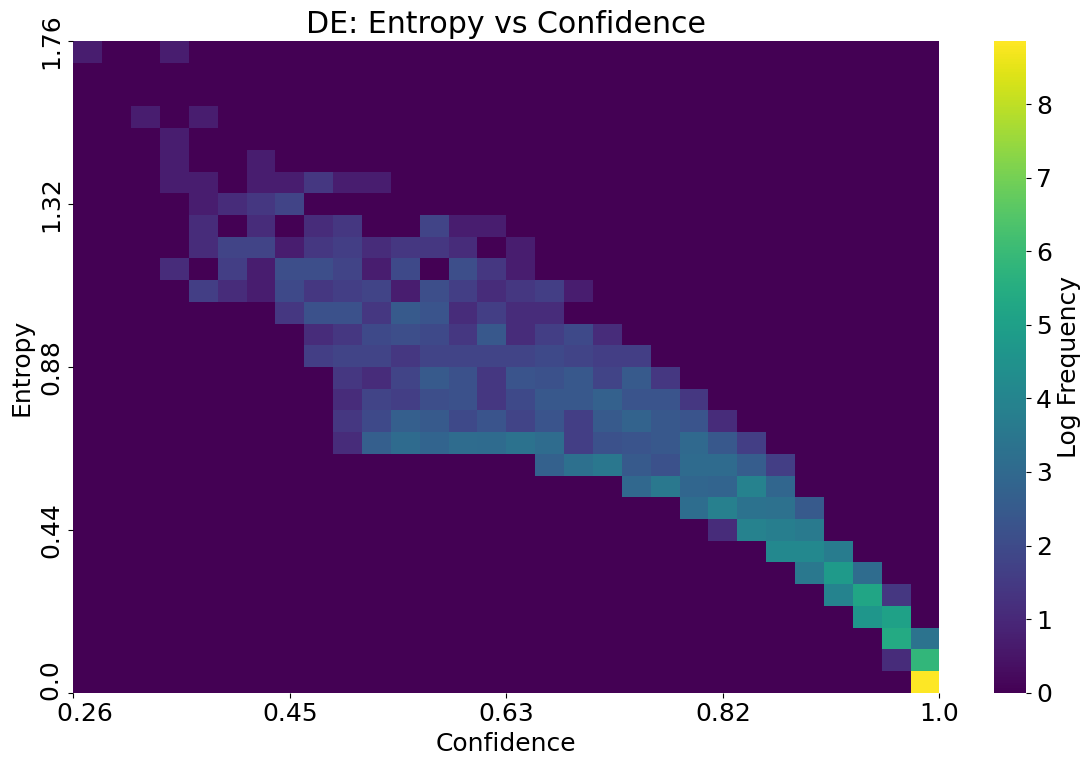}
    \end{minipage} \hspace{0.01\textwidth}
        \begin{minipage}[t]{0.3\textwidth}
    \includegraphics[width=\textwidth]{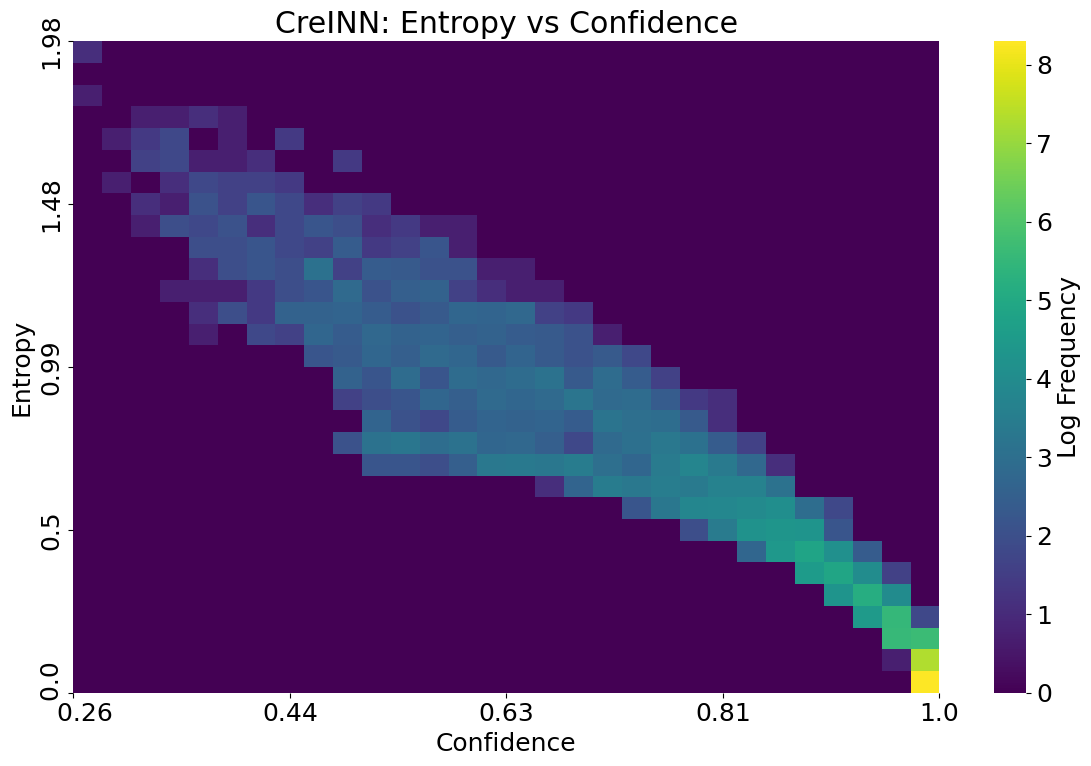}
    \end{minipage} \hspace{0.01\textwidth}
\\
    \begin{minipage}[t]{0.3\textwidth}
    \includegraphics[width=\textwidth]{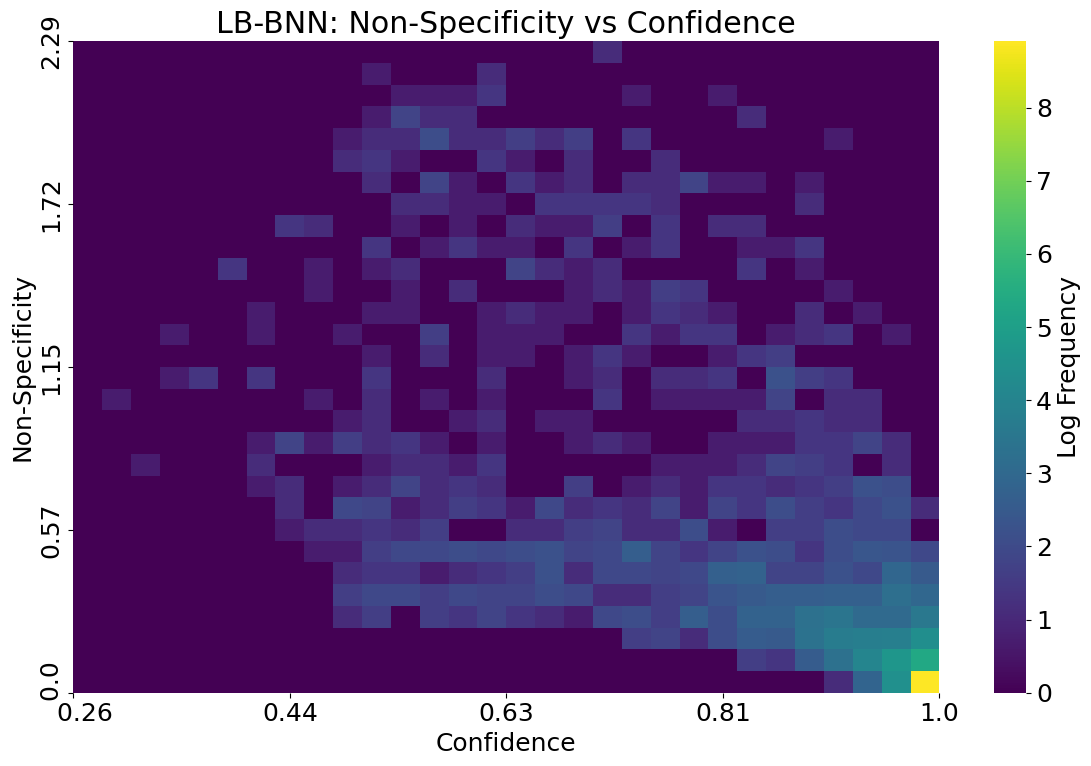}
        \caption*{(a) LB-BNN}
    \end{minipage} \hspace{0.01\textwidth}
        \begin{minipage}[t]{0.3\textwidth}
    \includegraphics[width=\textwidth]{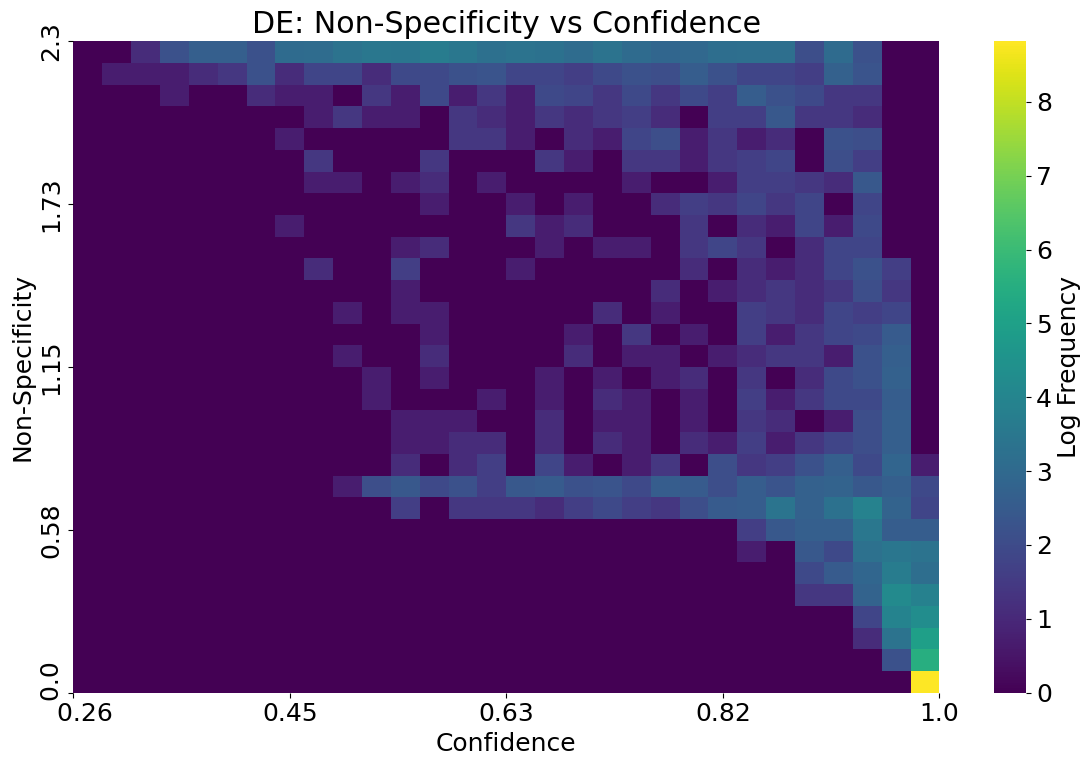}
            \caption*{(b) DE}
    \end{minipage}\hspace{0.01\textwidth}
        \begin{minipage}[t]{0.3\textwidth}
    \includegraphics[width=\textwidth]{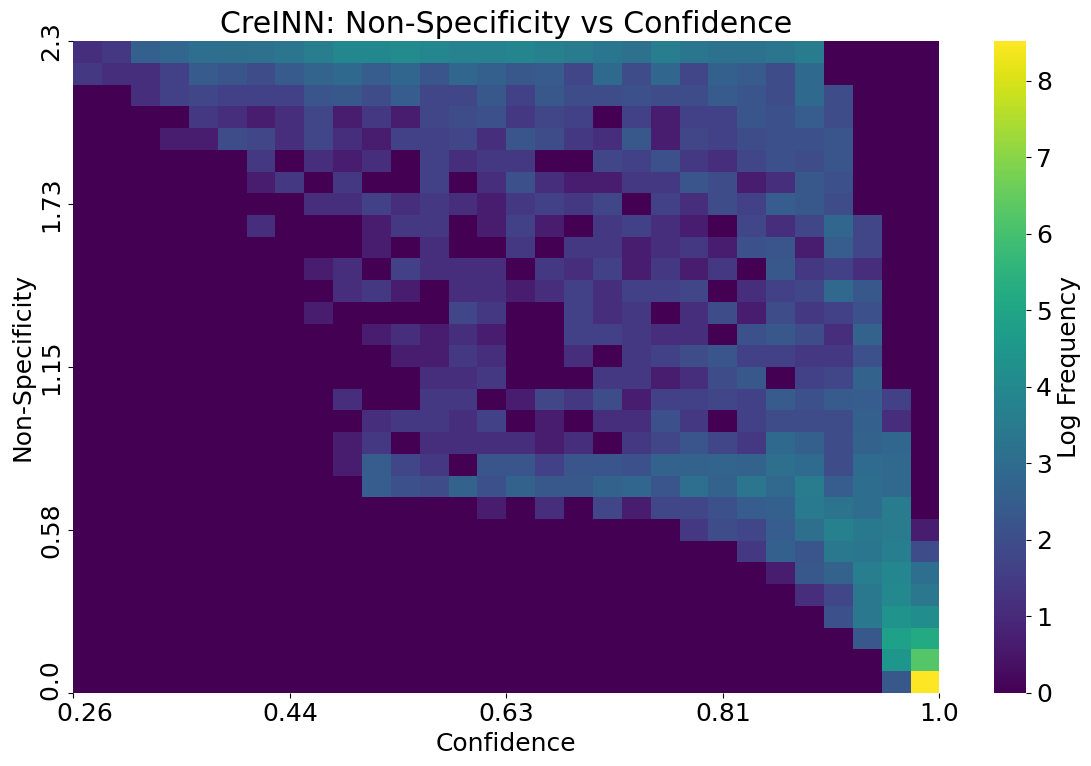}
            \caption*{(c) CreINN}
    \end{minipage}
    \\ 
        \begin{minipage}[t]{0.3\textwidth}
    \includegraphics[width=\textwidth]{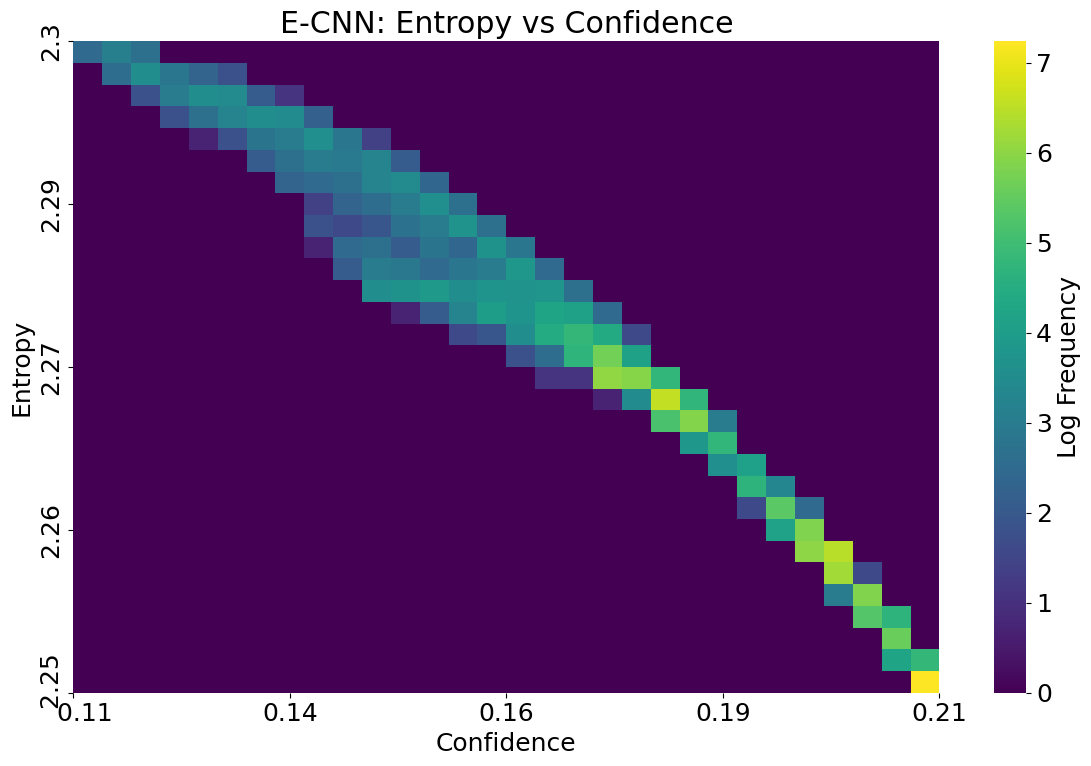}
    \end{minipage} \hspace{0.01\textwidth}
            \begin{minipage}[t]{0.3\textwidth}
    \includegraphics[width=\textwidth]{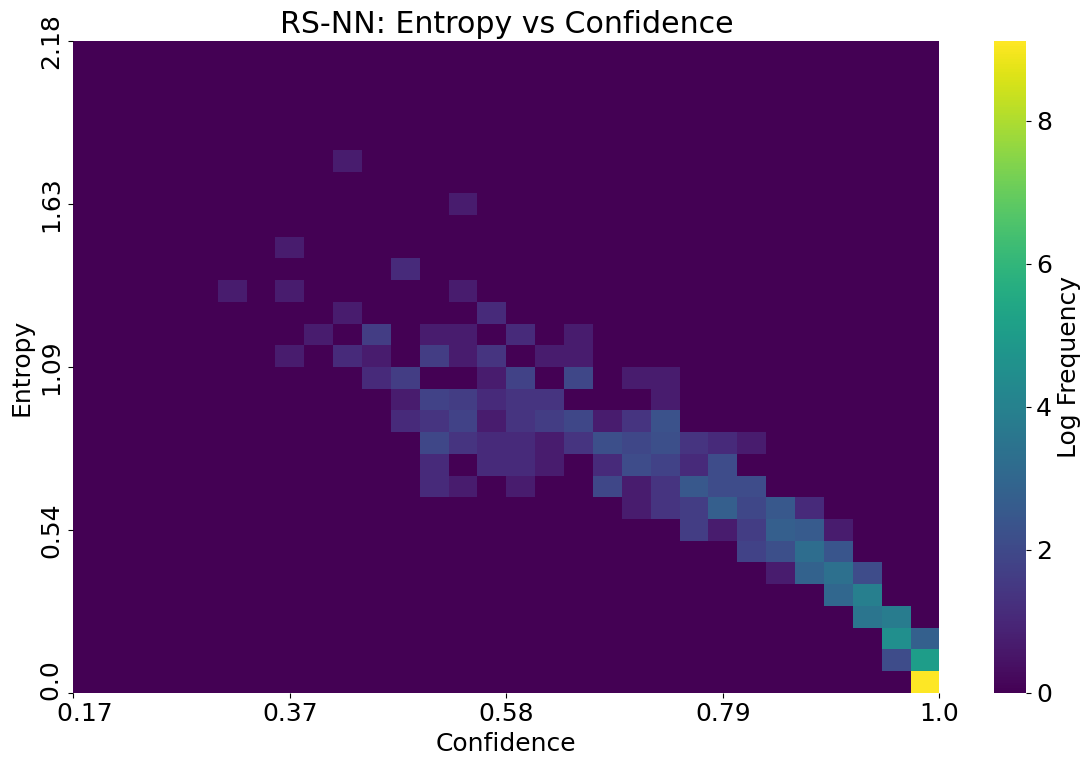}
    \end{minipage} \hspace{0.01\textwidth}
            \begin{minipage}[t]{0.3\textwidth}
    \includegraphics[width=\textwidth]{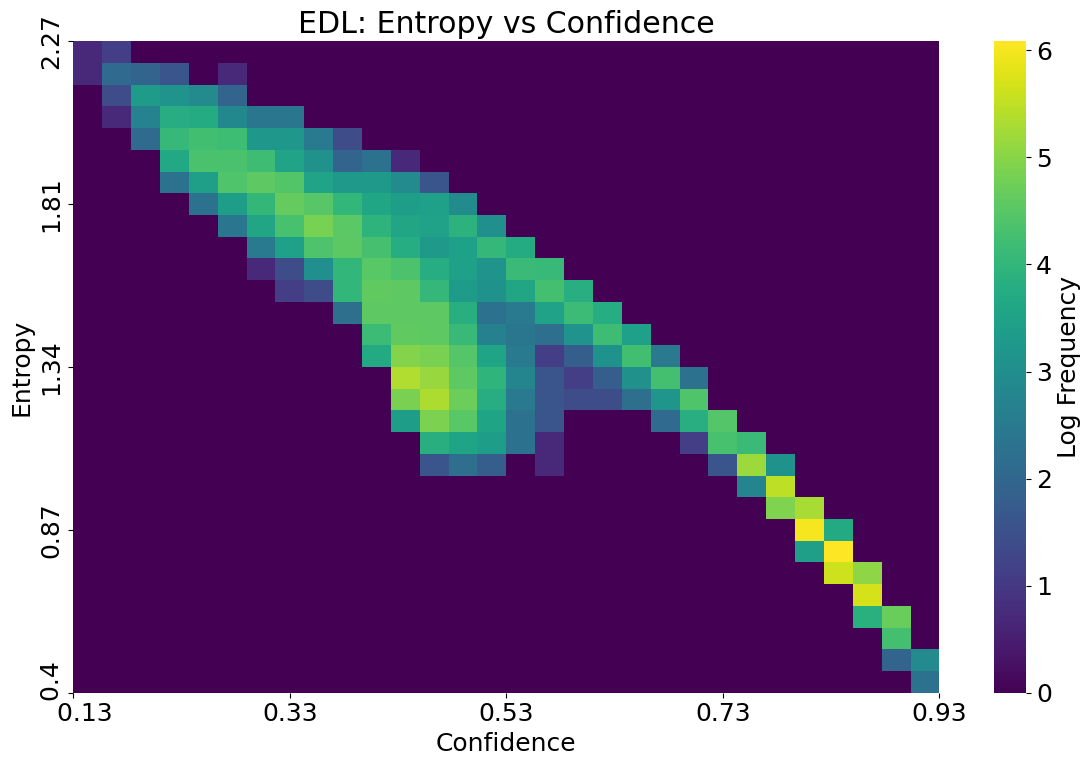}
    \end{minipage} 
    \\
        \begin{minipage}[t]{0.3\textwidth}
    \includegraphics[width=\textwidth]{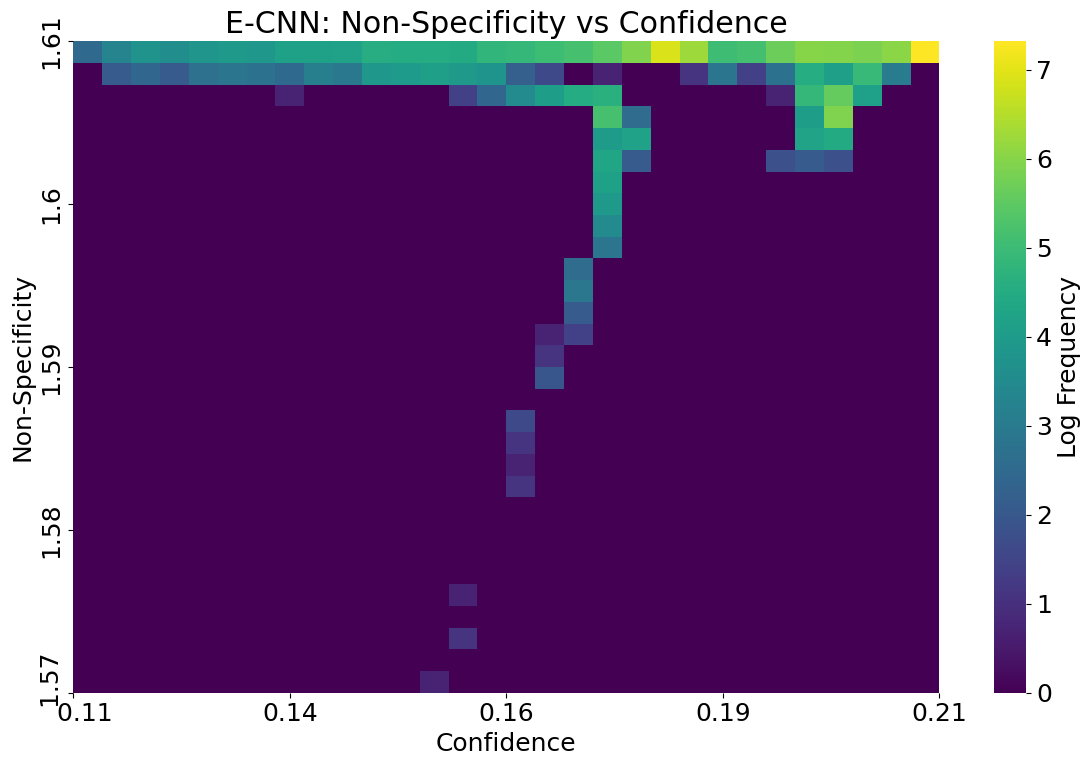}
    \caption*{(d) E-CNN}
    \end{minipage}\hspace{0.01\textwidth}
        \begin{minipage}[t]{0.3\textwidth}
    \includegraphics[width=\textwidth]{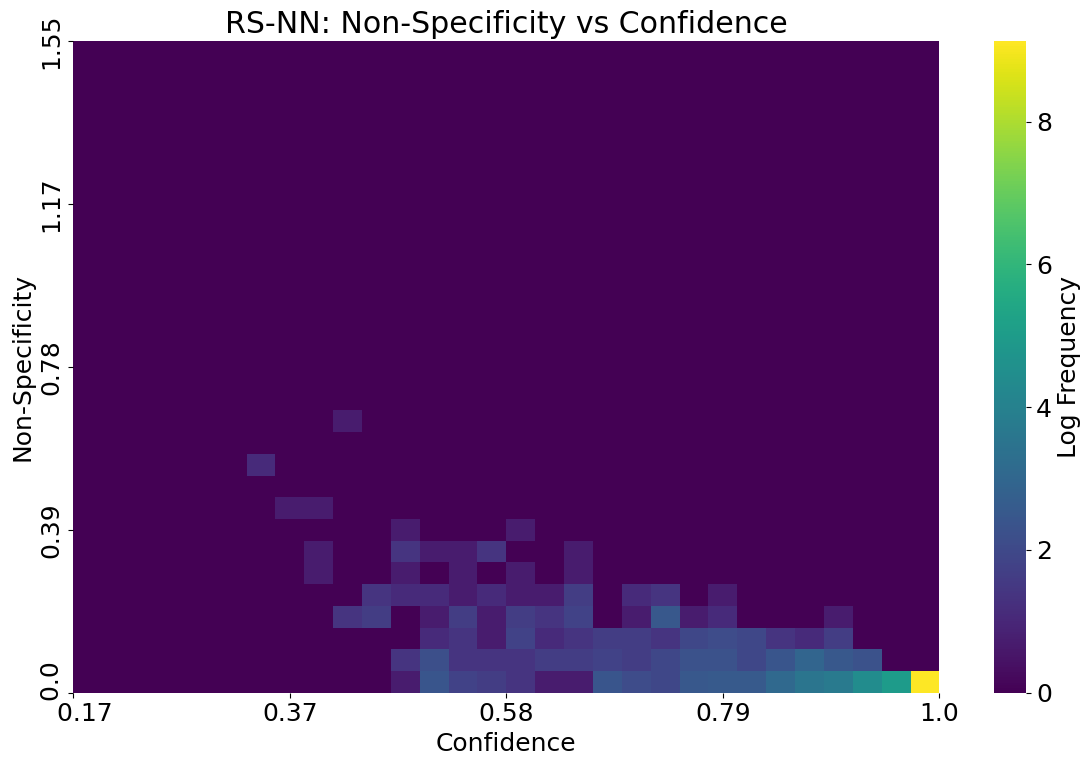}
    \caption*{(e) RS-NN}
    \end{minipage}\hspace{0.01\textwidth}
        \begin{minipage}[t]{0.3\textwidth}
    \includegraphics[width=\textwidth]{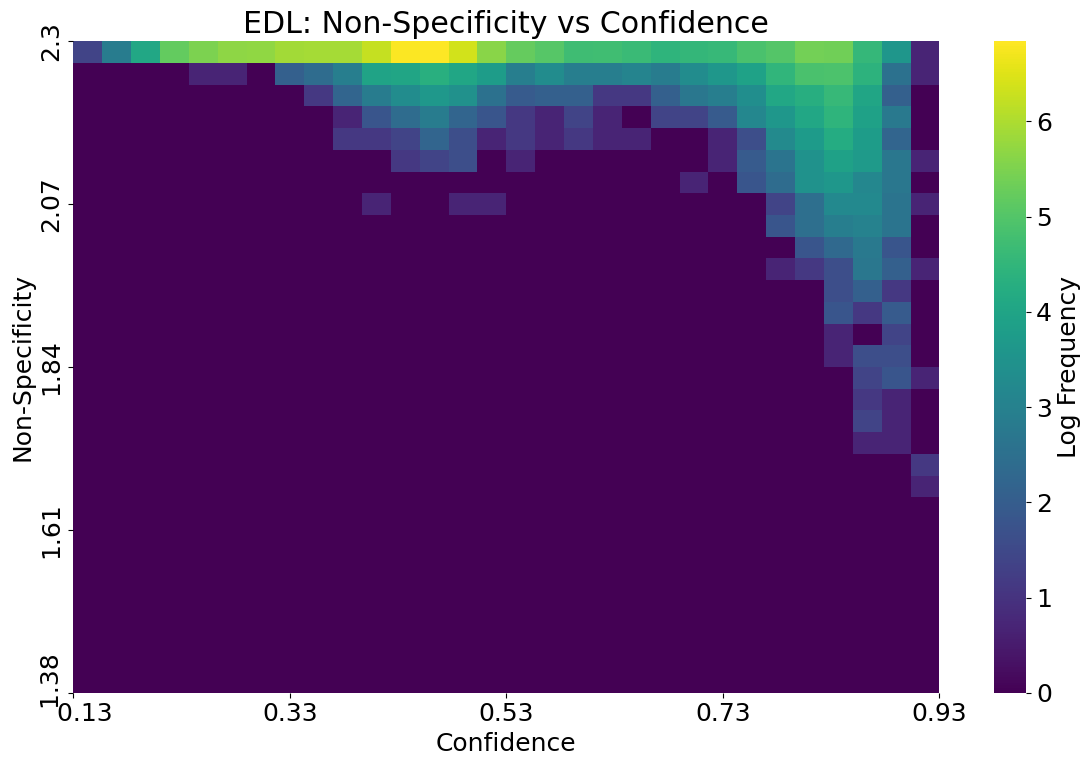}
    \caption*{(f) EDL}
    \end{minipage}
    \caption{Entropy vs Confidence (top) and Non-Specificity vs Confidence (bottom) for each of the models: (a) LB-BNN, (b) Deep Ensembles (DE), (c) CreINN, (d) E-CNN, (e) RS-NN and (f) EDL on the CIFAR-10 dataset. Entropy is strongly linked to confidence, whereas non-specificity shows minimal correlation with it.
}\label{fig:kl_vs_confidence}
\end{figure}

\subsection{Credal set size vs. Non-specificity} \label{app:credal_vs_non_specificity}

Figs. \ref{fig:app_credal_vs_non_specificity_cifar10}, \ref{fig:app_credal_vs_non_specificity_MNIST} and \ref{fig:app_credal_vs_non_specificity_cifar100} exhibits the correlation between credal set size and non-specificity for CIFAR-10, MNIST and CIFAR-100 respectively. Here, credal set size is computed by taking the average of the difference between maximal and minimal extremal probability Eq. \ref{eq:prho} for each class, and non-specificity is obtained using Eq. \ref{eq:non_spec}. There exists a direct correlation between the credal set size and non-specificity. This demonstrates the efficacy of using the non-specificity measure in the evaluation metric as it captures the uncertainty associated with model's prediction making the metric more wholesome. E-CNN produces a very wide credal set over all samples visualized in MNIST dataset indicating that it's a very imprecise model. 

\begin{figure}[!ht]
    \centering
    \begin{minipage}[t]{0.3\textwidth}
    \includegraphics[width=\textwidth]{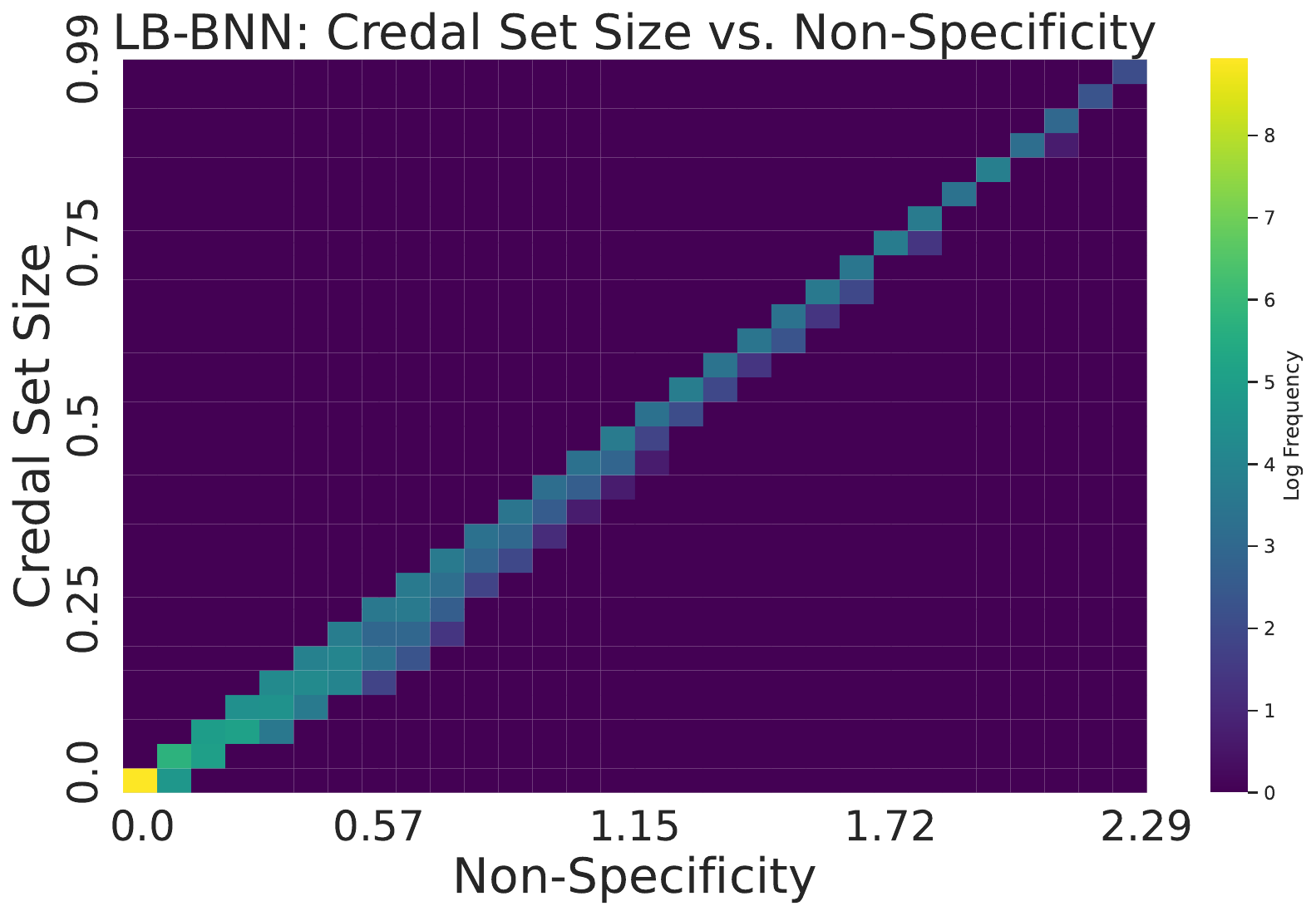}
    \caption*{(a)}
    \end{minipage} \hspace{0.01\textwidth}
    \begin{minipage}[t]{0.3\textwidth}
    \includegraphics[width=\textwidth]{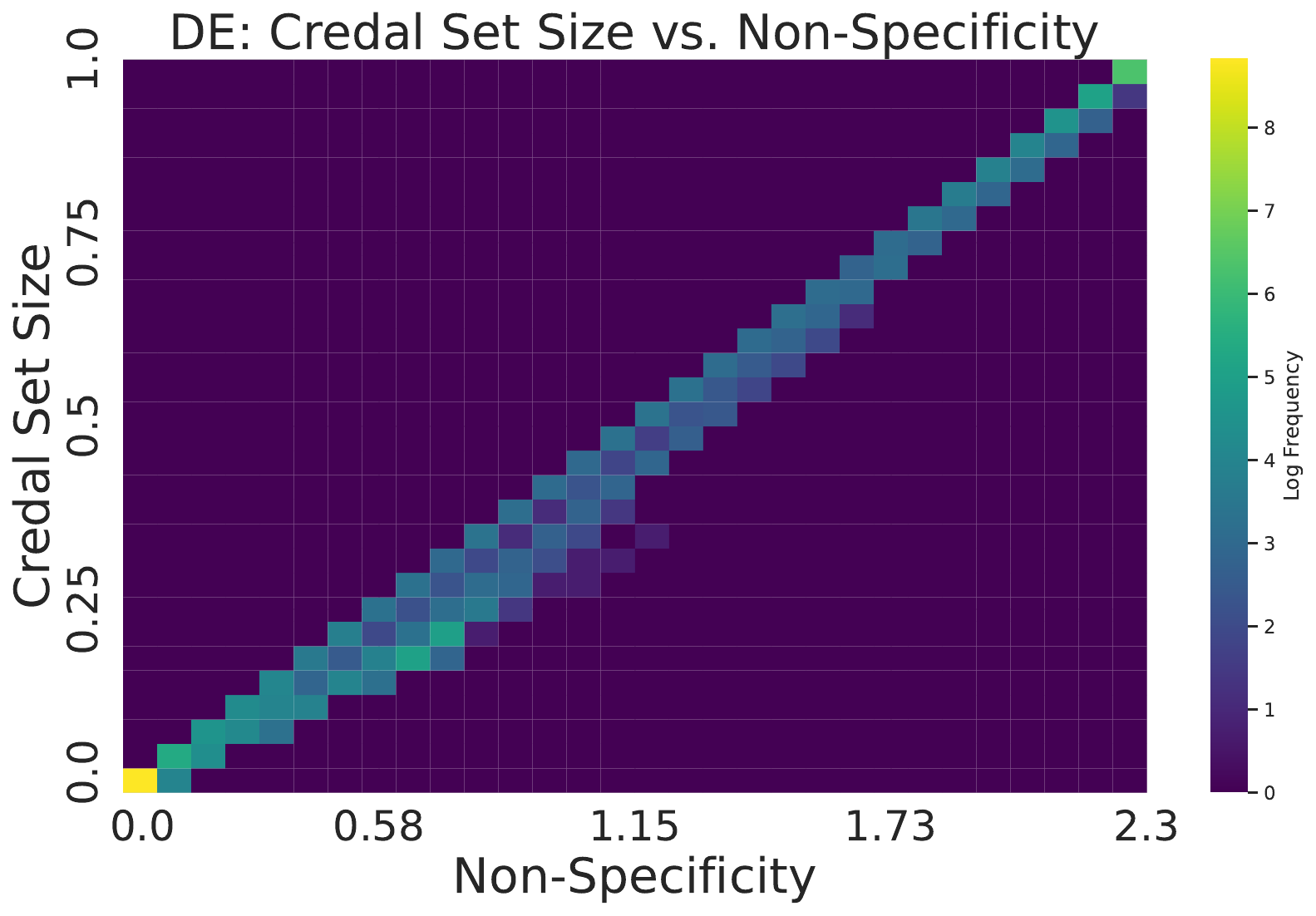}
    \caption*{(b)}
    \end{minipage} \hspace{0.01\textwidth}
        \begin{minipage}[t]{0.3\textwidth}
    \includegraphics[width=\textwidth]{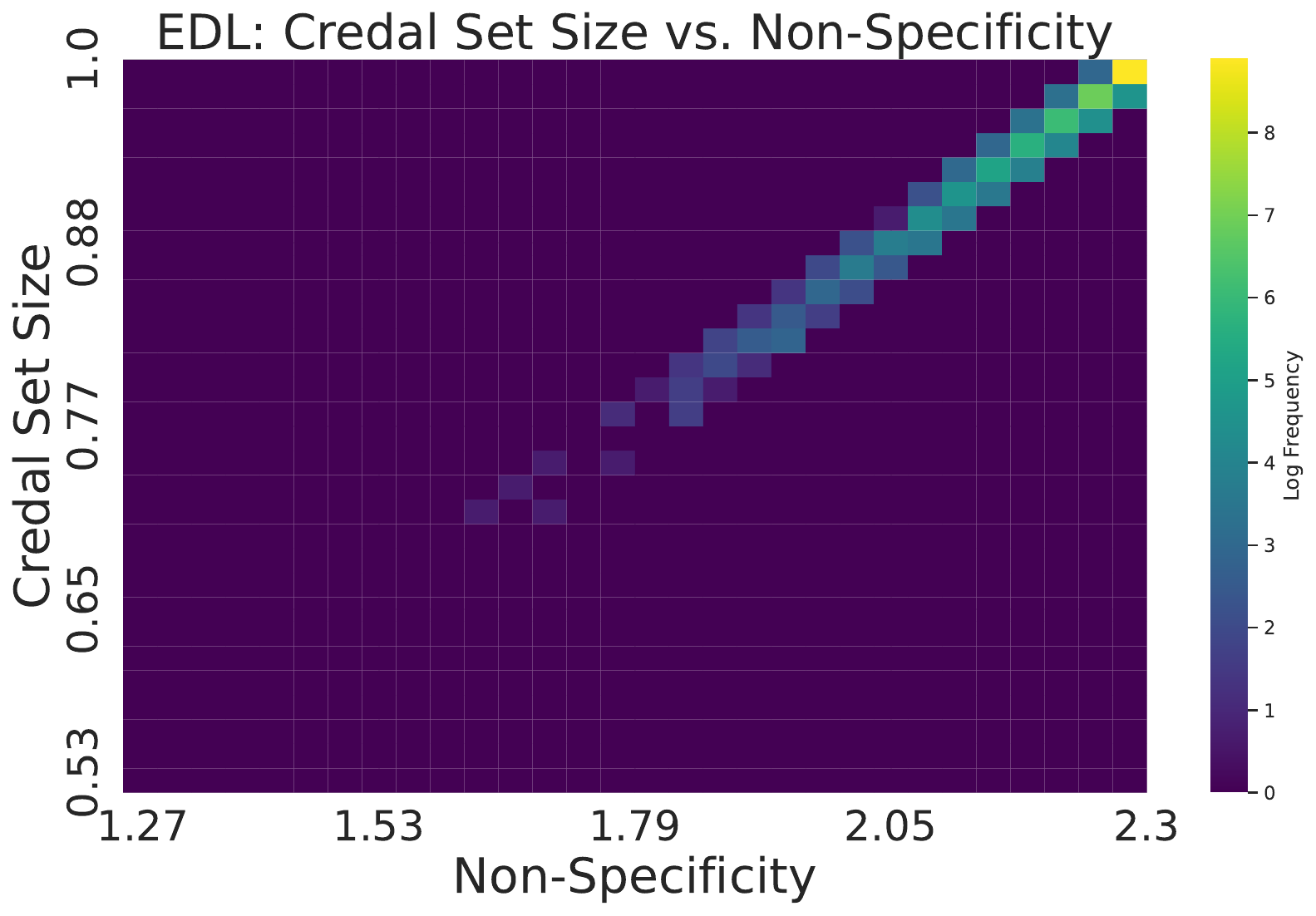}
    \caption*{(c)}
    \end{minipage}
    \\
    \begin{minipage}[t]{0.3\textwidth}
    \includegraphics[width=\textwidth]{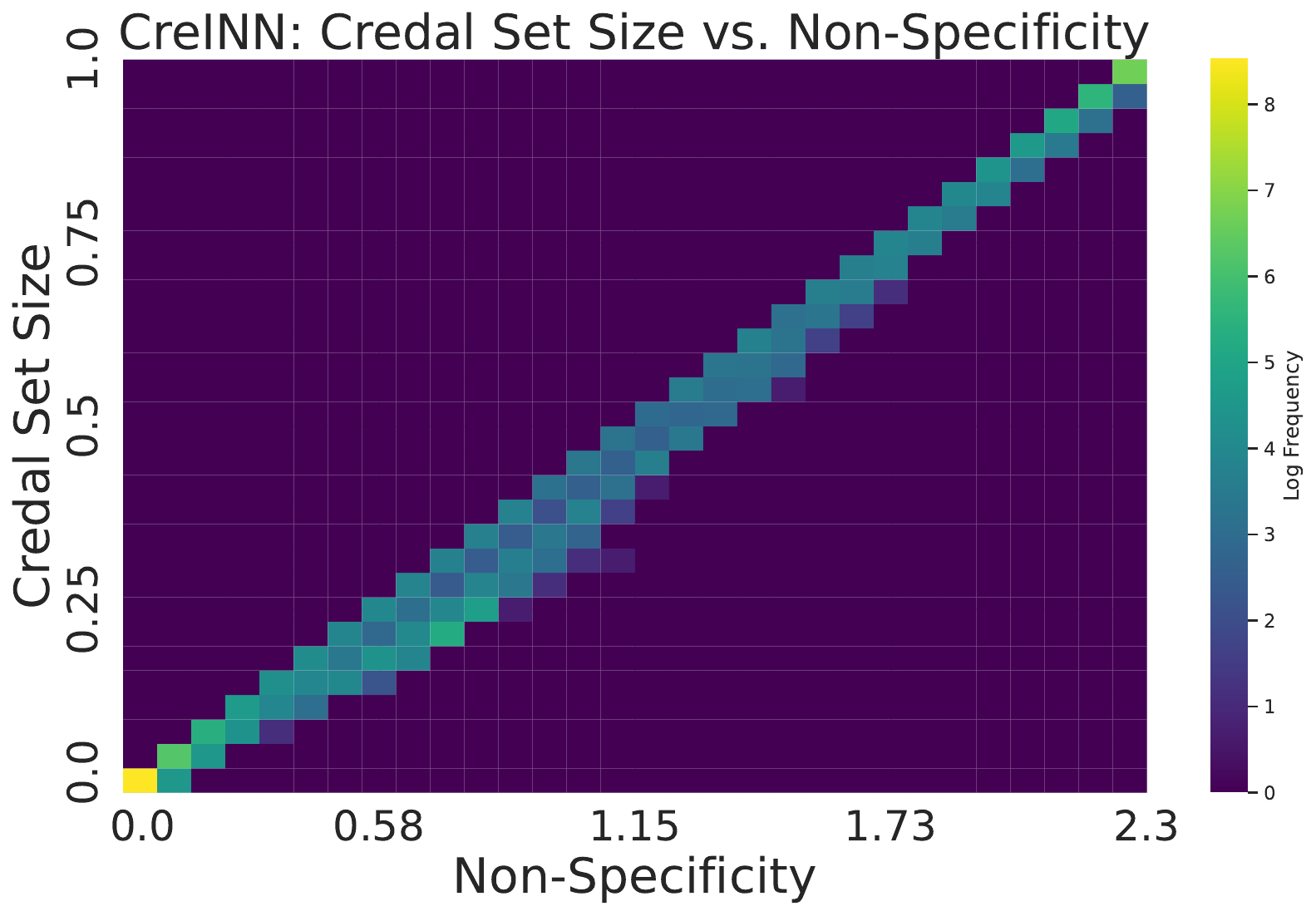}
    \caption*{(d)}
    \end{minipage} \hspace{0.01\textwidth}
        \begin{minipage}[t]{0.3\textwidth}
    \includegraphics[width=\textwidth]{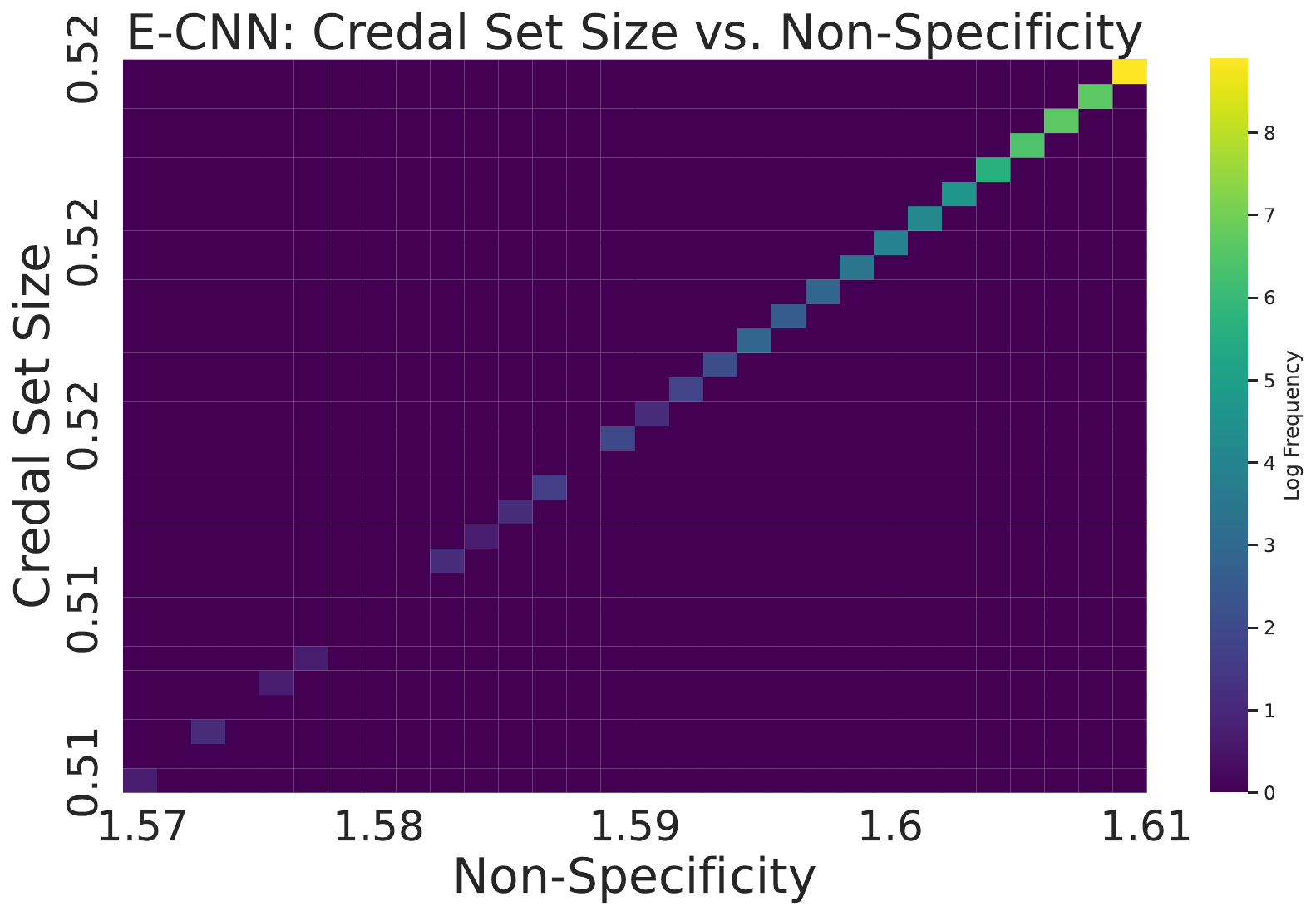}
    \caption*{(e)}
    \end{minipage}\hspace{0.01\textwidth}
            \begin{minipage}[t]{0.3\textwidth}
    \includegraphics[width=\textwidth]{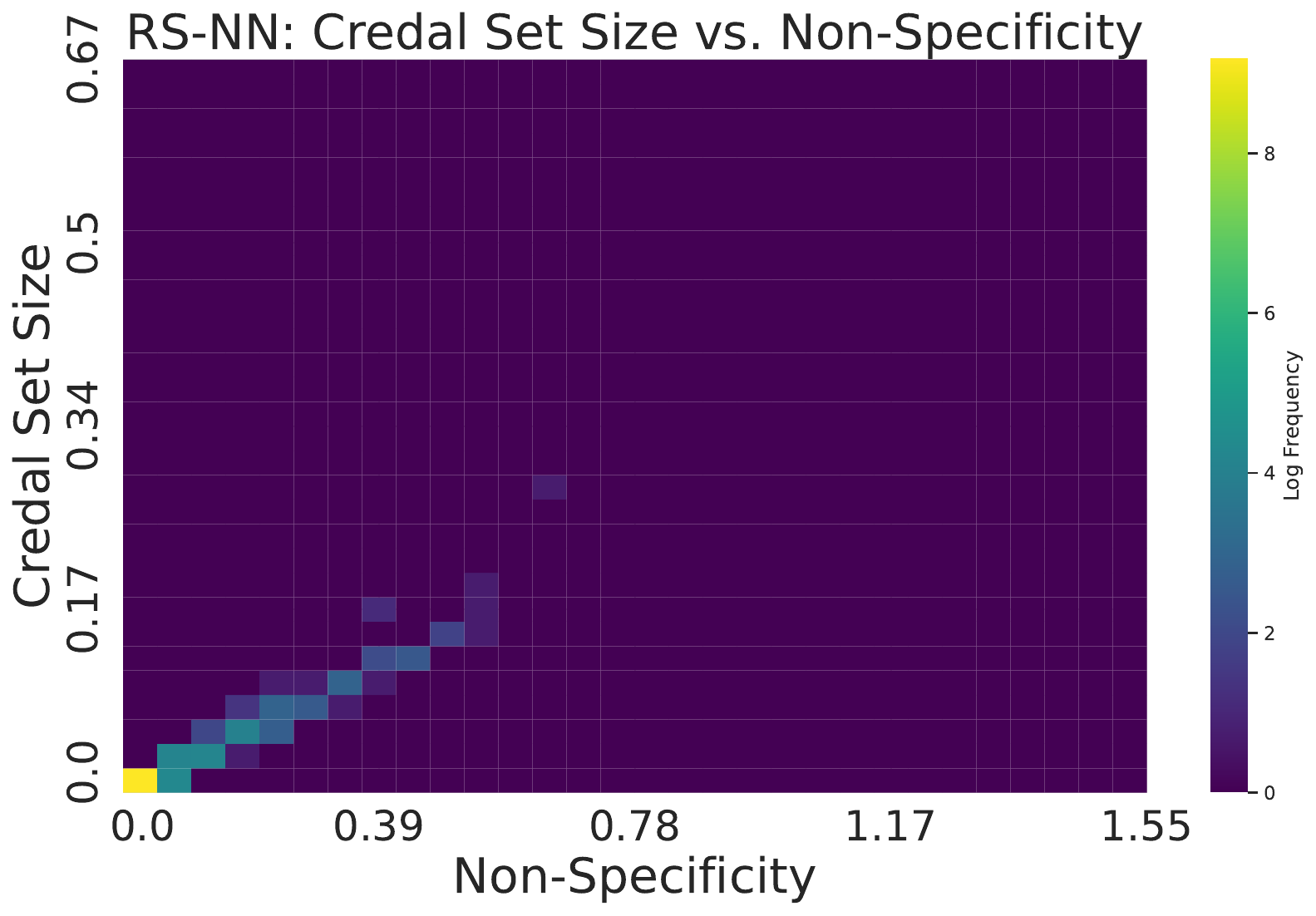}
    \caption*{(f)}
    \end{minipage}
    \caption{Credal Set Size vs. Non-Specificity heatmap for (a) LB-BNN, (b) Deep Ensembles (DE), (c) EDL, (d) CreINN, (e) E-CNN and (f) RS-NN on the CIFAR-10 dataset. Credal Set Size and Non-Specificity are directly correlated to each other. Log frequency is used to better showcase the trend.
}\label{fig:app_credal_vs_non_specificity_cifar10}
\end{figure}

\begin{figure}[!h]
    \centering
    \begin{minipage}[t]{0.3\textwidth}
    \includegraphics[width=\textwidth]{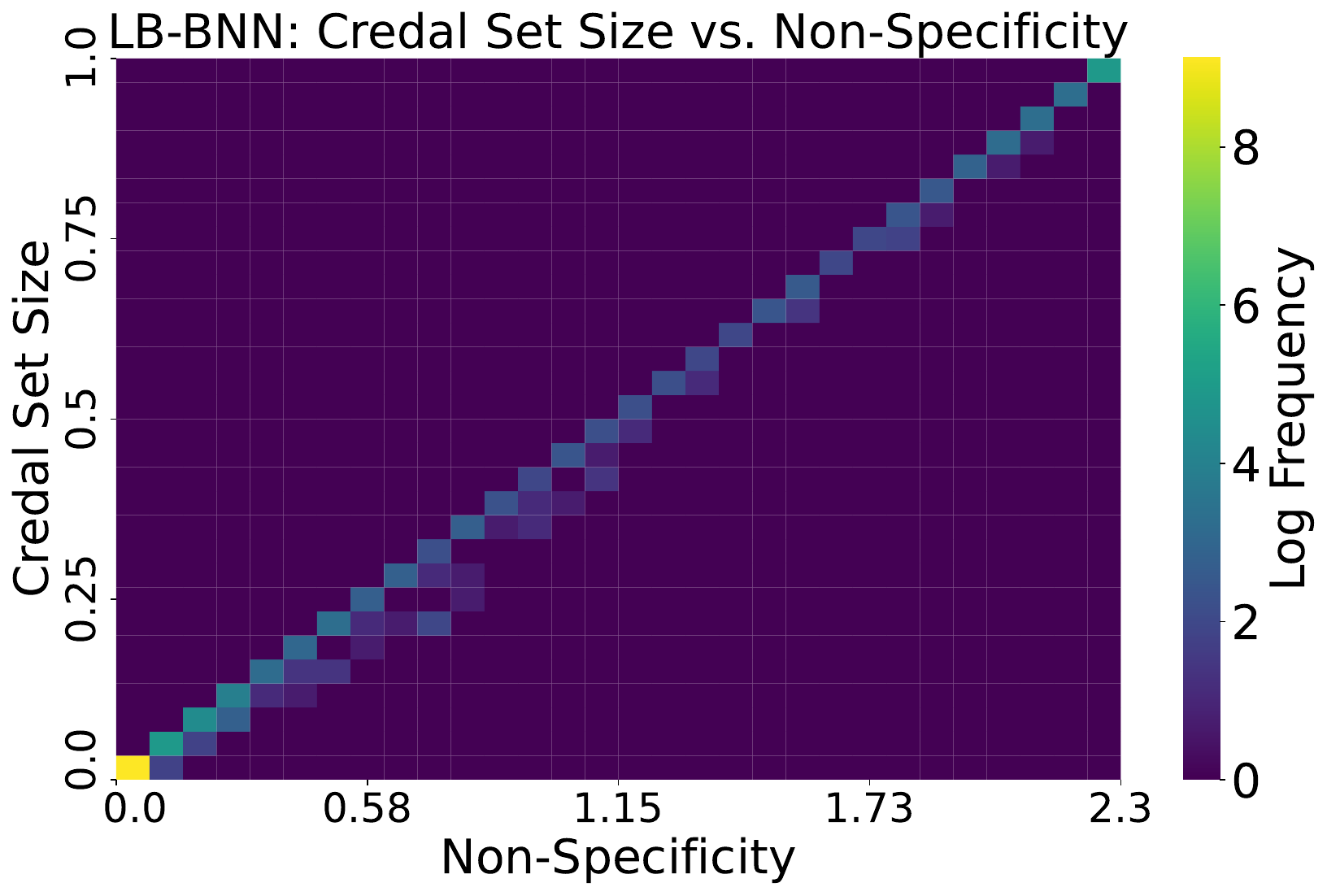}
    \caption*{(a)}
    \end{minipage} \hspace{0.01\textwidth}
    \begin{minipage}[t]{0.3\textwidth}
    \includegraphics[width=\textwidth]{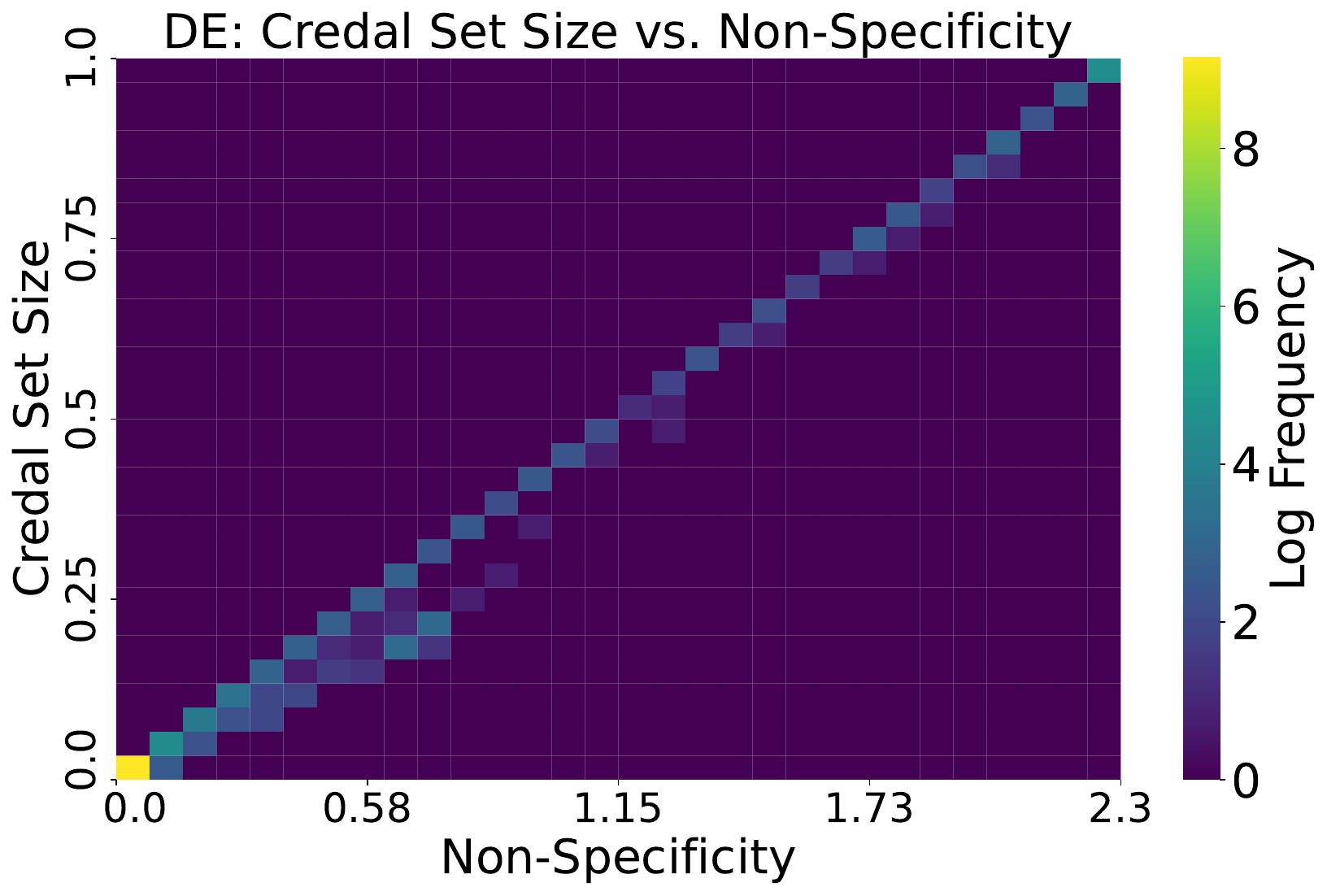}
    \caption*{(b)}
    \end{minipage} \hspace{0.01\textwidth}
        \begin{minipage}[t]{0.3\textwidth}
    \includegraphics[width=\textwidth]{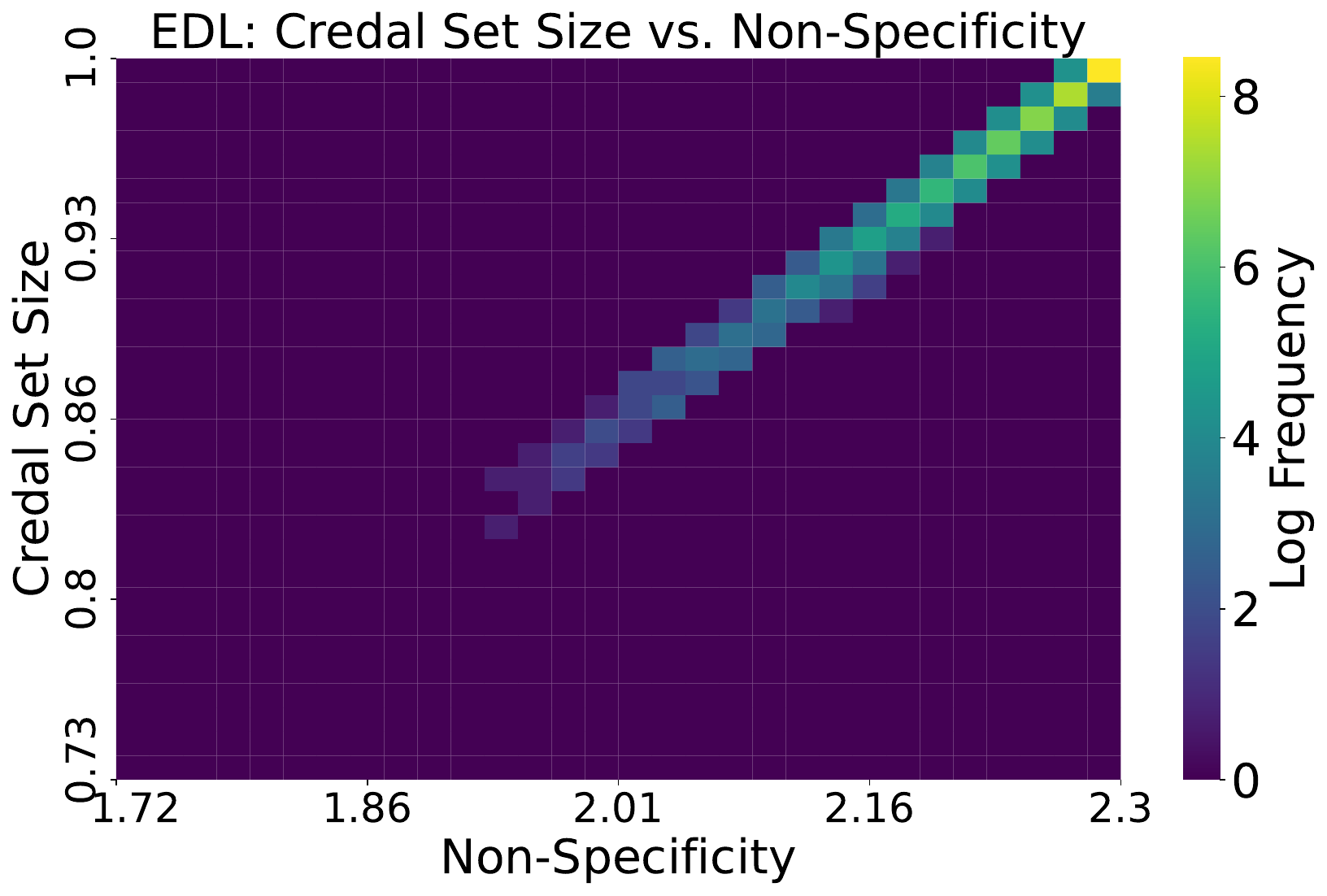}
    \caption*{(c)}
    \end{minipage}
    \\
    \begin{minipage}[t]{0.3\textwidth}
    \includegraphics[width=\textwidth]{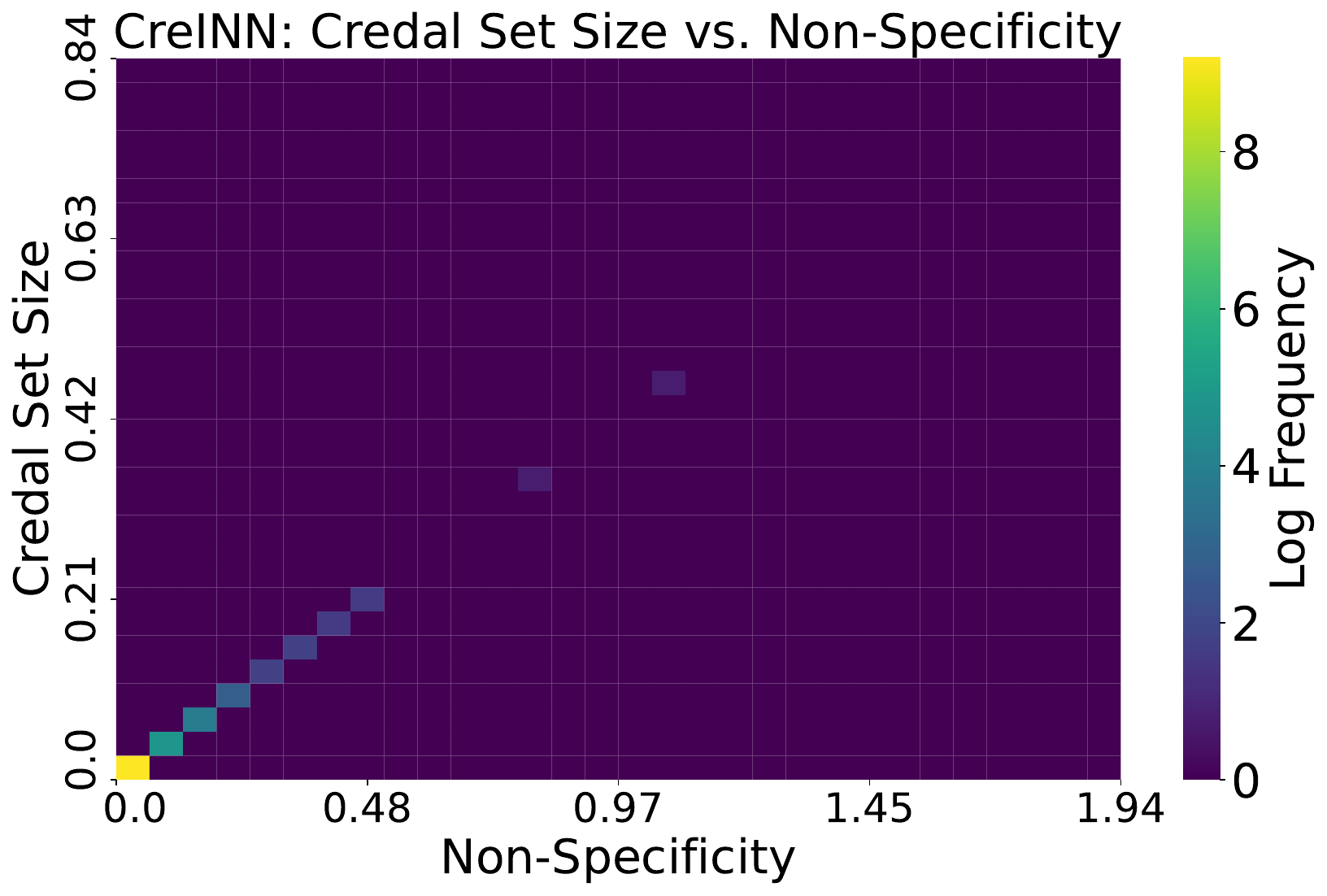}
    \caption*{(d)}
    \end{minipage} \hspace{0.01\textwidth}
        \begin{minipage}[t]{0.3\textwidth}
    \includegraphics[width=\textwidth]{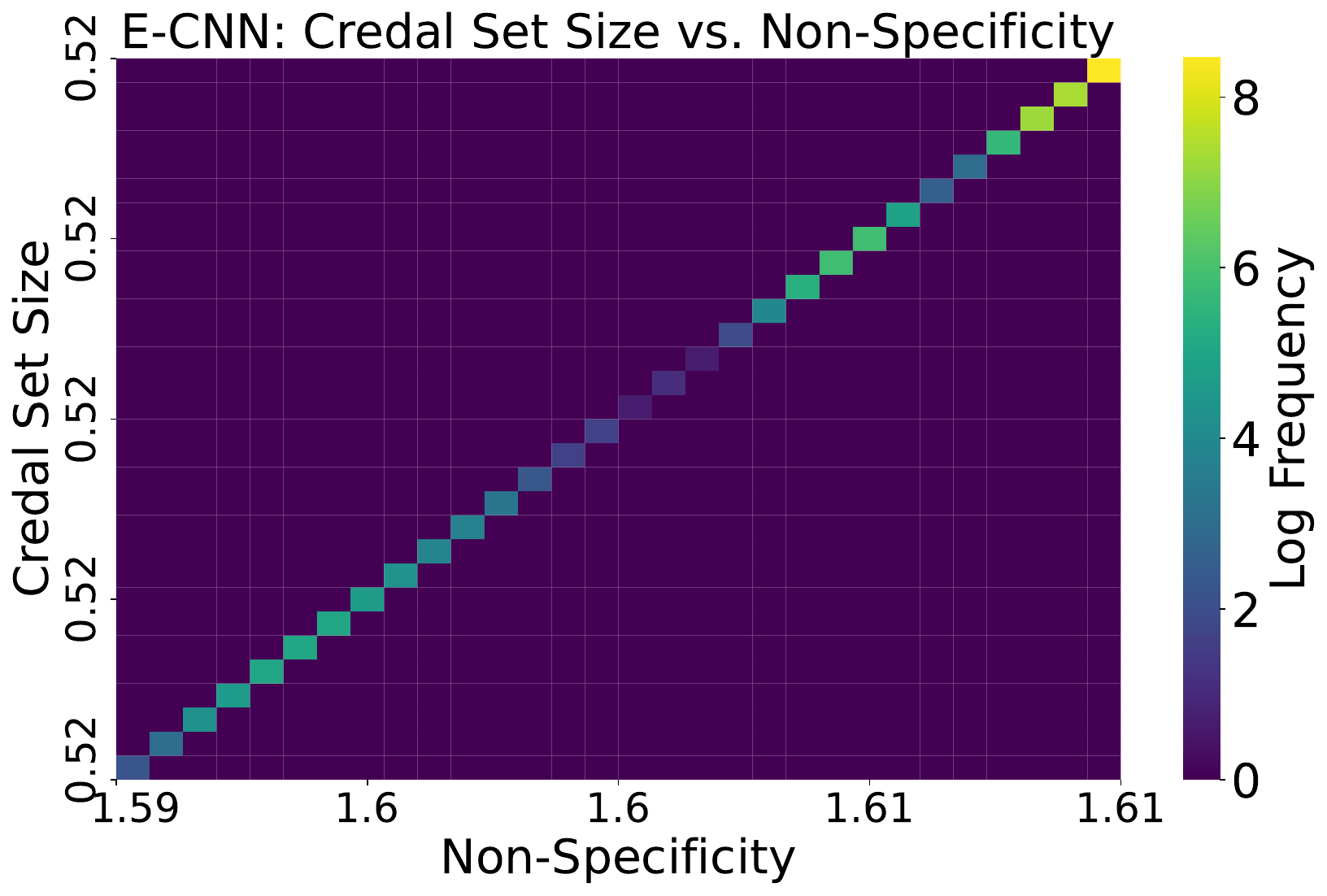}
    \caption*{(e)}
    \end{minipage}\hspace{0.01\textwidth}
            \begin{minipage}[t]{0.3\textwidth}
    \includegraphics[width=\textwidth]{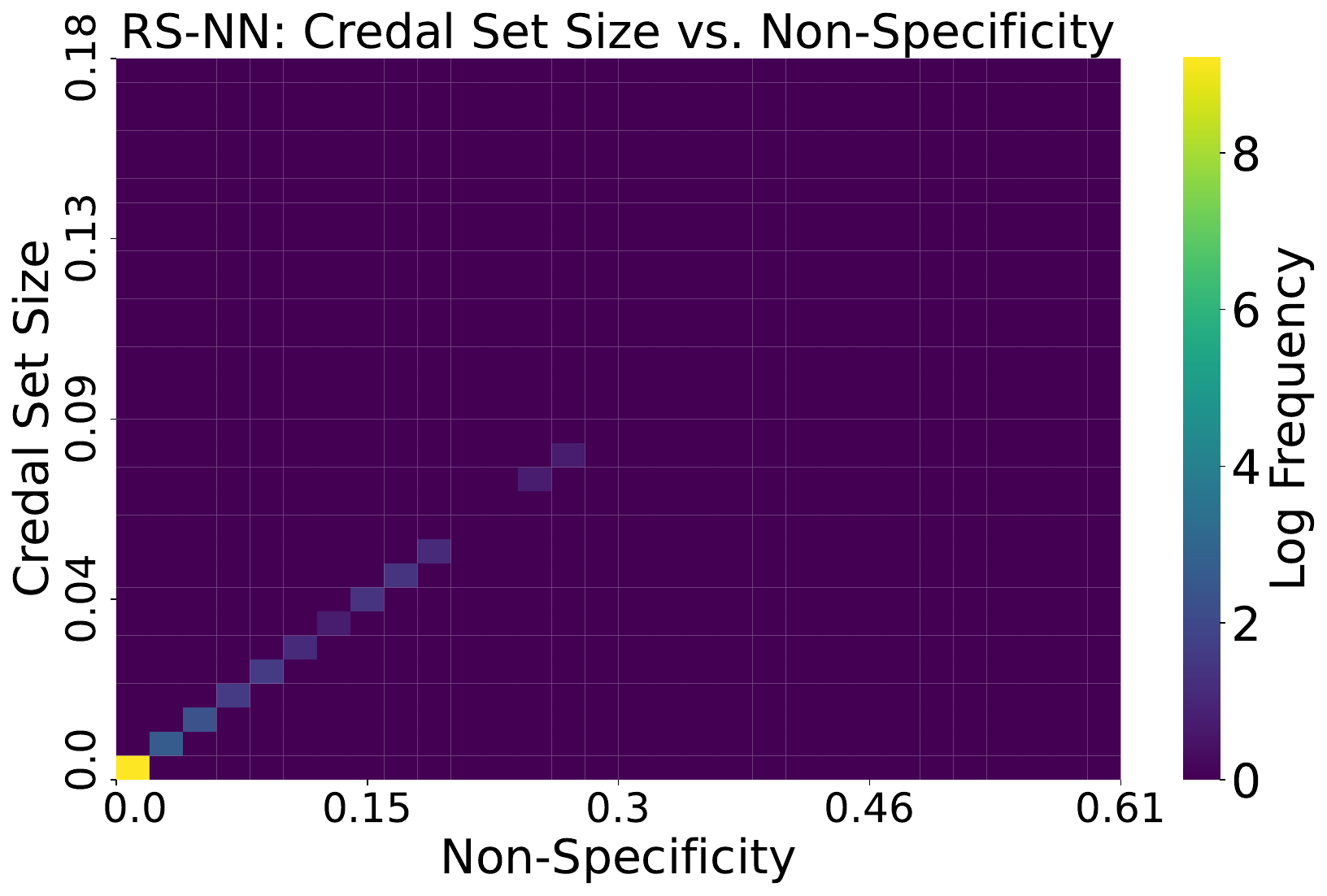}
    \caption*{(f)}
    \end{minipage}
    \caption{Credal Set Size vs. Non-Specificity heatmap for (a) LB-BNN, (b) Deep Ensembles (DE), (c) EDL, (d) CreINN, (e) E-CNN and (f) RS-NN on the MNIST dataset. Credal Set Size and Non-Specificity are directly correlated to each other. Log frequency is used to better showcase the trend.
}\label{fig:app_credal_vs_non_specificity_MNIST}
\end{figure}

\begin{figure}[!h]
    \centering
    \begin{minipage}[t]{0.3\textwidth}
    \includegraphics[width=\textwidth]{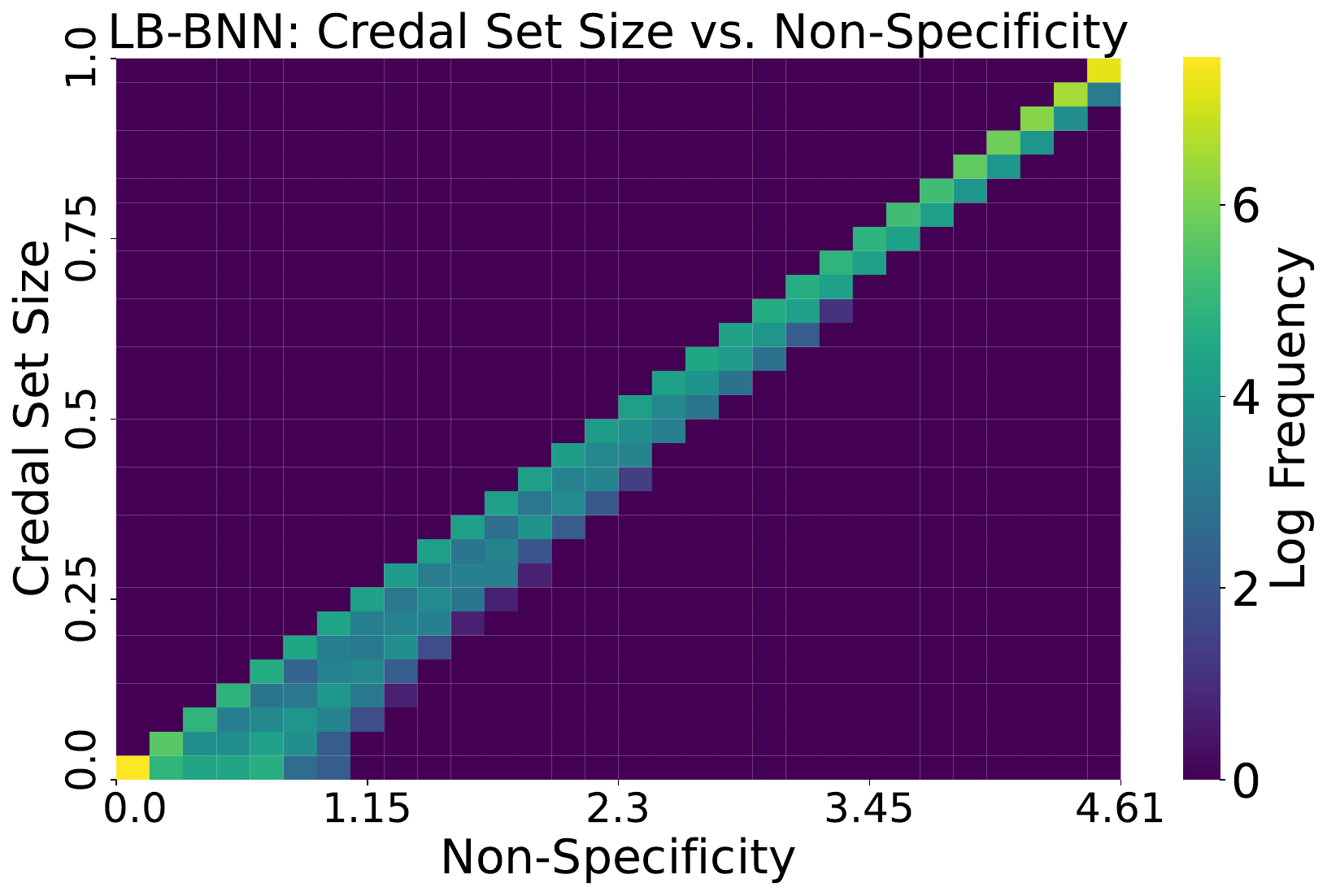}
    \caption*{(a)}
    \end{minipage} \hspace{0.01\textwidth}
    \begin{minipage}[t]{0.3\textwidth}
    \includegraphics[width=\textwidth]{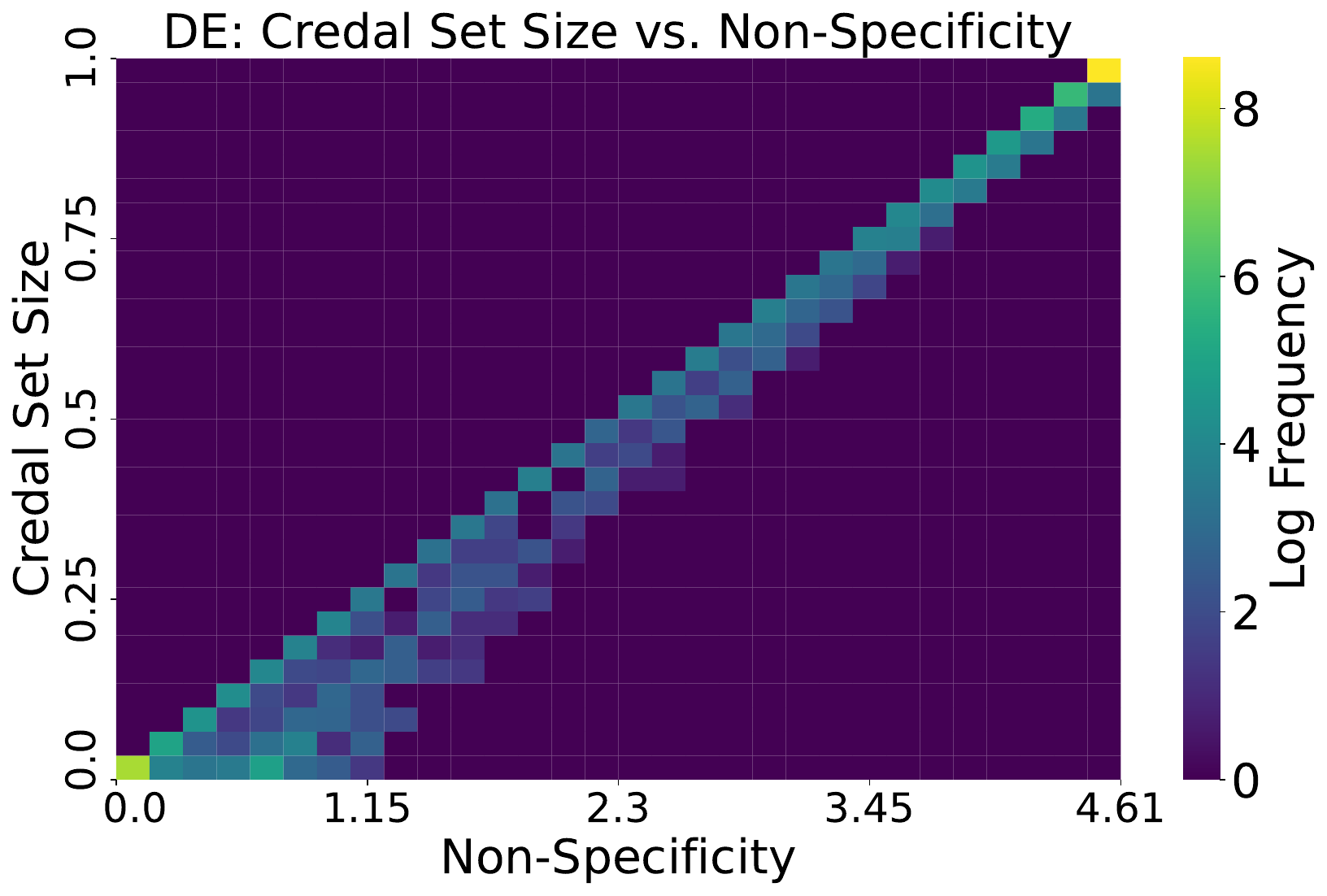}
    \caption*{(b)}
    \end{minipage} \hspace{0.01\textwidth}
        \begin{minipage}[t]{0.3\textwidth}
    \includegraphics[width=\textwidth]{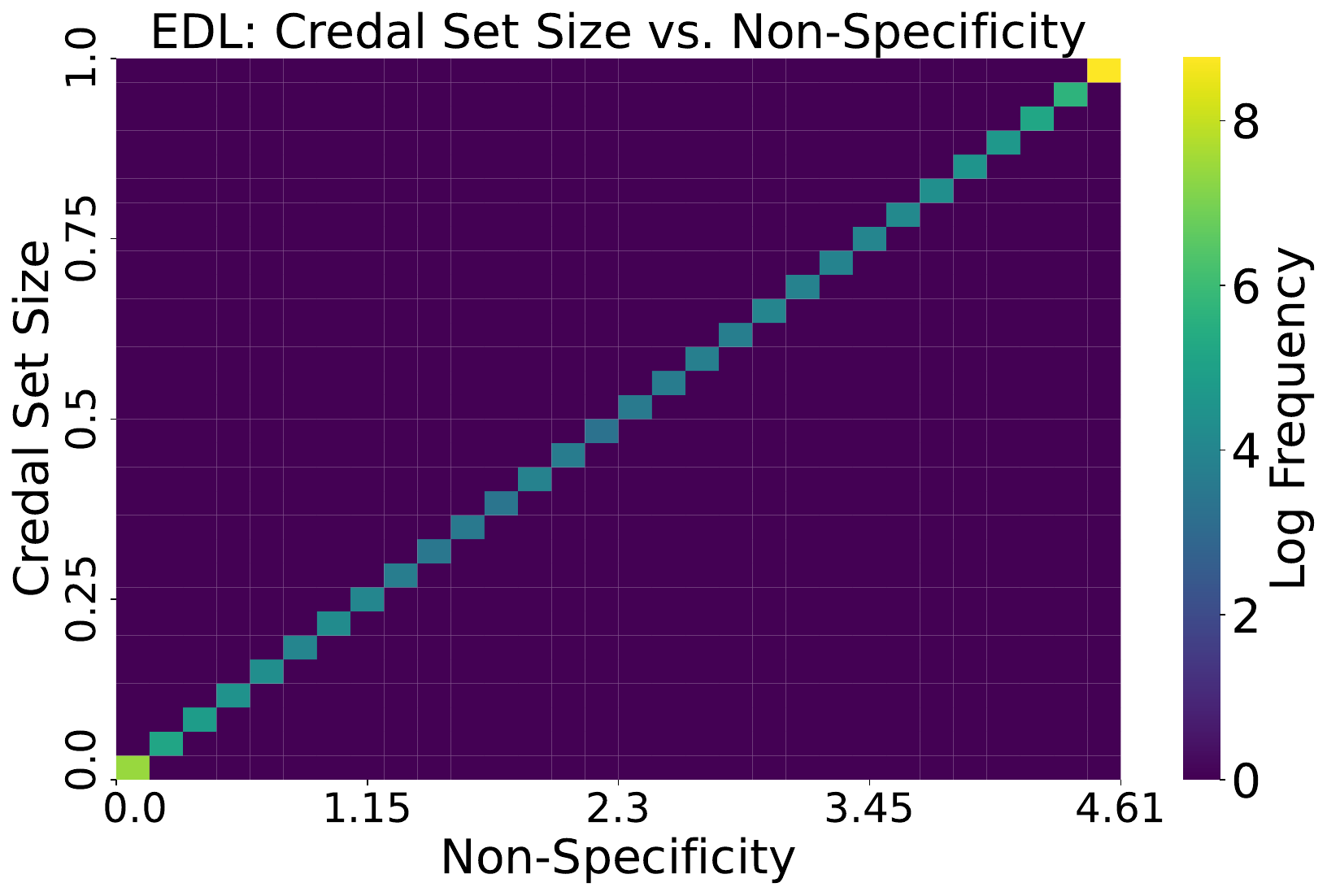}
    \caption*{(c)}
    \end{minipage}
    \\
        \begin{minipage}[t]{0.3\textwidth}
    \includegraphics[width=\textwidth]{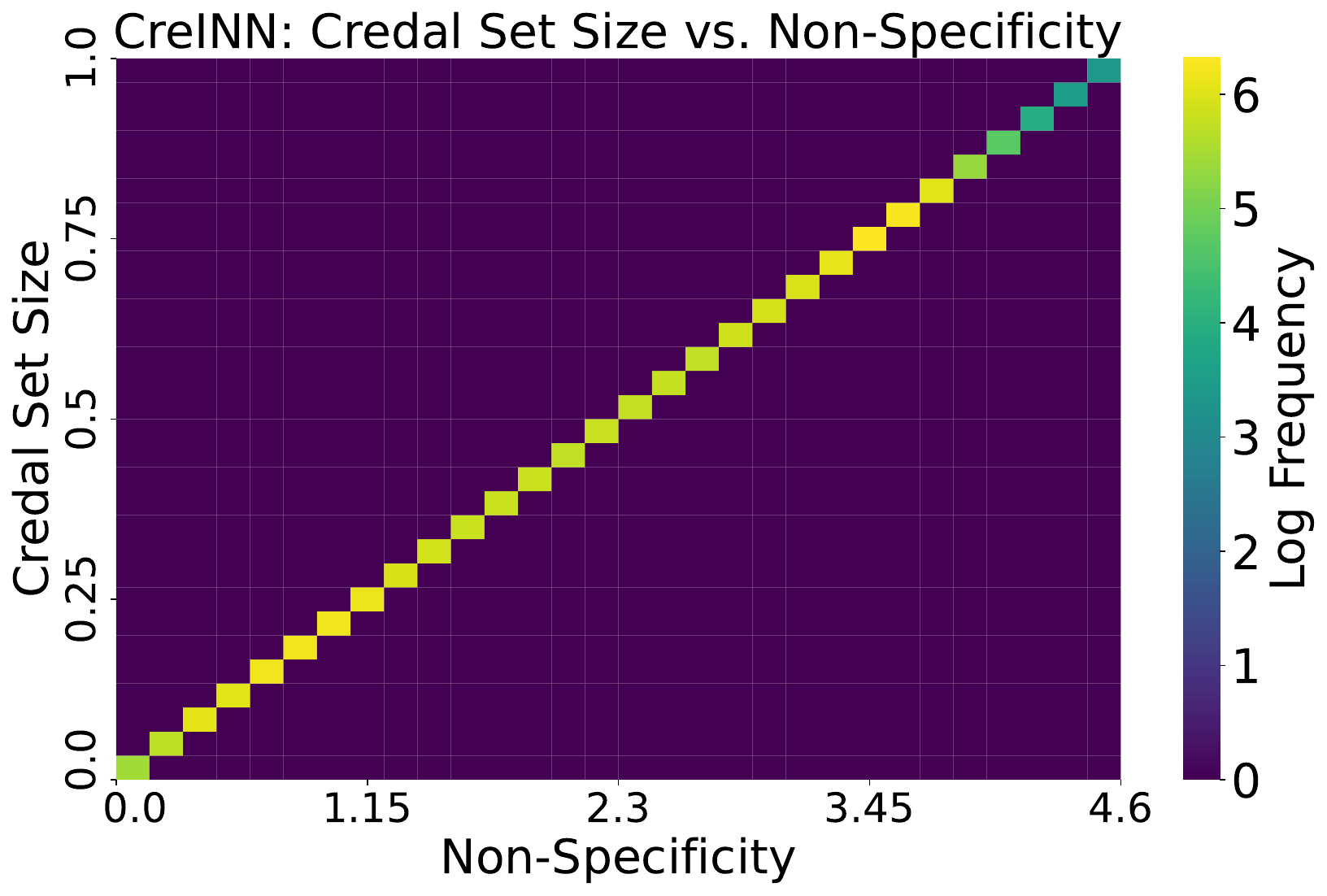}
    \caption*{(d)}
    \end{minipage}\hspace{0.01\textwidth}
            \begin{minipage}[t]{0.3\textwidth}
    \includegraphics[width=\textwidth]{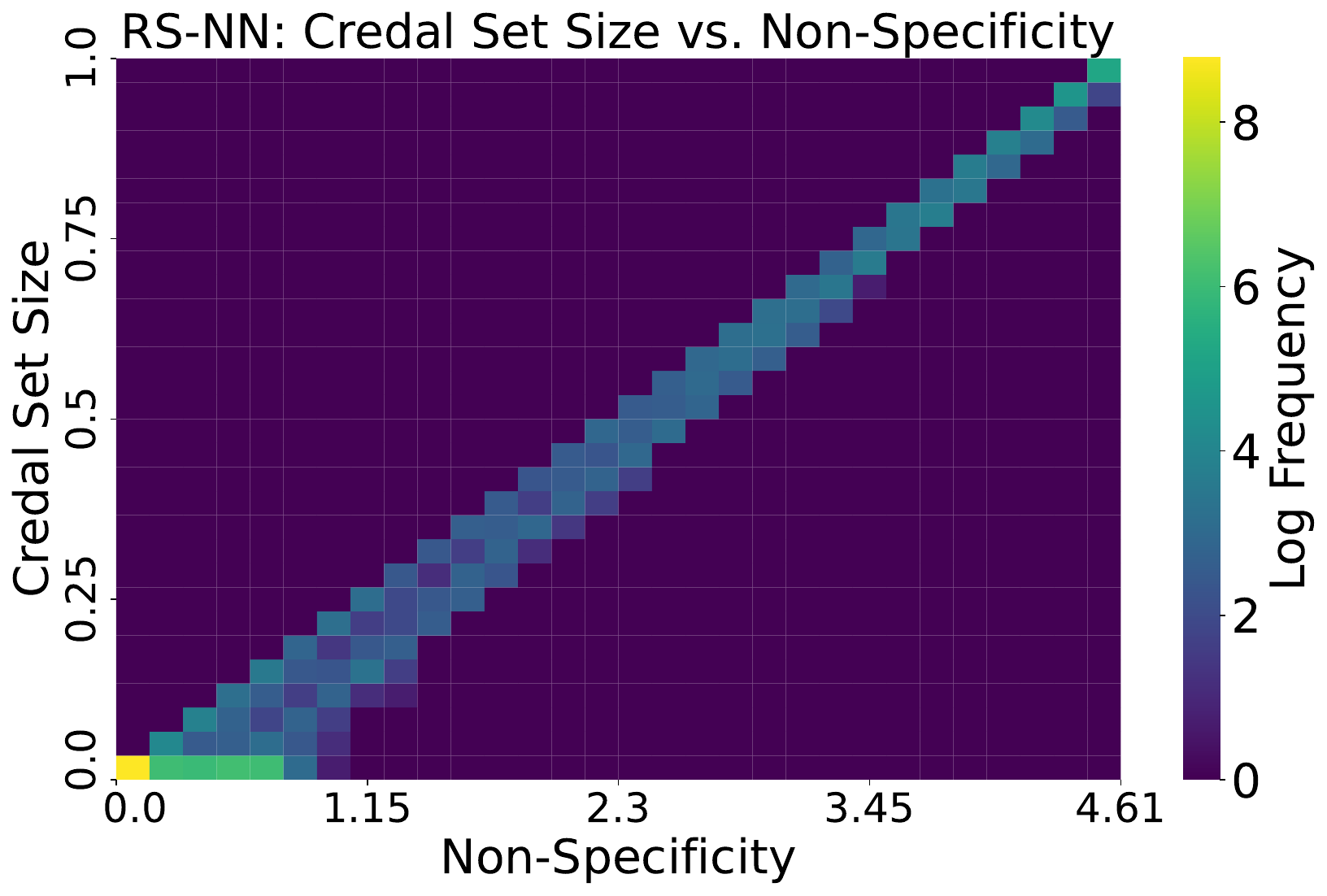}
    \caption*{(e)}
    \end{minipage}
    \caption{Credal Set Size vs. Non-Specificity heatmap for (a) LB-BNN, (b) Deep Ensembles (DE), (c) EDL, (d) CreINN and (e) RS-NN on the CIFAR-100 dataset. Credal Set Size and Non-Specificity are directly correlated to each other. Log frequency is used to better showcase the trend.
}\label{fig:app_credal_vs_non_specificity_cifar100}
\end{figure}

\end{document}